\documentclass{article}

\usepackage{arxiv}

\usepackage[utf8]{inputenc} 
\usepackage[T1]{fontenc}    
\usepackage{hyperref}       
\usepackage{url}            
\usepackage{booktabs}       
\usepackage{amsfonts}       
\usepackage{nicefrac}       
\usepackage{microtype}      
\usepackage{lipsum}		
\usepackage{graphicx}
\usepackage[square,sort,comma,numbers]{natbib}
\usepackage{doi}
\usepackage{xcolor}
\usepackage{array}
\usepackage{amsmath}

\def\OCML{OCML}
\def\MetaBCE{Meta-BCE}

\def\eg{\emph{e.g.}}

\def\etal{\emph{et al.}}

\title{One-Class Meta-Learning: Towards Generalizable Few-Shot Open-Set Classification}

\author{Jedrzej Kozerawski\\
	Department of Electrical \& Computer Engineering\\
	University of California, Santa Barbara\\
	Santa Barbara, CA \\
	\texttt{jkozerawski@ucsb.edu} \\
	\And
	Matthew Turk \\
	Toyota Technological Institute at Chicago\\
	6045 South Kenwood Ave,\\
	Chicago, IL \\
	\texttt{mturk@ttic.edu} \\
}

\date{}

\begin{document}
\maketitle

\begin{abstract}
	Real-world classification tasks are frequently required to work in an open-set setting. This is especially challenging for few-shot learning problems due to the small sample size for each known category, which prevents existing open-set methods from working effectively; however, most multiclass few-shot methods are limited to closed-set scenarios. In this work, we address the problem of few-shot open-set classification by first proposing methods for few-shot one-class classification and then extending them to few-shot multiclass open-set classification. We introduce two independent few-shot one-class classification methods: Meta Binary Cross-Entropy (\MetaBCE), which learns a separate feature representation for one-class classification, and One-Class Meta-Learning (\OCML), which learns to generate one-class classifiers given standard multiclass feature representation. Both methods can augment any existing few-shot learning method without requiring retraining to work in a few-shot multiclass open-set setting without degrading its closed-set performance. We demonstrate the benefits and drawbacks of both methods in different problem settings and evaluate them on three standard benchmark datasets, miniImageNet, tieredImageNet, and Caltech-UCSD-Birds-200-2011, where they surpass the state-of-the-art methods in the few-shot multiclass open-set and few-shot one-class tasks.
\end{abstract}



\section{Introduction}\label{sec:intro}

\begin{figure}[h!]
\begin{center}

\begin{tabular}{cc}
\includegraphics[width=8cm]{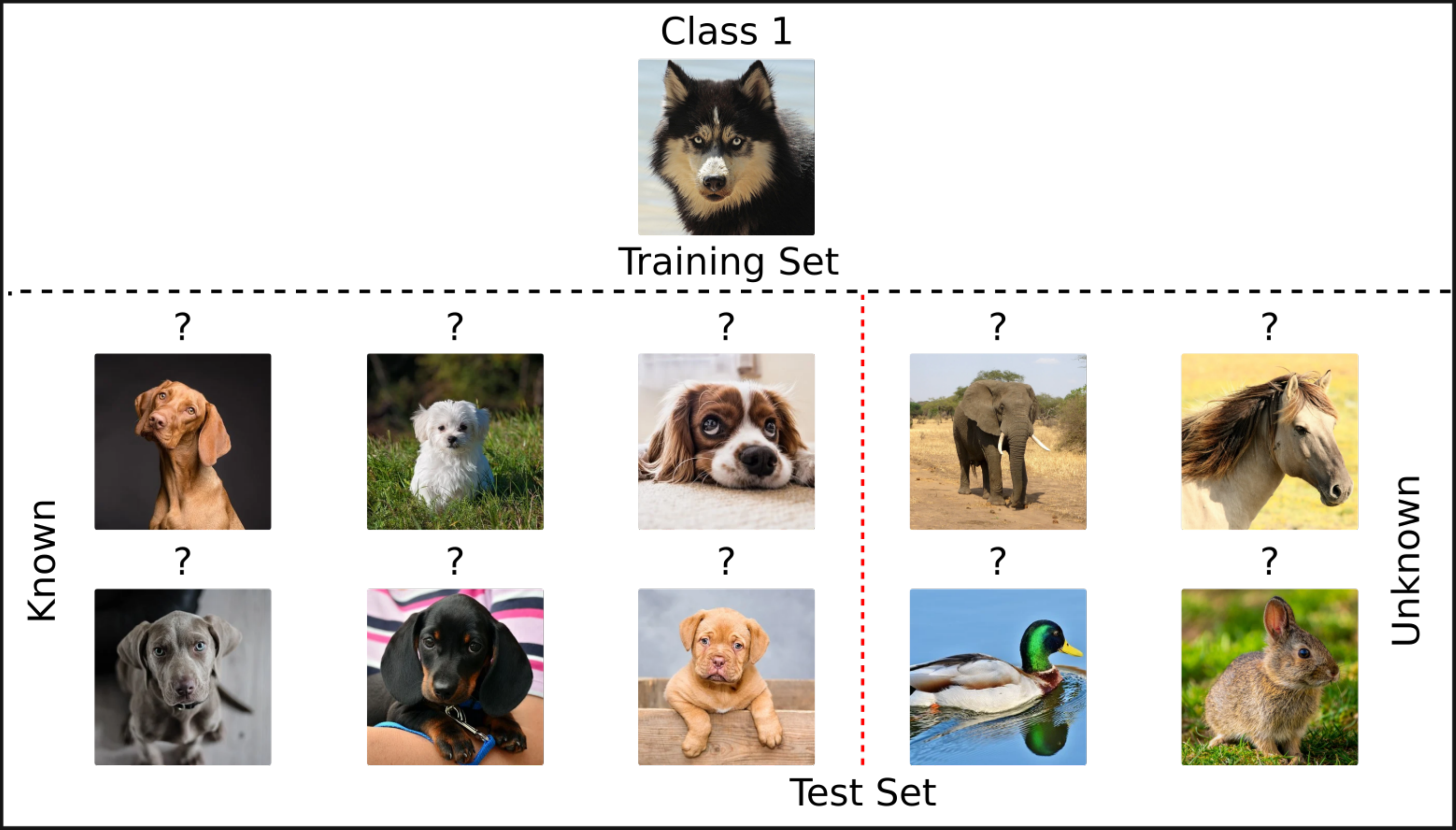}&
\includegraphics[width=8cm]{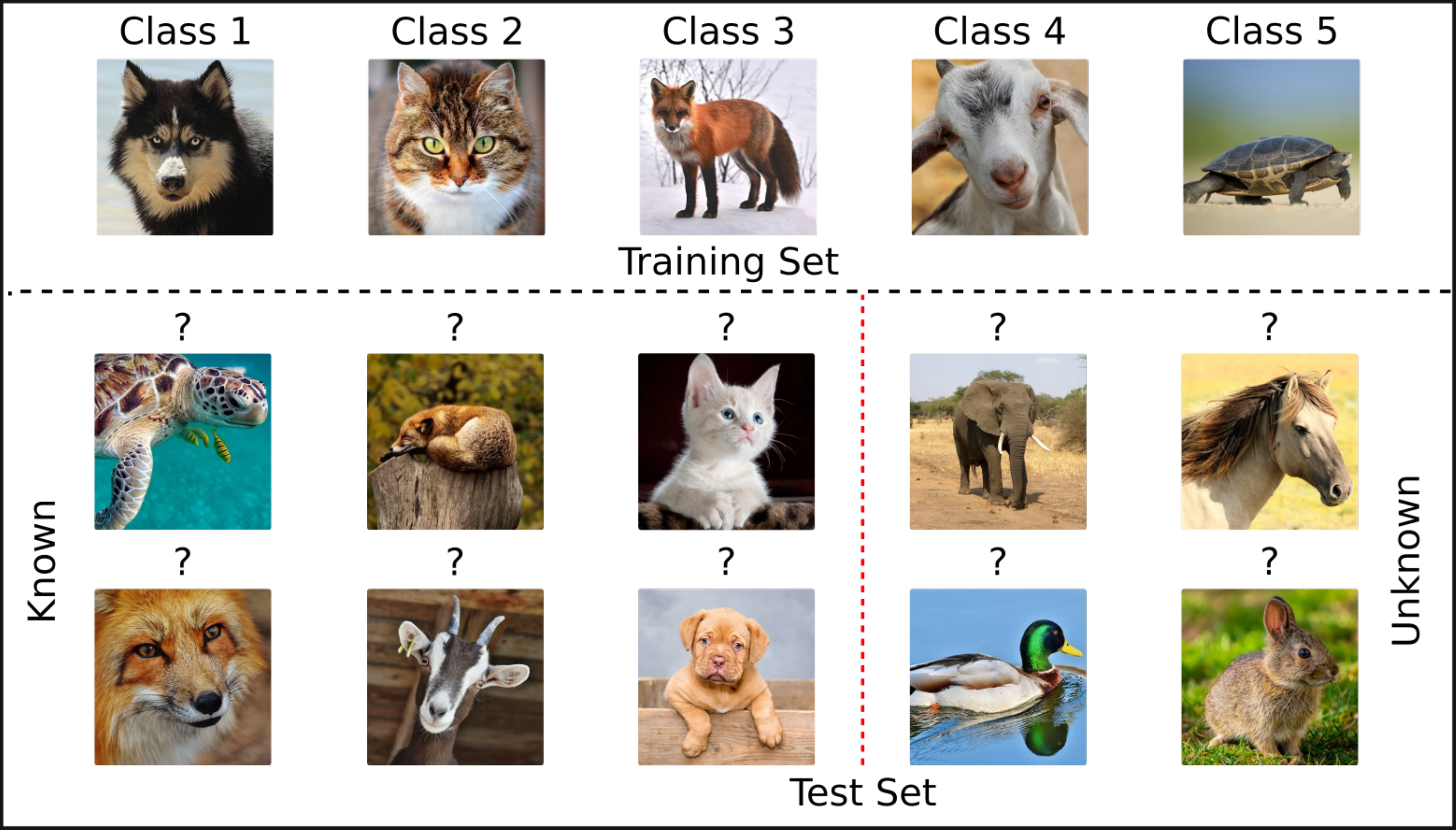}\\
(a) Few-Shot One-Class & (b) Few-Shot Multiclass Open-Set
\end{tabular}
\caption{ (a) Few-Shot One-Class classification and (b) Few-Shot Multiclass Open-Set problems. The training set (top) contains few (one in this case) labeled training examples from a single category (a) or five categories (b). The test set (bottom) contains unseen examples from known categories present in the training set treated as a known categories (Known - bottom left) and examples from unknown categories not present in the training set (Unknown - bottom right). The goal is to train a model given examples from the training set such that it would be able to differentiate between examples coming from the known categories (Known examples) and from other, unknown categories (Unknown examples), while simultaneously correctly assigning examples predicted as Known into one of $n$ known categories.}
\label{fig:data}
\end{center}
\end{figure}

Deep learning methods are able to achieve high performance on large-scale visual recognition tasks~\cite{huang2017densely,russakovsky2015imagenet,szegedy2016rethinking,he2016deep}, but the quality of the learned representations greatly depends on the amount of available training data. In many classification tasks this quantity is not sufficient to properly train a neural network that would generalize well to unseen examples. Obtaining more training data is often not feasible for numerous reasons, such as the natural rarity of specific categories (\eg,~rare diseases or events), necessity of fast adaptation to novel tasks (\eg,~early detection of new disease such as COVID-19 or recognition of a new type of car by the autonomous driving system), financial constraints, and researchers/scientists needing to train a specialized model on their own small-scale dataset.

Few-shot learning~\cite{ravi2016optimization, snell2017prototypical} is a problem in image classification dealing with small sets of training data per class, where the task is to train a classifier on data from a small, labeled support set $\mathcal{S}$ (consisting of data from $n$ categories, each represented with $k$ examples -- $k$-shot $n$-way classification) and use a query set $\mathcal{Q}$ to assess the quality of this classifier. Recent meta-learning approaches~\cite{snell2017prototypical, ye2020fewshot, gidaris2018dynamic}, designed for better generalization capabilities of trained models to unseen data, help increase the performance while simultaneously reducing the training time necessary to adjust to the novel data. However, existing approaches unfortunately assume that the query set $\mathcal{Q}$ has to contain examples only from the same $n$ categories as the $\mathcal{S}$ (a closed-set setting), which is very restrictive. Recently Liu~\etal~\cite{liu2020few} introduced the few-shot open-set setting, where during the inference time the query set $\mathcal{Q}$ might contain additional unknown open-set categories $\mathcal{U}$ (that are not present in the support set $\mathcal{S}$) that need to be differentiated from the known classes $\mathcal{K}$ ($\mathcal{Q} = \mathcal{K} \cup \mathcal{U}$) before performing the multiclass classification step. Standard open-set approaches~\cite{bendale2016towards, scheirer2012toward, rudd2017extreme} assume access to high number of per-category examples in order to model their distribution and detect out-of-distribution samples (unknown $\mathcal{U}$), however a small samples size in few-shot learning prevents those methods from working well (or frequently at all when number of per-category examples drops to one). This indicates the need for new meta-learning approaches capable of learning to detect the unknown examples. Initial approach introduced by Liu~\etal~\cite{liu2020few} unfortunately provides only a relative ranking score for all examples without any method to clearly differentiate between known examples $\mathcal{K}$ and unknown examples $\mathcal{U}$, provides low closed-set accuracy and cannot address a problem when $n=1$ since it operates on softmax scores. In this work we propose to address these issues by studying meta-learning approaches capable of out-of-distribution detection given only few training examples (few-shot one-class classification), and extending them to few-shot multiclass open-set problem setting (as seen in Fig.~\ref{fig:data}). Proposed approaches will have the following benefits:

\begin{itemize}
    \setlength{\parskip}{0pt}
    \setlength{\itemsep}{0pt plus 1pt}
    \item They can be used to augment any existing few-shot multiclass classification approach (such as FEAT~\cite{ye2020fewshot} or PEELER~\cite{liu2020few}) to operate in an open-set setting without requiring retraining and without a performance drop in the closed-set setting.
    \item After training, our approach can work in both the few-shot open-set and closed-set problem settings with any number of categories ($n$-way) and per-category examples ($k$-shot), even when both $n=1$ and $k=1$ (one-shot one-class).
    \item They do not require separate background (unknown) categories present during training (contrary to existing open-set methods).
\end{itemize}

Our contributions in this work can be summarized as:

\begin{itemize}
    \setlength{\parskip}{0pt}
    \setlength{\itemsep}{0pt plus 1pt}
    \item We present two novel meta-learning methods for few-shot one-class image classification that are capable of augmenting any existing few-shot multiclass classification approach to work in few-shot multiclass open-set classification setting.
    \item We verify the value of the proposed approaches using few-shot one-class and few-shot multiclass open-set experiments by reporting performance on two benchmark few-shot datasets.
\end{itemize}

\section{Related work}\label{sec:related_work}

\subsection{Few-shot classification}

Few-shot classification refers to a problem where a model is trained to generalize to novel, unseen samples in the query set $\mathcal{Q}$ represented by a small number of examples in the support set $\mathcal{S}$. There are two main types of approaches when addressing this problem.
One type of methods is concentrated around metric learning for better similarity and relation embeddings. Siamese networks~\cite{kochsiamese} compute the similarity score between two images, while Matching Networks~\cite{vinyals2016matching} learn classifiers for novel categories based on a mapping from a small support set of examples (input-label pairs) to a classifier for the given example. Snell~\etal~\cite{snell2017prototypical} used Prototypical Networks for few-shot learning by representing each class by the mean of its examples in an embedding space learned by the neural network. Sung~\etal.~\cite{sung2018learning} presented Relation Networks that consist of two modules, an embedding module and a relation module, learning the appropriate relation between a query image and each of $k$ images in $k$-way classification. Liu~\etal~\cite{liu2020few} used a modified prototypical network approach to tackle the few-shot open-set problem. Qi~\etal~\cite{qi2018low} proposed to imprint the weights of a new classifier with an embedding vector extracted from a base classifier pre-trained on known categories. This idea is very similar to the Gidaris and Komodakis approach~\cite{gidaris2018dynamic}. Ye~\etal~\cite{ye2020fewshot} proposed to use a transformer on top of the prototypical network embeddings to learn a better mapping for class representations. Another set of ideas focused on optimization approaches to solve this problem. Ravi and Larochelle~\cite{ravi2016optimization} used an LSTM to optimize updates while training a network on a meta-set consisting of multiple datasets. Li~\etal~\cite{li2017meta} presented a Meta-SGD approach based on Stochastic Gradient Descent. Finn~\etal~\cite{finn2017model} introduced Model-Agnostic Meta-Learning (MAML).

\subsection{One-class classification}

In one-class classification only data from positive category is available during training. Sch{\"o}lkopf~\etal~\cite{scholkopf2000support} addressed this with a One-Class Support Vector Machine (OC-SVM) which maximizes the margin between the origin and one-class samples. Chen~\etal~\cite{chen2001one} also used the One-Class SVM in the image retrieval problem. Tax and Duin~\cite{tax1999data,tax1999support} introduced Support Vector Data Description (SVDD) and later augmented the method by generating artificial outliers~\cite{tax2001uniform}. Ruff~\etal~\cite{pmlr-v80-ruff18a} proposed Deep Support Vector Data Description (Deep-SVDD) to train a deep feature extractor jointly with the one-class classification objective. Perera and Patel~\cite{perera2019learning} introduced two loss functions (compactness and descriptiveness loss) together with a template matching matching framework for deep one-class classification. Sabokrou~\etal~\cite{sabokrou2018adversarially} utilized a two-network architecture trained in a GAN-style adversatial learning framework for adversarially learned one-class classification. Kemmler~\etal~\cite{kemmler2013one} utilized Gaussian Process (GP) priors for one-class classification. Kozerawski and Turk~\cite{kozerawski2018clear} proposed CLEAR to predict the hyperparameters of a one-class SVM classifier given a single positive example (one-shot one-class classification).

\subsection{Open-set classification}

Open-set classification is a machine learning problem when during the inference stage the set of observable examples (the query set $\mathcal{Q}$) can include unknown examples $\mathcal{U}$ coming from unknown categories apart from known examples $\mathcal{K}$ coming from known categories (present in the support set $\mathcal{S})$. Scheirer~\etal~\cite{scheirer2012toward} introduced a new variant of SVM (1-vs-set machine) based on the risk minimization. Bendale~\etal~\cite{bendale2016towards} proposed OpenMax, a method using extreme value theory to re-evaluate logit values. Ge~\etal~\cite{ge2017generative} augmented OpenMax to a generative variant called G-OpenMax. Neal~\etal~\cite{neal2018open} also proposed a generative approach sythesizing open-set images while training, which helps detecting unknown examples during the inference time. Liu~\etal~\cite{liu2019large} introduced a method for long-tail open-set recognition with distance-based methods. Dhamija~\etal~\cite{dhamija2018reducing} proposed two losses (Entropic Open-Set and Objectosphere) to maximize the difference in the softmax output for known and unknown samples. Yoshihashi~\etal~\cite{yoshihashi2019classification} proposed CROSR (Classification-Reconstruction learning for Open-Set Recognition) where they jointly perform classification and reconstruction of the input data. Rudd~\etal~\cite{rudd2017extreme} introduced extreme value machines which utilize extreme value theory to model the probability of example coming from unknown category. Liu~\etal~\cite{liu2020few} proposed a new problem setting of few-shot open-set recognition and utilized method based on Prototypical Networks~\cite{snell2017prototypical} combined with entropy-based loss function.

\section{Meta Learning for Few-Shot Open-Set Classification}\label{sec:methods}

Many traditional open-set classification approaches~\cite{rudd2017extreme, liu2020energy, sun2020conditional, bendale2016towards} differentiate known from unknown examples by modeling the distribution of known classes and frequently focus on modeling the tail of this distribution using the Extreme Value Theory. In few-shot learning modeling the distribution of a known category and analyzing the tail of this distribution when the number of examples is close to zero is not feasible (or even not possible in case of number of examples $k=1$). Additionally, in few-shot learning the set of known categories is different during meta-training and meta-testing phase, thus preventing standard open-set methods from working efficiently. For these reasons there is a need for an open-set meta-learning approach that can work well in few-shot setting. Lack of sufficient training examples is also problematic for existing one-class classification approaches~\cite{scholkopf2000support, tax2004support, pmlr-v80-ruff18a} that need abundance of positive training examples to model well the distribution of the positive class in order to detect any anomalous examples (out-of-distribution detection). We propose to solve mentioned problems with an introduction of two separate few-shot one-class meta-learning approaches capable of detection of unknown examples in both one-class and multiclass open-set settings. Proposed methods work as separate modules that can augment any existing closed-set few-shot multiclass classification method to work in few-shot multiclass open-set setting without a need to retrain it.

\begin{figure}[h!]
\centerline{\fbox{\includegraphics[width=1.0\textwidth]{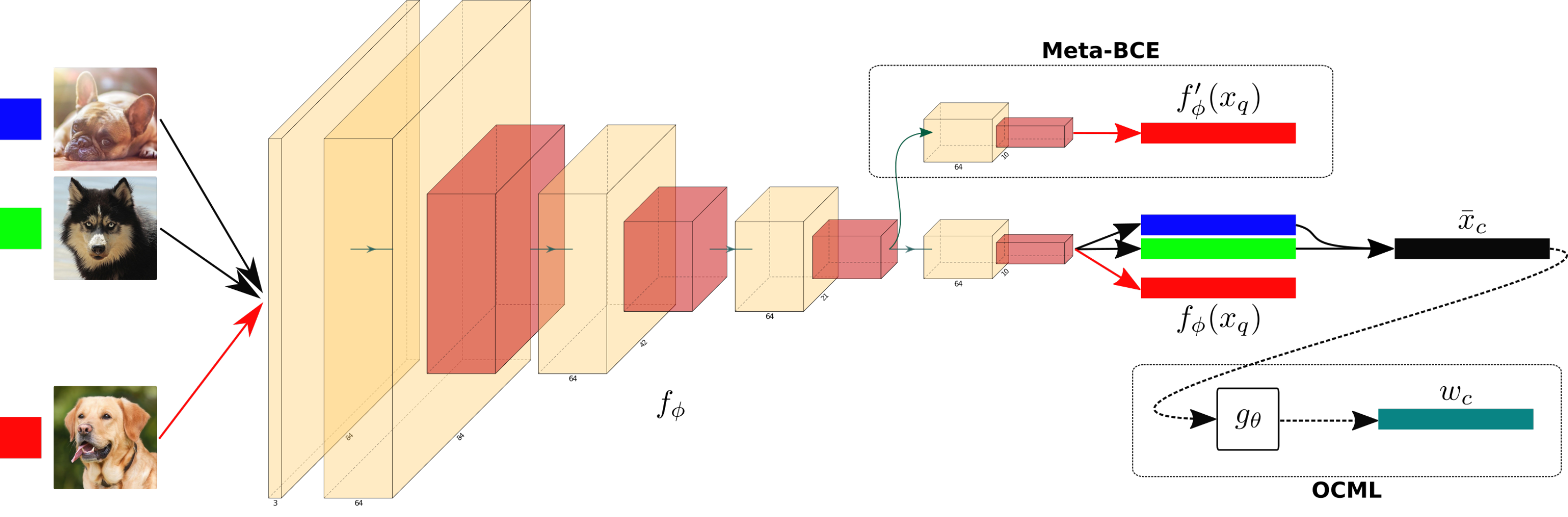}}}
\caption{An overview of \OCML~and \MetaBCE~classification methods. Both approaches are independent and standalone modules (indicated with dashed lines) trained and utilized separately from each other. $k$ images belonging to the known support class $c$ have their features extracted using a CNN $f_{\phi}$ and a prototype representation $\bar{x}_c$ is calculated, which is used for closed-set classification and as an input to \OCML. In the example above, $k=2$ with two upper images ({\color{blue} blue} and {\color{green} green}) as the known examples from the support class $c$, and the {\color{red} red} example is a query example. One-class classification is performed either using either Eq.~\ref{eq:BCE_probability} (when using \MetaBCE) or Eq.~\ref{eq:OCML_probability} (when using \OCML).}
\label{fig:method_flow}

\end{figure}

\subsection{Meta Binary Cross-Entropy (\MetaBCE)}

Let us divide the dataset $\mathcal{D}$ by following a standard few-shot meta-learning setting~\cite{vinyals2016matching, ravi2016optimization} into three separate meta-sets: meta-training meta-set $\mathcal{D}_{meta-train}$, meta-validation meta-set $\mathcal{D}_{meta-val}$, and meta-testing meta-set $\mathcal{D}_{meta-test}$. Each of the meta-sets has a separate, non-overlapping set of categories. We utilize episodic training as is a standard practice in few-shot learning~\cite{vinyals2016matching, snell2017prototypical, ye2020fewshot}.

Following the work of Liu~\etal~\cite{liu2020few}, let us use a separate branch of the feature extractor $f_{\phi}$ indicated as $f_{\phi}'$, but instead of using it for multiclass features (as done by Liu) let us use it to learn a separate feature representation for one-class classification (instead of only multiclass classification). We hypothesize that high quality feature representation for one-class classification differs from that for multiclass classification (which is backed by our results). Standard multiclass classifiers answer the question ``Which of the $n$ classes does this new, unknown example resemble the most?''. Answering this question depends highly on the composition of the few-shot task (i.e., how similar are the categories and how many of them there are). In order to eliminate that dependency, we need to learn a different feature space representation where the probability of a new example belonging or not belonging to the known category depends only on the known positive examples from that category irrespective of other categories. We propose to use a binary cross-entropy loss in a meta-learning setting to simultaneously learn a one-class feature representation on the branch $f_{\phi}'$ and a one-class classification decision boundary:

\begin{equation}
\mathcal{L}_{Meta\mbox{-}BCE} = -\frac{1}{N} \sum_{(x_i, y_i) \in \mathcal{D}_{meta-train}}  \sum_{c=1}^{N} (y_{i,c}\log p(y_c|x_i) + (1-y_{i,c})(1-p(y_c|x_i)))
\label{eq:BCE_loss}
\end{equation}

\noindent where $y_{i,c}$ is a binary indicator (0 or 1) if class label $c$ is the correct classification for observation $i$, and $p(y_c|x_i)$ is the probability of the example $x_i$ belonging to the category $y_c$:

\begin{equation}
\begin{split}
p(y=c|x_i) = \frac{1}{1+exp( -(-d(f_{\phi}'(x_i), \bar{x}_c) - t)) }
= \frac{1}{1+exp( d(f_{\phi}'(x_i), \bar{x}_c) + t) }
\end{split}
\label{eq:BCE_probability}
\end{equation}

\noindent where $t$ is a learnable parameter. You can see the feature extractor, $f_{\phi}$, one-class branch $f_{\phi}'$, and \MetaBCE~approach in Figure~\ref{fig:method_flow}

\subsection{One-Class Meta-Learning (\OCML)}

One-class classifiers should be ideally adapted to the class of interest, which in few-shot scenario is difficult to achieve given small sample size. We propose to use meta-learning to learn how to generate dynamically parameters of a one-class classifier for a few-shot category. Kozerawski and Turk addressed this issued by allowing to dynamically predict the parameters of a one-class SVM classifier~\cite{kozerawski2018clear}, however the method was limited to only a one-shot setting as the transfer learning module was using a feature vector of a single image as an input, and the method had multiple, separate stages of training with a final classification limited to using SVMs or logistic regression. To overcome these limitation we propose One-Class Meta-Learning (\OCML), a method that is able to dynamically create a one-class neural network classifier for a novel category with more than a single image in the support set $\mathcal{S}$ and is trained jointly with the feature extractor in a single stage. \OCML~has a transfer learning module $g_{\theta}$ that learns how to transform the feature representation of a category $c$ ($\bar{x}_c$) to a weight vector $w_c$ of the one-class classifier for the category $c$:

\begin{equation}
w_c= g_{\theta}( f_{\phi}( x_c ) ) 
\label{eq:OCML_vector}
\end{equation}

After obtaining the weight of the one-class classifier $w_c$, we can calculate the probability of a novel example $x_i$ belonging to the category $c$:

\begin{equation}
p(y=c|x_i) = \frac{1}{1+exp( -w_c \cdot f_{\phi}(x_i) ) }
\label{eq:OCML_probability}
\end{equation}

We use the binary cross-entropy loss using the probabilities calculated with the Eq.~\ref{eq:OCML_probability} to train jointly both networks $f_{\phi}$  and $g_{\theta}$. The overview of \OCML~can be seen in Figure~\ref{fig:method_flow}.

\subsection{Few-Shot Open-Set}

In the standard open-set setting we have access to an abundance of examples from known classes; however, in the few-shot open-set setup the known classes will be revealed in the meta-testing phase (as the model is trained on a separate, non-overlapping set of classes in  meta-training). This means that the algorithm should be able to work on any given set of known categories. To address this issue, we utilize a divide-and-conquer approach and divide the problem of detecting samples not belonging to any of $n$ known categories into $n$ smaller problems of detecting samples not belonging to a single category (one-class problem). This allows the use of meta-learning more efficiently, as one-class learning depends only on positive examples from a single class, whereas multiclass open-set depends also on the composition of the task (what classes are known and how many of them). Additionally, few-shot one-class classification can be thought of as a special case of few-shot multiclass open-set classification, where in the $n$-way $k$-shot scenario, $n=1$. In order to adapt both \OCML~and \MetaBCE~to a multiclass scenario ($n\geq2$) we treat the few-shot open-set problem as an ensemble of $n$ few-shot one-class problems and modify the prediction equation for \MetaBCE:

\begin{equation}
p_(y\in\mathcal{U}|x_i) = \max_{c\in n} (1 - \frac{1}{1+exp(-(-d(f_{\phi}'(x_i), \bar{x}_c) - t)) })
= \min_{c\in n} (\frac{1}{1+exp(d(f_{\phi}'(x_i), \bar{x}_c) + t) })
\label{eq:fsos_BCE_probability}
\
\end{equation}
\noindent and for \OCML:

\begin{equation}
p(y\in\mathcal{U}|x_i) = \max_{c\in n} (1 - \frac{1}{1+exp(-g_{\theta}( f_{\phi}( \bar{x}_c ) ) \cdot f_{\phi}(x_i) ) }) 
= \min_{c\in n} (\frac{1}{1+exp(-g_{\theta}( f_{\phi}( \bar{x}_c ) ) \cdot f_{\phi}(x_i) ) })
\label{eq:fsos_OCML_probability}
\end{equation}

\section{Experimental Results}\label{sec:results}

\subsection{Datasets and performance metrics}

We conducted experiments on three benchmark datasets for few-shot learning: miniImageNet~\cite{vinyals2016matching}, CUB-200-2011~\cite{WahCUB_200_2011}, and tieredImageNet~\cite{ren2018meta}. The miniImageNet dataset is a set of 100 categories (a subset of ImageNet categories~\cite{russakovsky2015imagenet}) with 600 images per category. We follow the meta-training/meta-validation/meta-testing split of the 100 categories to 64/16/20~\cite{ravi2016optimization}. The CUB-200-2011 dataset has 200 visual categories and we follow meta-training/meta-validation/meta-testing split as in Ye~\etal~\cite{ye2020fewshot} to 100/50/50 categories respectively. The tieredImageNet dataset has 608 visual categories and we follow meta-training/meta-validation/meta-testing split as in Ye~\etal~\cite{ren2018meta} to 351/97/60 categories respectively.

To evaluate all methods we have used three metrics in the few-shot one-class settings: accuracy, F1-score, and AUROC (Area Under ROC curve) score. To evaluate the methods in the few-shot open-set setting we have used four metrics: closed-set accuracy (dubbed here accuracy), normalized accuracy (NA), F1-open score, and AUROC score. For details on the performance metrics used here please see Appendix~\ref{sec:fsos_metrics}.


\subsection{Few-Shot One-Class}

In Table~\ref{table:fsoc_mini_val} we present the results for our two proposed approaches (\MetaBCE~and \OCML) on the newly-introduced few-shot one-class task on the miniImageNet dataset~\cite{vinyals2016matching}. We compare our methods with benchmark one-class approaches such as DeepSVDD~\cite{pmlr-v80-ruff18a}, One-Class SVM~\cite{scholkopf2000support}, SVDD~\cite{tax2004support}, DeepAnomaly~\cite{golan2018deep}, and with a one-shot one-class approach introduced by Kozerawski and Turk~\cite{kozerawski2018clear} (CLEAR). As hypothesised, standard many-shot one-class approaches do not translate well to the few-shot setting. DeepSVDD~\cite{pmlr-v80-ruff18a} and DeepAnomaly provide only a relative ranking score for all examples, thus allowing to calculate only AUROC score without accuracy or F1-score. Their AUROC scores are lower than SVDD's~\cite{tax2004support} and OCSVM's~\cite{scholkopf2000support}, which would indicate that a deep network approach might require more training examples to work correctly. In the $1$-shot setting, both OCSVM and SVDD completely overfit to the single examples resulting in a random classification performance, and CLEAR~\cite{kozerawski2018clear} has slightly better classification scores, but lower AUROC score. \OCML~has the best performance in the $1$-shot setting with an accuracy of $68.05\%$, F1-score of $0.620$ and second highest AUROC score of $0.759$. \MetaBCE~has the second highest performance in the $1$-shot setting (accuracy of $57.57\%$ and F1-score of $0.283$) and the highest AUROC score ($0.764$). In the $5$-shot setting, \MetaBCE~becomes the best method with accuracy of $77.38\%$ and F1-score of $0.793$, better than \OCML~(accuracy $77.74\%$ and F1-score $0.693$). Both introduced methods (\MetaBCE~and \OCML) perform much better than any other approaches, with \MetaBCE~performing the best in the $5$-shot setting and \OCML~in the $1$-shot setting. We also provide an upper-bound accuracy obtained using FEAT~\cite{ye2020fewshot} for a $k$-shot $2$-way classification setting when the unknown class is treated as a known one with provided training examples. Calculated upper-bound accuracy shows there is still room for progress as $1$-shot accuracy is $79.76\%$ and $5$-shot accuracy is $88.61\%$. The difference in performance between threshold and \MetaBCE~confirms that multiclass and one-class classification requires different feature representations. The results on the CUB-200-2011 dataset~\cite{WahCUB_200_2011} and on tieredImageNet~\cite{ren2018meta} follow the above conclusions, with \OCML~achieving best performance in the $1$-shot setting ($0.701$ F1-score on CUB and $0.800$ on tieredImageNet), and \MetaBCE~performing best in the $5$-shot setting on CUB ($0.783$ F1-score), while \OCML~performs best on tieredImageNet ($0.878$ F1-score). For details on the CUB-200-2011 and tieredImageNet results please see Appendix~\ref{sec:fsoc_additional}.

\begin{table}[h!]
\begin{center}
\footnotesize
\begin{tabular}{l|c | c c c}
\hline
 & & Accuracy (\%) $\uparrow$ & F1-score $\uparrow$ & AUROC $\uparrow$\\
\cline{3-5} Method & Arch.& \multicolumn{3}{c}{1-shot}\\ \hline
Proto Net~\cite{snell2017prototypical} + DeepSVDD~\cite{pmlr-v80-ruff18a}& Conv64 & - & - & $0.675\pm0.029$ \\
Proto Net~\cite{snell2017prototypical} + DeepAnomaly~\cite{golan2018deep}& Conv64 & - & - & $0.633\pm0.011$\\
Proto Net~\cite{snell2017prototypical} + SVDD~\cite{tax2004support}& Conv64 & $50.00\pm0.00$ & $0.000\pm0.000$ & $0.703\pm0.010$\\
Proto Net~\cite{snell2017prototypical} + OCSVM~\cite{scholkopf2000support}& Conv64 & $50.00\pm0.00$ & $0.000\pm0.000$ & $0.702\pm0.011$\\
Proto Net~\cite{snell2017prototypical} + Threshold& Conv64 & $53.18\pm6.65$ & $0.131\pm0.279$ & $0.706\pm0.012$\\
CLEAR~\cite{kozerawski2018clear} & Conv64 & $50.42\pm1.03$ & $0.077\pm0.102$ & $0.507\pm0.009$\\
Proto Net~\cite{snell2017prototypical} + \MetaBCE~[ours]& Conv64 & $57.57\pm0.78$ & $0.283\pm0.019$ & $\bf{0.764\pm0.016}$\\
Proto Net~\cite{snell2017prototypical} + \OCML~[ours]& Conv64 & $\bf{68.05\pm0.99}$ & $\bf{0.620\pm0.014}$ & $\bf{0.759\pm0.012}$\\
\hline Upper-bound (supervised FEAT~\cite{ye2020fewshot})$^1$& Conv64 & $79.76\pm0.91$ & - & -\\
\hline & & \multicolumn{3}{c}{5-shot}\\ \hline
Proto Net~\cite{snell2017prototypical} + DeepSVDD~\cite{pmlr-v80-ruff18a}& Conv64 & - & - & $0.714\pm0.009$ \\
Proto Net~\cite{snell2017prototypical} + DeepAnomaly~\cite{golan2018deep}& Conv64 & - & - & $0.729\pm0.011$\\
Proto Net~\cite{snell2017prototypical} + SVDD~\cite{tax2004support}& Conv64 & $50.12\pm0.02$ & $0.005\pm0.001$ & $0.747\pm0.016$\\
Proto Net~\cite{snell2017prototypical} + OCSVM~\cite{scholkopf2000support}& Conv64 & $51.22\pm0.01$ & $0.053\pm0.006$ & $0.746\pm0.016$ \\
Proto Net~\cite{snell2017prototypical} + Threshold& Conv64 & $60.93\pm6.09$ & $0.584\pm0.168$ & $0.742\pm0.019$\\
Proto Net~\cite{snell2017prototypical} + \MetaBCE~[ours]& Conv64 & $\bf{77.38\pm0.67}$ & $\bf{0.793\pm0.005}$ & $\bf{0.850\pm0.007}$\\
Proto Net~\cite{snell2017prototypical} + \OCML~[ours]& Conv64 & $74.74\pm0.88$ & $0.693\pm0.013$ & $\bf{0.854\pm0.009}$\\
\hline Upper-bound (supervised FEAT~\cite{ye2020fewshot})$^1$& Conv64 & $88.61\pm0.63$ & - & -\\
\hline
\end{tabular}
\end{center}
\caption{Experimental results on miniImageNet dataset for few-shot one-class classification.  The best results are shown in {\bf bold}. $^1$ Supervised two-class classification.}
\label{table:fsoc_mini_val}

\end{table}


\subsection{Few-Shot Open-Set}

In Table~\ref{table:fsos_mini_val} we provide the experimental results for the few-shot open-set task on the miniImageNet dataset~\cite{vinyals2016matching}. We compare our approaches with Gaussian Embedding (GaussE)~\cite{liu2020few} and PEELER~\cite{liu2020few} by Liu~\etal, where they introduced the few-shot open-set problem. We also compare our method with existing (non-few-shot) open-set state-of-the-art methods such as Open-Max~\cite{bendale2016towards}, Counterfactual~\cite{neal2018open}, Entropic Open-Set Loss~\cite{dhamija2018reducing}, and Objectosphere Loss~\cite{dhamija2018reducing}. Following Liu~\etal~\cite{liu2020few} we perform experiments with ResNet-18 network and in accordance to standard few-shot learning practices we add experiments with Conv64 backbone as well.

\begin{table}[h!]
\begin{center}
\footnotesize
\begin{tabular}{l|c |c c c c}
\hline
& & Accuracy (\%) $\uparrow$ & NA (\%) $\uparrow$ & F1-open $\uparrow$ & AUROC $\uparrow$\\ 
\cline{3-6} Method & Arch. & \multicolumn{4}{c}{1-shot}\\ \hline
GaussE~\cite{liu2020few} + OpenMax~\cite{bendale2016towards}$^1$ & Res18 & $57.89\pm0.59$ & - & - & $ 0.589\pm0.006$\\
GaussE~\cite{liu2020few} + Counterfactual~\cite{neal2018open}$^1$ & Res18 & $57.89\pm0.59$ & - & - & $0.522\pm0.006$\\
GaussE~\cite{liu2020few}$^1$ & Res18 & $57.89\pm0.59$ & - & - & $0.587\pm0.006$\\
PEELER~\cite{liu2020few}$^1$ & Res18 & $58.31\pm0.58$ & - & - & $0.617\pm0.006$ \\
PEELER~\cite{liu2020few} + threshold & Res18 & $56.98\pm0.88$ & $44.77\pm0.70$ & $0.357\pm0.009$ & $0.626\pm0.008$\\
PEELER~\cite{liu2020few} + Entropic Loss~\cite{dhamija2018reducing} & Res18 & $56.10\pm0.86$ & $48.36\pm0.69$ & $0.367\pm0.008$ & $0.631\pm0.008$\\
PEELER~\cite{liu2020few} + Objectosphere~\cite{dhamija2018reducing} & Res18 & $52.63\pm0.86$ & $27.31\pm0.37$ & $0.060\pm0.007$ & $0.548\pm0.008$\\
PEELER~\cite{liu2020few} + \MetaBCE~[ours] & Res18 & $56.98\pm0.88$ & $36.94\pm0.44$ & $0.252\pm0.007$ & $0.606\pm0.008$ \\
PEELER~\cite{liu2020few} + \OCML~[ours] & Res18 & $56.98\pm0.88$ & $57.60\pm0.76$ & $0.380\pm0.006$ & $0.607\pm0.007$ \\
FEAT~\cite{ye2020fewshot} + threshold & Res18 & $\bf{66.42\pm0.65}$ & $56.53\pm0.89$ & $\bf{0.442\pm0.005}$ & $\bf{0.686\pm0.008}$\\
FEAT~\cite{ye2020fewshot} + \MetaBCE~[ours] & Res18 & $\bf{66.42\pm0.65}$ & $34.70\pm0.31$ & $0.232\pm0.006$ & $0.624\pm0.006$\\ 
FEAT~\cite{ye2020fewshot} + \OCML~[ours] & Res18 & $\bf{66.42\pm0.65}$ & $\bf{59.79\pm0.54}$ & $\bf{0.440\pm0.005}$ & $0.623\pm0.006$\\ \hline

PEELER~\cite{liu2020few} + threshold & Conv64 & $51.38\pm0.81$ & $42.48\pm0.75$ & $0.310\pm0.009$ & $0.568\pm0.007$\\
PEELER~\cite{liu2020few} + Entropic Loss~\cite{dhamija2018reducing} & Conv64 & $51.11\pm0.82$ & $41.81\pm0.68$ & $0.299\pm0.008$ & $0.594\pm0.008$\\
PEELER~\cite{liu2020few} + Objectosphere~\cite{dhamija2018reducing} & Conv64 & $49.01\pm0.85$ & $33.79\pm0.68$ & $0.204\pm0.009$ & $0.519\pm0.009$\\
PEELER~\cite{liu2020few} + \MetaBCE~[ours] & Conv64 & $51.38\pm0.81$ & $36.11\pm0.41$ & $0.233\pm0.007$ & $0.600\pm0.008$ \\
PEELER~\cite{liu2020few} + \OCML~[ours] & Conv64 & $51.38\pm0.81$ & $\bf{55.80\pm0.78}$ & $0.349\pm0.006$ & $0.613\pm0.008$ \\
FEAT~\cite{ye2020fewshot} + threshold & Conv64 & $\bf{55.01\pm0.62}$ & $38.80\pm0.40$ & $0.279\pm0.005$ & $0.557\pm0.006$\\
FEAT~\cite{ye2020fewshot} + \MetaBCE~[ours] & Conv64 & $\bf{55.01\pm0.62}$ & $33.64\pm0.30$ & $0.205\pm0.006$ & $0.616\pm0.006$\\
FEAT~\cite{ye2020fewshot} + \OCML~[ours] & Conv64 & $\bf{55.01\pm0.62}$ & $\bf{55.81\pm0.53}$ & $\bf{0.373\pm0.005}$ & $\bf{0.626\pm0.006}$\\

\hline& & \multicolumn{4}{c}{5-shot}\\ \hline
GaussE~\cite{liu2020few} + OpenMax~\cite{bendale2016towards}$^1$ & Res18 & $75.31\pm0.76$ & - & - & $0.675\pm0.007$\\
GaussE~\cite{liu2020few} + Counterfactual~\cite{neal2018open}$^1$ & Res18 & $75.31\pm0.76$ & - & - & $0.533\pm0.006$\\
GaussE~\cite{liu2020few}$^1$ & Res18 & $75.31\pm0.76$ & - & - & $0.665\pm0.007$\\
PEELER~\cite{liu2020few}$^1$ & Res18 & $75.08\pm0.72$ & - & - & $0.699\pm0.007$\\
PEELER~\cite{liu2020few} + threshold & Res18 & $73.04\pm0.68$ & $52.78\pm0.83$ & $0.472\pm0.009$ & $0.677\pm0.008$\\
PEELER~\cite{liu2020few} + Entropic Loss~\cite{dhamija2018reducing} & Res18 & $70.98\pm0.64$ & $59.26\pm0.69$ & $0.491\pm0.006$ & $0.688\pm0.007$\\
PEELER~\cite{liu2020few} + Objectosphere~\cite{dhamija2018reducing} & Res18 & $66.27\pm0.77$ & $26.29\pm0.19$ & $0.034\pm0.005$ & $0.564\pm0.008$\\
PEELER~\cite{liu2020few} + \MetaBCE~[ours] & Res18 & $73.04\pm0.68$ & $70.98\pm0.58$ & $0.495\pm0.004$ & $0.646\pm0.005$ \\
PEELER~\cite{liu2020few} + \OCML~[ours] & Res18 & $73.04\pm0.68$ & $65.64\pm0.49$ & $0.498\pm0.004$ & $0.661\pm0.006$ \\
FEAT~\cite{ye2020fewshot} + threshold & Res18 & $\bf{80.26\pm0.45}$ & $63.89\pm1.02$ & $0.537\pm0.003$ & $\bf{0.727\pm0.007}$\\
FEAT~\cite{ye2020fewshot} + \MetaBCE~[ours] & Res18 & $\bf{80.26\pm0.45}$ & $\bf{73.32\pm0.49}$ & $\bf{0.553\pm0.003}$ & $0.667\pm0.005$\\
FEAT~\cite{ye2020fewshot} + \OCML~[ours] & Res18 & $\bf{80.26\pm0.45}$ & $64.36\pm0.48$ & $0.544\pm0.004$ & $0.682\pm0.006$\\ \hline

PEELER~\cite{liu2020few} + threshold & Conv64 & $66.92\pm0.66$ & $47.24\pm0.83$ & $0.399\pm0.009$ & $0.608\pm0.007$\\
PEELER~\cite{liu2020few} + Entropic Loss~\cite{dhamija2018reducing} & Conv64 & $65.06\pm0.71$ & $47.77\pm0.80$ & $0.386\pm0.009$ & $0.625\pm0.008$\\
PEELER~\cite{liu2020few} + Objectosphere~\cite{dhamija2018reducing} & Conv64 & $62.04\pm0.71$ & $35.10\pm0.73$ & $0.219\pm0.010$ & $0.526\pm0.009$\\
PEELER~\cite{liu2020few} + \MetaBCE~[ours] & Conv64 & $66.92\pm0.66$ & $\bf{68.74\pm0.56}$ & $0.460\pm0.004$ & $0.645\pm0.005$ \\
PEELER~\cite{liu2020few} + \OCML~[ours] & Conv64 & $66.92\pm0.66$ & $63.57\pm0.48$ & $0.466\pm0.004$ & $0.666\pm0.006$ \\
FEAT~\cite{ye2020fewshot} + threshold & Conv64 & $\bf{70.70\pm0.50}$ & $45.85\pm0.46$ & $0.389\pm0.005$ & $0.595\pm0.006$\\
FEAT~\cite{ye2020fewshot} + \MetaBCE~[ours] & Conv64 & $\bf{70.70\pm0.50}$ & $\bf{69.10\pm0.45}$ & $\bf{0.487\pm0.004}$ & $0.669\pm0.005$\\
FEAT~\cite{ye2020fewshot} + \OCML~[ours] & Conv64 & $\bf{70.70\pm0.50}$ & $61.29\pm0.48$ & $\bf{0.490\pm0.004}$ & $\bf{0.680\pm0.006}$\\

\hline
\end{tabular}
\end{center}
\caption{Experimental results on miniImageNet dataset for few-shot $5$-way open-set classification with $5$ open-set categories. The best results are shown in {\bf bold}. $^1$ Results from Liu~\etal~\cite{liu2020few}}
\label{table:fsos_mini_val}

\end{table}

First of all, it is worth noticing that PEELER and Gaussian Embedding (GaussE) introduced by Liu~\etal~\cite{liu2020few} do not provide a method for clearly differentiating between known and unknown examples at inference time and only provide a relative AUROC score and accuracy on closed-set (known) examples. Additionally, their method operates on softmax scores thus making it not suitable for scenario when $n=1$ (few-shot one-class). We have reproduced PEELER approach using the original authors' implementation and augmented it with different existing methods of detecting unknown examples in the query set: threshold method, Entropic Open-Set Loss~\cite{dhamija2018reducing}, and Objectosphere Loss~\cite{dhamija2018reducing}. Out of all methods, Objectosphere Loss~\cite{dhamija2018reducing} performs the worst, obtaining the lowest normalized accuracy, F1-open score and AUROC score both in $1$-shot and $5$-shot settings with ResNet-18 and Conv64 networks, and results in a lower closed-set accuracy compared to other methods. This confirms that methods performing well in standard open-set tasks, do not have to transfer well to few-shot open-set tasks. In this specific scenario, Objectosphere~\cite{dhamija2018reducing} learns to produce features of lower magnitude for unseen classes, however, in the few-shot setting all classes in the meta-testing meta-set $\mathcal{D}_{meta-test}$ will be considered as unseen since there is no overlap in between categories present in the meta-training meta-set and meta-testing meta-set. Entropic Open-Set Loss~\cite{dhamija2018reducing} aims to increase entropy for unknown examples, which seems to translate better to few-shot setting compared to simple threshold or Objectosphere~\cite{dhamija2018reducing} for both settings ($1$-shot and $5$-shot) and both networks (ResNet-18 and Conv64), however it results in a lower closed-set accuracy.

When we augment PEELER~\cite{liu2020few} with an ensemble of our proposed few-shot one-class methods (\MetaBCE~or \OCML) we can significantly increase the open-set performance without influencing the closed-set accuracy. The results confirm the conclusion from the experiments on few-shot one-class task: \MetaBCE~results in a better performance for larger values of $k$ (per-category examples) with normalized accuracy of $70.98\%$ (ResNet-18 and $k=5$) and \OCML~results in a better performance for smaller values of $k$ with a normalized accuracy of $57.60\%$ (ResNet-18 and $k=1$). The great benefit of our approaches is that they do not degrade the performance of closed-set accuracy (compared to Entropic Open-Set Loss~\cite{dhamija2018reducing} and Objectosphere~\cite{dhamija2018reducing}) as they work as separate modules trained on top of existing few-shot learning methods. Additionally, since \OCML~and \MetaBCE~are trained as few-shot one-class methods, they do not require separate background (or unknown) categories present in the training set (contrary to PEELER~\cite{liu2020few}, Entropic Open-Set Loss~\cite{dhamija2018reducing}, Objectosphere~\cite{dhamija2018reducing} or OpenMax~\cite{bendale2016towards}).

Standard multiclass open-set methods such as PEELER, OpenMax, or Entropic Open-Set Loss cannot be adapted to the special case when number of known categories ($n$) drops to one (few-shot one-class setting) since all of them are softmax-based, which prevents them from working in the single-category scenario. Our proposed approaches tackle both problems with superior performance. Additionally, in the few-shot open-set setting, existing meta-learning models due to the training process are tailored to the closed-set setting, thus achieving worse performance on the open-set setting -- as seen in Table~\ref{table:fsos_mini_val} through the comparison of Prototypical Networks~\cite{snell2017prototypical} or FEAT~\cite{ye2020fewshot} when using the multiclass feature space (``+ theshold'') versus when using a one-class feature space (\MetaBCE).

Since both \MetaBCE~and \OCML~can be added to any existing few-shot closed-set approaches to allow them to work in open-set settings, we combine them with state-of-the-art few-shot closed-set method (FEAT) proposed by Ye~\etal~\cite{ye2020fewshot}. FEAT results with \MetaBCE~and \OCML~significantly outperform all the other existing methods in all performance metrics with normalized accuracy of $59.79\%$ with \OCML~for $k=1$ and $73.32\%$ with \MetaBCE~for $k=5$ (with ResNet-18). The results on the CUB-200-2011 dataset~\cite{WahCUB_200_2011} and on tieredImageNet~\cite{ren2018meta} uphold these conclusions, as FEAT + \OCML~achieves the best performance in the $1$-shot setting ($60.21\%$ normalized accuracy on CUB and $67.76\%$ on tieredImageNet) and FEAT + \MetaBCE~performs best in the $5$-shot setting on the CUB dataset ($72.04\%$ normalized accuracy), while FEAT + \OCML~achieves the best performance on the tieredImageNet dataset ($77.96\%$ normalized accuracy). For details on the CUB-200-2011 and tieredImageNet results, please see Appendix~\ref{sec:fsos_additional}.

\section{Conclusions}\label{sec:conclusion}

We proposed two novel methods for few-shot one-class classification (\MetaBCE~and \OCML) that can augment any existing few-shot learning method (such as PEELER~\cite{liu2020few} or FEAT~\cite{ye2020fewshot}) to work in the few-shot open-set setting. These methods do not require retraining of the existing few-shot method, do not degrade its performance in the closed-set setting, and (contrary to existing open-set methods) do not require separate background categories present during the training phase. Our approaches surpass the state-of-the-art methods in few-shot one-class and few-shot multiclass open-set classification, with \MetaBCE~performing better when the number of per-category examples is higher, and \OCML~performing optimally for smaller values of per-category examples. Training high-quality models quickly and efficiently with smaller amounts of data and with the ability to work in an open-set setting is an important future direction for machine learning.

\bibliographystyle{IEEEtran}
\bibliography{main}

\appendix
\section{Performance metrics}\label{sec:fsos_metrics}

{\bf Few-Shot One-Class metrics.} For the few-shot one-class experiments we have used multiple metrics to capture the performance of tested models on the meta-testing meta-set: classification accuracy, AUROC (Area Under ROC curve) score, and F1-score.

\noindent{\bf Few-Shot Open-Set metrics.} To obtain fair comparison with Liu~\etal~\cite{liu2020few} we use the same metrics in the few-shot experiments: closed-set accuracy (dubbed here accuracy) and AUROC (Area Under ROC curve) score. However, these measures alone are not sufficient to capture the quality of open-set methods (impact of unknown examples on the classification performance on known or closed-set samples), and for this reason we will utilize two commonly used metrics in open-set classification introduced by J{\'u}nior~\etal~\cite{junior2017nearest}: F1-open score and Normalized Accuracy. Where F1-open score can be calculated using following formula:

\begin{equation}
\mbox{F1-open} = 2\times\frac{P_{mi}\times R_{mi}}{P_{mi}+R_{mi}}
\label{eq:fsos_f_score}
\end{equation}

\begin{equation}
P_{mi} = \frac{\sum_{i=1}^N TP_i }{\sum_{i=1}^N (TP_i + FP_i)},~~~R_{mi} = \frac{\sum_{i=1}^N TP_i }{\sum_{i=1}^N (TP_i + FN_i)}
\label{eq:fsos_prec_rec}
\end{equation}

\noindent And Normalized Accuracy is:

\begin{equation}
NA = \lambda \times AKS+(1-\lambda)\times AUS
\label{eq:normalized_accuracy}
\end{equation}

\noindent where $\lambda$ is a weight hyperparameter balancing the AKS and AUS (set here to $\lambda=0.5$). The AKS is the Accuracy on Known Samples:
\begin{equation}
AKS = \frac{\sum_{i=1}^N (TP_i + TN_i)}{\sum_{i=1}^N (TP_i + TN_i + FP_i +FN_i)}
\label{eq:aks_junior}
\end{equation}

\noindent and AUS is the Accuracy on Unknown Samples:
\begin{equation}
AUS = \frac{TU}{TU + FU}
\label{eq:aus}
\end{equation}

\noindent where TP is the number of True Positives, TN is the number of True Negatives, FP is the number of False Positives, FN is the number of False Negatives, TU is the number of True Unknowns, and FU is the number of False Unknowns.

The proposed way of measuring AKS (Eq.~\ref{eq:aks_junior}) in a multiclass scenario results in weighing heavily the number of True Negatives (TN) in the formula thus skewing the quality of the metric. We propose to utilize the following way of measuring the AKS, which is more frequently used in a multiclass classification:

\begin{equation}
AKS = \frac{1}{M} \sum_{i=1}^{M} I(\hat{y}_i = y_i)
\label{eq:aks}
\end{equation}

\noindent where $M$ is number of known examples, $y_i$ is the true label of an i-th examples, $\hat{y}_i$ is the predicted label for the i-th example, and $I()$ is the indicator function, which returns 1 if the prediction matches the ground truth and 0 otherwise.

Open-Set Classification Rate Curve (OSCRC) metric introduced by Dhamija \etal~\cite{dhamija2018reducing} assumes that the open-set method does not provide a clear classification decision (whether a novel examples in known or unknown) and scans throug different softmax threshold values in order to produce the curve. Our approaches provide a clear classification decision (a single operating point on the OSCRC curve), thus calculation of OSCRC metric is not possible.
\section{Testing procedure}
\subsection{Few-Shot One-Class}
Every method is tested using the same testing procedure. The testing procedure consists of $M_{episodes}$ testing episodes, each consisting of a support set $\mathcal{S}$ in form of $1$-way $k$-shot setting, a query set $\mathcal{Q}$ with $q$ unseen examples from the category present in the support set and with $q$ examples from another, unknown category. All episodes contain data from the meta-testing meta-set $\mathcal{D}_{meta-test}$. We report the average performance with a $95\%$ confidence interval. In all the experiments we have set $q=15$, $M_{episodes}=10000$.
\subsection{Few-Shot Open-Set}
Every method is tested using the same testing procedure. The testing procedure consists of $M_{episodes}$ testing episodes, each consisting of a support set $\mathcal{S}$ in form of $n$-way $k$-shot setting, a query set $\mathcal{Q}$ with $q$ per-category unseen examples from $N$ categories present in the support set and with $q$ per-category examples from $n_{U}$ unknown categories. All episodes contain data from the meta-testing meta-set $\mathcal{D}_{meta-test}$. We report the average performance with a $95\%$ confidence interval. In all the experiments we have set $q=15$, $M_{episodes}=10000$,  and $n_{U}=N$.

\section{Additional experiments: CUB-200-2011 and tieredImageNet}
\subsection{Few-Shot One-Class}\label{sec:fsoc_additional}

In Tables~\ref{table:fsoc_cub_val} and~\ref{table:fsoc_tier_val} we present the results for few-shot one-class experiments on the CUB-200-2011 dataset~\cite{WahCUB_200_2011} and tieredImageNet dataset~\cite{ren2018meta} respectively. The performance of all methods confirms the conclusions from the miniImageNet experiments. Standard one-class approaches (DeepSVDD~\cite{pmlr-v80-ruff18a}, DeepAnomaly~\cite{golan2018deep}, OCSVM~\cite{scholkopf2000support}, SVDD~\cite{tax2004support}) do not work well in the few-shot setting (especially when considering accuracy and F1-score metrics in $1$-shot setting). Prototypical Networks~\cite{snell2017prototypical} with \OCML~perform the best in the $1$-shot setting with $72.14\%$ accuracy, $0.701$ F1-score, and $0.801$ AUROC score on CUB dataset and with $72.13\%$ accuracy, $0.724$ F1-score, and $0.800$ AUROC score on tieredImageNet dataset. We can see that the supervised upper-bound calculated using FEAT~\cite{ye2020fewshot} still gives a lot of room for improvement with accuracy of $84.25\%$ on CUB and $88.54\%$ on tieredImageNet dataset. In the $5$-shot setting, on CUB dataset \OCML~has the best accuracy ($78.74\%$) and AUROC score ($0.874$), while \MetaBCE~has the best F1-score of $0.783$. On tieredImageNet \OCML~has the best accuracy ($78.89\%$), highest F1-score ($0.803$) and AUROC score ($0.878$). The supervised two-class upper-bound of FEAT~\cite{ye2020fewshot} rises higher as well to $91.26\%$ on CUB and to $94.56\%$ on tieredImageNet. 

\begin{table}[h!]
\begin{center}
\footnotesize
\begin{tabular}{l|c | c c c}
\hline
 & & Accuracy (\%) $\uparrow$ & F1-score $\uparrow$ & AUROC $\uparrow$\\
\cline{3-5} Method & Arch. & \multicolumn{3}{c}{1-shot}\\ \hline
Proto Net~\cite{snell2017prototypical} + DeepSVDD~\cite{pmlr-v80-ruff18a}& Conv64 & - & - & $0.603\pm0.007$\\
Proto Net~\cite{snell2017prototypical} + DeepAnomaly~\cite{golan2018deep}& Conv64 & - & - & $0.655\pm0.011$\\
Proto Net~\cite{snell2017prototypical} + SVDD~\cite{tax2004support}& Conv64 & $50.00\pm0.00$ & $0.000\pm0.000$ & $0.599\pm0.030$\\
Proto Net~\cite{snell2017prototypical} + OCSVM~\cite{scholkopf2000support}& Conv64 & $50.00\pm0.00$ & $0.000\pm0.000$ & $0.633\pm0.015$\\
Proto Net~\cite{snell2017prototypical} + Threshold& Conv64 & $50.05\pm0.04$ & $0.002\pm0.001$ & $0.697\pm0.005$\\
CLEAR~\cite{kozerawski2018clear} & Conv64 & $50.24\pm0.36$ & $0.008\pm0.012$ & $0.620\pm0.120$\\
Proto Net~\cite{snell2017prototypical} + \MetaBCE~[ours]& Conv64 & $54.72\pm1.29$ & $0.169\pm0.044$ & $0.774\pm0.009$ \\
Proto Net~\cite{snell2017prototypical} + \OCML~[ours]& Conv64 & $\bf{72.14\pm0.56}$ & $\bf{0.701\pm0.016}$ & $\bf{0.801\pm0.006}$ \\
\hline Upper-bound (supervised FEAT~\cite{ye2020fewshot})$^1$& Conv64 & $84.25\pm0.89$ &  - & -\\
\hline & & \multicolumn{3}{c}{5-shot}\\ \hline
Proto Net~\cite{snell2017prototypical} + DeepSVDD~\cite{pmlr-v80-ruff18a}& Conv64 & - & - & $0.656\pm0.007$ \\
Proto Net~\cite{snell2017prototypical} + DeepAnomaly~\cite{golan2018deep}& Conv64 & - & - & $0.762\pm0.010$\\
Proto Net~\cite{snell2017prototypical} + SVDD~\cite{tax2004support}& Conv64 & $50.61\pm0.41$ & $0.036\pm0.021$ & $0.609\pm0.036$ \\
Proto Net~\cite{snell2017prototypical} + OCSVM~\cite{scholkopf2000support}& Conv64 & $52.40\pm0.45$ & $0.130\pm0.019$ & $0.656\pm0.019$ \\
Proto Net~\cite{snell2017prototypical} + Threshold& Conv64 & $64.92\pm1.29$ & $0.590\pm0.063$ & $0.725\pm0.003$ \\
Proto Net~\cite{snell2017prototypical} + \MetaBCE~[ours]& Conv64 & $76.16\pm0.58$ & $\bf{0.783\pm0.007}$ & $0.839\pm0.006$ \\
Proto Net~\cite{snell2017prototypical} + \OCML~[ours]& Conv64 & $\bf{78.74\pm1.07}$ & $0.774\pm0.019$ & $\bf{0.874\pm0.010}$\\
\hline Upper-bound (supervised FEAT~\cite{ye2020fewshot})$^1$& Conv64 & $91.26\pm0.56$ & - & -\\
\hline
\end{tabular}
\end{center}
\caption{Experimental results on CUB-200-2011 dataset for few-shot one-class classification. The best results are shown in {\bf bold}. $^1$ Supervised two-class classification.}
\label{table:fsoc_cub_val}
\end{table}

\begin{table}[h!]
\begin{center}
\footnotesize
\begin{tabular}{l|c | c c c}
\hline
 & & Accuracy (\%) $\uparrow$ & F1-score $\uparrow$ & AUROC $\uparrow$\\
\cline{3-5} Method & Arch. & \multicolumn{3}{c}{1-shot}\\ \hline
Proto Net~\cite{snell2017prototypical} + DeepSVDD~\cite{pmlr-v80-ruff18a}& Conv64 & - & - & $0.656\pm0.004$\\
Proto Net~\cite{snell2017prototypical} + DeepAnomaly~\cite{golan2018deep}& Conv64 & - & - & $0.603\pm0.012$\\
Proto Net~\cite{snell2017prototypical} + SVDD~\cite{tax2004support}& Conv64 & $50.00\pm0.00$ & $0.000\pm0.000$ & $0.669\pm0.003$\\
Proto Net~\cite{snell2017prototypical} + OCSVM~\cite{scholkopf2000support}& Conv64 & $50.00\pm0.00$ & $0.000\pm0.000$ & $0.676\pm0.003$\\
Proto Net~\cite{snell2017prototypical} + Threshold& Conv64 & $50.48\pm0.11$ & $0.017\pm0.003$ & $0.704\pm0.008$\\
CLEAR~\cite{kozerawski2018clear} & Conv64  & $50.53\pm0.53$ & $0.552\pm0.006$ & $0.509\pm0.008$ \\
Proto Net~\cite{snell2017prototypical} + \MetaBCE~[ours]& Conv64 & $55.87\pm0.38$ & $0.208\pm0.010$ & $0.755\pm0.009$\\
Proto Net~\cite{snell2017prototypical} + \OCML~[ours]& Conv64 & $\bf{72.13\pm0.33}$ & $\bf{0.724\pm0.003}$ & $\bf{0.800\pm0.004}$\\
\hline Upper-bound (supervised FEAT~\cite{ye2020fewshot})$^1$& ResNet12 & $88.54\pm0.81$ &  - & -\\
\hline & & \multicolumn{3}{c}{5-shot}\\ \hline
Proto Net~\cite{snell2017prototypical} + DeepSVDD~\cite{pmlr-v80-ruff18a}& Conv64 & - & - & $0.678\pm0.003$ \\
Proto Net~\cite{snell2017prototypical} + DeepAnomaly~\cite{golan2018deep}& Conv64 & - & - & $0.719\pm0.011$\\
Proto Net~\cite{snell2017prototypical} + SVDD~\cite{tax2004support}& Conv64 & $50.52\pm0.04$ & $0.021\pm0.001$ & $0.724\pm0.003$ \\
Proto Net~\cite{snell2017prototypical} + OCSVM~\cite{scholkopf2000support}& Conv64 & $52.37\pm0.09$ & $0.106\pm0.003$ & $0.712\pm0.003$ \\
Proto Net~\cite{snell2017prototypical} + Threshold& Conv64 & $61.90\pm0.54$ & $0.426\pm0.011$ & $0.741\pm0.008$\\
Proto Net~\cite{snell2017prototypical} + \MetaBCE~[ours]& Conv64 & $75.62\pm0.65$ & $0.774\pm0.005$ & $0.831\pm0.007$\\
Proto Net~\cite{snell2017prototypical} + \OCML~[ours]& Conv64 & $\bf{78.89\pm0.67}$ & $\bf{0.803\pm0.006}$ & $\bf{0.878\pm0.007}$\\
\hline Upper-bound (supervised FEAT~\cite{ye2020fewshot})$^1$& ResNet12 & $94.56\pm0.51$ & - & -\\
\hline
\end{tabular}
\end{center}
\caption{Experimental results on tieredImageNet dataset for few-shot one-class classification. The best results are shown in {\bf bold}. $^1$ Supervised two-class classification.}
\label{table:fsoc_tier_val}
\end{table}

\subsection{Few-Shot Open-Set}\label{sec:fsos_additional}

In Tables~\ref{table:fsos_cub_val} and~\ref{table:fsos_tier_val} we can see the performance comparison of few-shot open-set methods on CUB-200-2011 dataset~\cite{WahCUB_200_2011} and tieredImageNet~\cite{ren2018meta}. We can see that both \MetaBCE~and \OCML~do not degrade the closed-set accuracy (dubbed accuracy here), hence methods initially performing well in this setting (\eg~FEAT~\cite{ye2020fewshot}) keep their superior performance. On CUB dataset \OCML~performs the best in the $1$-shot setting with normalized accuracy of $60.21\%$ and F1-open score of $0.414$ (with FEAT~\cite{ye2020fewshot}) and AUROC score of $0.654$ with Prototypical Networks~\cite{snell2017prototypical}. In the $5$-shot setting the best performing method is \MetaBCE~with normalized accuracy of $72.04\%$ and F1-open score of $0.541$. Similarly to $1$-shot setting, Prototypical Networks~\cite{snell2017prototypical} with \OCML~has the best AUROC score of $0.736$. On tieredImageNet dataset \OCML~performs the best in the $1$-shot setting with normalized accuracy of $67.76\%$ and F1-open score of $0.478$ (with FEAT~\cite{ye2020fewshot}). In the $5$-shot setting the best performing method is \OCML~with normalized accuracy of $77.96\%$ and F1-open score of $0.577$.

\begin{table}[h!]
\begin{center}
\footnotesize
\begin{tabular}{l|c |c c c c}
\hline
& & Accuracy (\%) $\uparrow$ & NA (\%) $\uparrow$ & F1-open $\uparrow$ & AUROC $\uparrow$\\ 
\cline{3-6} Method & Arch. & \multicolumn{4}{c}{1-shot}\\ \hline
PEELER~\cite{liu2020few} & Conv64 & $52.56\pm0.73$ & - & - & $0.549\pm0.003$\\
PEELER~\cite{liu2020few} + threshold & Conv64 & $52.56\pm0.73$ & $26.51\pm1.48$ & $0.042\pm0.036$ & $0.605\pm0.007$\\
Proto Nets~\cite{snell2017prototypical} + threshold & Conv64 & $41.72\pm1.03$ & $25.03\pm0.02$ & $0.001\pm0.001$ & $0.566\pm0.009$\\
Proto Nets~\cite{snell2017prototypical} + \MetaBCE~[ours] & Conv64 & $41.72\pm1.03$ & $35.66\pm4.85$ & $0.229\pm0.060$  & $0.613\pm0.019$\\
Proto Nets~\cite{snell2017prototypical} + \OCML~[ours] & Conv64 & $41.72\pm1.03$ & $48.61\pm5.72$ & $0.371\pm0.035$ & $\bf{0.654\pm0.010}$\\
FEAT~\cite{ye2020fewshot} + threshold & Conv64 & $\bf{63.16\pm0.74}$ & $32.69\pm0.29$ & $0.199\pm0.005$ & $0.537\pm0.007$\\
FEAT~\cite{ye2020fewshot} + \MetaBCE~[ours] & Conv64 & $\bf{63.16\pm0.74}$ & $32.70\pm0.26$ & $0.190\pm0.005$ & $0.648\pm0.006$\\
FEAT~\cite{ye2020fewshot} + \OCML~[ours] & Conv64 & $\bf{63.16\pm0.74}$ & $\bf{60.21\pm0.80}$ & $\bf{0.414\pm0.005}$ & $0.556\pm0.005$\\

\hline& & \multicolumn{4}{c}{5-shot}\\ \hline
Proto Nets~\cite{snell2017prototypical} + OpenMax~\cite{bendale2016towards} & Conv64 & $70.97\pm0.99$ & $25.60\pm0.15$ & $0.02\pm0.005$ & $0.574\pm0.011$\\
PEELER~\cite{liu2020few} & Conv64 & $71.65\pm0.79$ & - & - & $0.637\pm0.004$ \\
PEELER~\cite{liu2020few} + threshold & Conv64 & $71.65\pm0.79$ & $42.79\pm6.17$ & $0.348\pm0.062$ & $0.665\pm0.010$\\
Proto Nets~\cite{snell2017prototypical} + threshold & Conv64 & $70.97\pm0.99$ & $37.62\pm1.56$ & $0.298\pm0.022$ & $0.606\pm0.009$\\
Proto Nets~\cite{snell2017prototypical} + \MetaBCE~[ours] & Conv64 & $70.97\pm0.99$ & $68.59\pm2.64$ & $0.509\pm0.009$   & $0.661\pm0.031$\\
Proto Nets~\cite{snell2017prototypical} + \OCML~[ours] & Conv64 & $70.97\pm0.99$ & $47.70\pm9.15$ & $0.426\pm0.103$ & $\bf{0.736\pm0.007}$\\
FEAT~\cite{ye2020fewshot} + threshold & Conv64 & $\bf{77.47\pm0.55}$ & $35.31\pm0.34$ & $0.261\pm0.005$ & $0.568\pm0.007$\\
FEAT~\cite{ye2020fewshot} + \MetaBCE~[ours] & Conv64 & $\bf{77.47\pm0.55}$ & $\bf{72.04\pm0.43}$ & $\bf{0.541\pm0.004}$ & $0.713\pm0.005$\\
FEAT~\cite{ye2020fewshot} + \OCML~[ours] & Conv64 & $\bf{77.47\pm0.55}$ & $66.95\pm0.60$ & $0.513\pm0.004$ & $0.573\pm0.005$\\
\hline
\end{tabular}

\end{center}
\caption{Experimental results on CUB-200-2011 dataset for few-shot $5$-way open-set classification with $5$ open-set categories. The best results are shown in {\bf bold}. }
\label{table:fsos_cub_val}
\end{table}

\begin{table}[h!]
\begin{center}
\footnotesize
\begin{tabular}{l|c |c c c c}
\hline
& & Accuracy (\%) $\uparrow$ & NA (\%) $\uparrow$ & F1-open $\uparrow$ & AUROC $\uparrow$\\ 
\cline{3-6} Method & Arch. & \multicolumn{4}{c}{1-shot}\\ \hline
PEELER~\cite{liu2020few} & Conv64 & $38.55\pm0.43$ & - & - & $0.546\pm0.004$\\
Proto Nets~\cite{snell2017prototypical} + threshold & Conv64 & $43.77\pm0.47$ & $25.38\pm0.05$ & $0.011\pm0.001$ & $0.576\pm0.004$\\
Proto Nets~\cite{snell2017prototypical} + \MetaBCE~[ours] & Conv64 & $43.77\pm0.47$ & $33.74\pm0.21$ & $0.339\pm0.004$ & $0.623\pm0.004$\\
Proto Nets~\cite{snell2017prototypical} + \OCML~[ours] & Conv64 & $43.77\pm0.47$ & $56.06\pm0.47$ & $0.299\pm0.003$ & $0.659\pm0.004$\\
FEAT~\cite{ye2020fewshot} + threshold & ResNet12 & $\bf{70.88\pm0.72}$ & $50.97\pm0.06$ & $0.463\pm0.007$ & $\bf{0.697\pm0.006}$\\
FEAT~\cite{ye2020fewshot} + \MetaBCE~[ours] & ResNet12 & $\bf{70.88\pm0.72}$ & $39.26\pm0.36$ & $0.308\pm0.006$ & $0.592\pm0.006$\\
FEAT~\cite{ye2020fewshot} + \OCML~[ours] & ResNet12 & $\bf{70.88\pm0.72}$ & $\bf{67.76\pm0.86}$ & $\bf{0.478\pm0.005}$ & $0.610\pm0.006$\\

\hline& & \multicolumn{4}{c}{5-shot}\\ \hline
Proto Nets~\cite{snell2017prototypical} + OpenMax~\cite{bendale2016towards} & Conv64 & $71.04\pm0.40$ & $25.73\pm0.04$ & $0.024\pm0.001$ & $0.585\pm0.003$\\
PEELER~\cite{liu2020few} & Conv64 & $67.71\pm0.42$ & - & - & $0.635\pm0.004$\\
Proto Nets~\cite{snell2017prototypical} + threshold & Conv64 & $71.04\pm0.40$ & $38.08\pm0.25$ & $0.302\pm0.004$ & $0.615\pm0.004$\\
Proto Nets~\cite{snell2017prototypical} + \MetaBCE~[ours] & Conv64 & $71.04\pm0.40$ & $70.46\pm0.37$ & $0.495\pm0.003$ & $0.672\pm0.004$\\
Proto Nets~\cite{snell2017prototypical} + \OCML~[ours] & Conv64 & $71.04\pm0.40$ & $72.87\pm0.36$ & $0.509\pm0.003$ & $0.727\pm0.004$\\
FEAT~\cite{ye2020fewshot} + threshold & ResNet12 & $\bf{84.22\pm0.53}$ &  $67.91\pm0.05$ & $0.545\pm0.005$ & $\bf{0.768\pm0.005}$\\
FEAT~\cite{ye2020fewshot} + \MetaBCE~[ours] & ResNet12 & $\bf{84.22\pm0.53}$ & $76.71\pm0.73$ & $0.570\pm0.004$ & $0.636\pm0.005$\\
FEAT~\cite{ye2020fewshot} + \OCML~[ours] & ResNet12 & $\bf{84.22\pm0.53}$ & $\bf{77.96\pm0.63}$ & $\bf{0.577\pm0.003}$ & $0.657\pm0.006$\\
\hline
\end{tabular}

\end{center}
\caption{Experimental results on tieredImageNet dataset for few-shot $5$-way open-set classification with $5$ open-set categories. The best results are shown in {\bf bold}. }
\label{table:fsos_tier_val}
\end{table}
\section{Implementation details}

We have utilized PyTorch~\cite{NEURIPS2019_9015} to train all models. In order to implement results with PEELER~\cite{liu2020few} and FEAT~\cite{ye2020fewshot} we have used the original implementations supplied by the authors of PEELER\footnote{https://github.com/BoLiu-SVCL/meta-open} and FEAT\footnote{https://github.com/Sha-Lab/FEAT}. We have used Prototypical Networks implementation introduced from the authors of FEAT as well. \MetaBCE~uses a separate branch of the main feature extractor ($f_{\phi}'$) which in case of Conv64 network is the last convnet block (2 last convolutional layers). \OCML~uses a transfer learning module $g_{\theta}$ which in case of a Conv64 architecture is a single fully connected linear layer with input and output dimension of $1600$. As both our methods augment existing few-shot closed-set approaches, all training hyperparameters are consistent with original hyperparameters introduced by the authors of the closed-set approaches (PEELER, FEAT or Prototypical Networks). We will release the code necessary to reproduce the results from this paper along all pre-trained models upon the publication.
\section{Ablation studies}
\subsection{\OCML~$g_{\theta}$ architecture}

\begin{table}[h!]
\begin{center}
\begin{tabular}{|l|l|l|}
\hline
$g_{\theta}$ architecture name & Layer 1 & Layer 2\\ \hline
1 layer & FC layer (1600, 1600) & -\\
2 layers, middle dim 100 & FC layer (1600, 100) & FC layer (100, 1600)\\
2 layers, middle dim 500 & FC layer (1600, 500) & FC layer (500, 1600)\\
2 layers, middle dim 1000 & FC layer (1600, 1000) & FC layer (1000, 1600)\\
\hline
\end{tabular}
\end{center}
\caption{Tested architectures for the transfer learning module $g_{\theta}$ for \OCML~approach.}
\label{table:ocml_architecture_table}
\end{table}

\begin{figure}[h!]
\begin{center}
\begin{tabular}{cc}
\includegraphics[width=8cm]{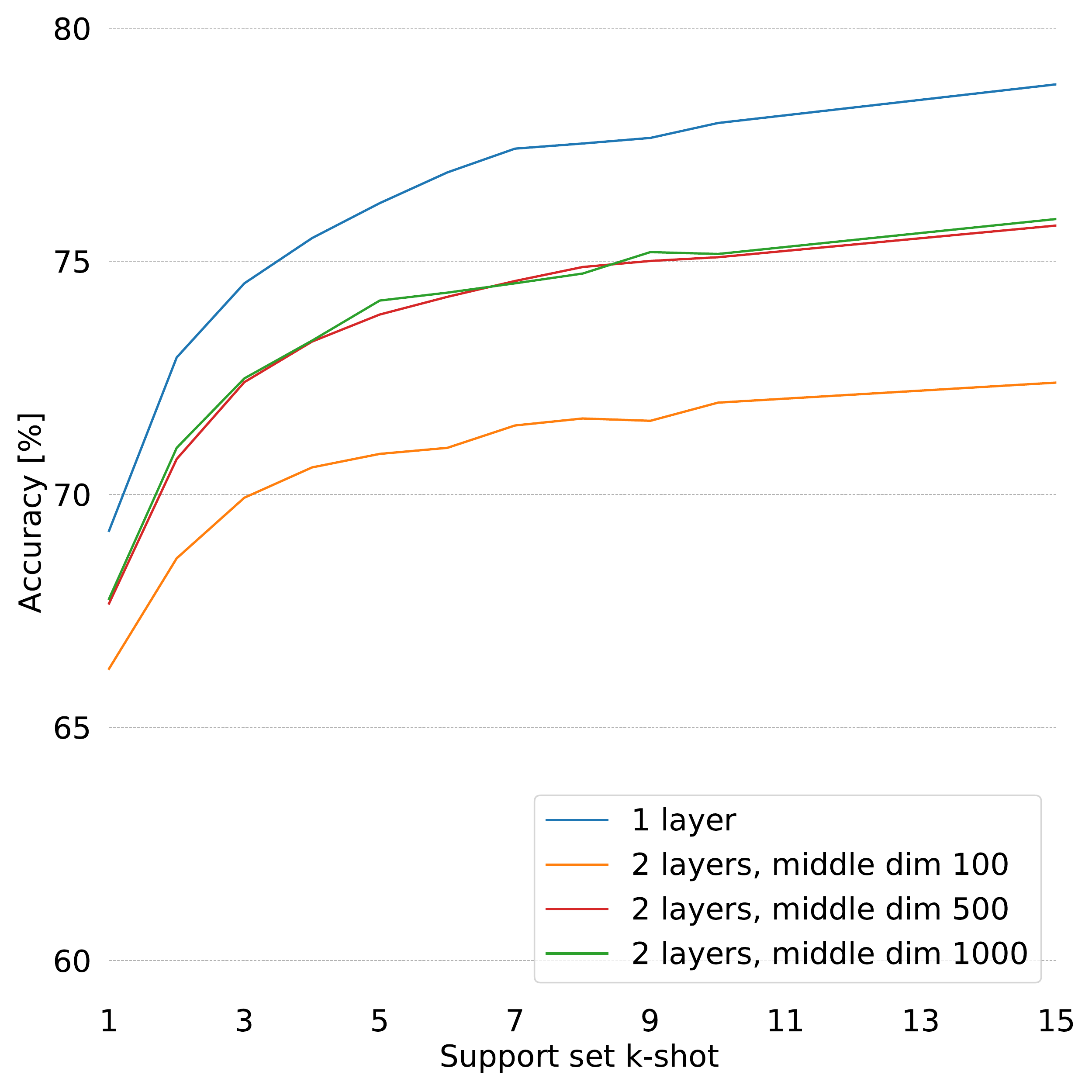}&
\includegraphics[width=8cm]{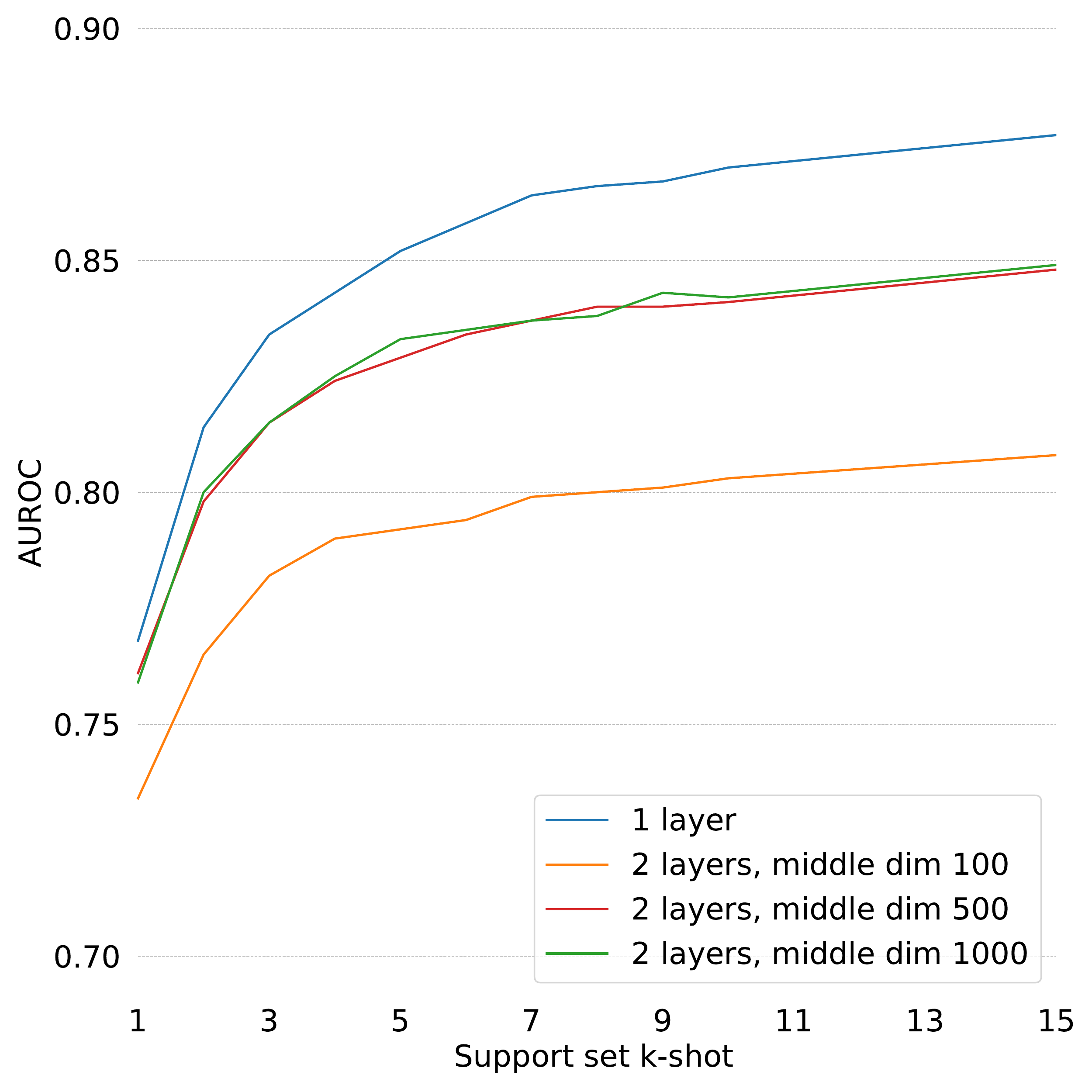}\\
(a) One-Class Accuracy & (b) AUROC \\ 
\end{tabular}
\caption{Ablation study results for different architectures of the transfer learning module $g_{\theta}$ for \OCML~approach. Results for four different architectures are presented for different $k$-shot scenarios (different sizes of the support set $\mathcal{S}$ of the meta-testing meta-set $\mathcal{D}_{meta-test}$).}
\label{fig:ocml_architecture_ablation}
\end{center}
\end{figure}

\OCML~method utilizes a transfer learning module $g_{\theta}$ transforming a class-level representation $\bar{x}_c$ into the weights $w_c$ of a classifier for a category $c$. We performed an ablation study comparing the impact of the architecture of the transfer learning module $g_{\theta}$ on the performance of the network. In Table~\ref{table:ocml_architecture_table} we can see four architectures of the transfer learning module $g_{\theta}$ being tested. All four architectures consist of only fully connected layers (FC) with input, output dimensions in the parenthesis. The ablation study is performed on miniImageNet dataset with the Conv64 network, which has a feature space dimensionality of $1600$, hence the basic architecture for $g_{\theta}$ has only a single FC layer keeping the dimensionality the same. With three additional architectures we have tested whether adding more layers to the transfer learning module is beneficial and what is the impact of the latent space dimensionality (middle dim in Table~\ref{table:ocml_architecture_table}). All experiments were repeated five times and the average performance along the $95\%$ confidence interval was reported.

The results of the experiments for all four settings present in Table~\ref{table:ocml_architecture_table} are visible in Figure~\ref{fig:ocml_architecture_ablation}. On the left we can see the impact of the architecture on the accuracy of the method for various sizes of the support set category in the meta-testing meta-set (various $k$ in the $k$-shot scenario). On the right of Figure~\ref{fig:ocml_architecture_ablation} we can see the impact of the architecture on the AUROC score. The single-layer architecture achieves the best performance (both accuracy and AUROC score) across all $k$ in the $k$-shot setting. Among the two-layer architectures, the one with the latent dimensionality of $100$ achieves the lowest performance across all $k$, which might indicate that the low-dimensional space is not enough to reliably compress the necessary information. Two-layer architectures with a higher latent dimensionality (500 and 1000) have similar performance, although lower than the single-layer architecture. The performance of all the methods increases with with number of examples ($k$) in the support set $\mathcal{S}$ of the meta-testing meta-set $\mathcal{D}_{meta-test}$.

\subsection{\MetaBCE~separate branch}

\begin{table}[h!]
\begin{center}
\footnotesize
\begin{tabular}{l|c |c c c c}
\hline
& & Accuracy (\%) $\uparrow$ & NA (\%) $\uparrow$ & F1-open $\uparrow$ & AUROC $\uparrow$\\ 
\cline{3-6} Method & Arch. & \multicolumn{4}{c}{1-shot}\\ \hline
Proto Nets~\cite{snell2017prototypical} + \MetaBCE~[ours] & Conv64 & $41.72\pm1.03$ & $35.66\pm4.85$ & $0.229\pm0.060$  & $0.613\pm0.019$\\
Proto Nets~\cite{snell2017prototypical} + \MetaBCE$_C$~[ours] & Conv64 & $41.72\pm1.03$ & $28.66\pm3.78$ & $0.080\pm0.077$  & $0.629\pm0.009$\\

\hline& & \multicolumn{4}{c}{5-shot}\\ \hline
Proto Nets~\cite{snell2017prototypical} + \MetaBCE~[ours] & Conv64 & $70.97\pm0.99$ & $68.59\pm2.64$ & $0.509\pm0.009$   & $0.661\pm0.031$\\
Proto Nets~\cite{snell2017prototypical} + \MetaBCE$_C$~[ours] & Conv64 & $70.97\pm0.99$ & $37.13\pm1.53$ & $0.233\pm0.162$  & $0.684\pm0.011$\\
\hline
\end{tabular}

\end{center}
\caption{Experimental results on CUB-200-2011 dataset for few-shot $5$-way open-set classification with $5$ open-set categories.}
\label{table:fsos_mbce_ablation}
\end{table}

The main method of \MetaBCE~ is using an auxiliary branch $f_{\phi}'$ to produce features for the one-class classifier. We can also modify \MetaBCE~to use the main branch of the feature extractor $f_{\phi}$ to calculate the one-class feature vectors. In order to do this, we will use a separate module $h_{\chi}$ to transform the multiclass classification feature vector $f_{\phi}(x)$ to a \MetaBCE~one-class classification feature vector $h_{\chi}(f_{\phi}(x))$ and use it for one-class predictions:

\begin{equation}
\begin{split}
p_(y=c|x) &= \frac{1}{1+exp( -(-d((h_{\chi}(f_{\phi}(x)), \bar{x}_c) - t)) }\\
&= \frac{1}{1+exp( d(h_{\chi}(f_{\phi}(x)), \bar{x}_c) + t) }
\end{split}
\label{eq:BCE2_probability}
\end{equation}

\noindent This version of the \MetaBCE~will be called \MetaBCE$_C$. We can see the accuracy comparison of the above two approaches on the CUB-200-2011 dataset~\cite{WahCUB_200_2011} in Table~\ref{table:fsos_mbce_ablation}. Both methods do not degrade the closed-set accuracy, however the results for \MetaBCE~using a separate branch to calculate the embeddings ($f_{\phi}'$) are significantly better than when using the main branch (\MetaBCE$_C$). This pertains to normalized accuracy and F1-open scores, where \MetaBCE~achieves $35.66\%$ normalized accuracy in $1$-shot setting (compared to $28.66\%$ by \MetaBCE$_C$) and $68.59\%$ in $5$-shot setting (compared to $37.13\%$ by \MetaBCE$_C$). However, the AUROC scores are slightly better when using the main branch (\MetaBCE$_C$). This might indicate that multiclass embeddings obtained in the main branch of the feature extractor help slightly with the overall ranking of known vs unknown examples (as indicated by AUROC score), but a separate feature extraction branch leading to a separate, dedicated one-class embedding allows to obtain better separability of the feature space leading to better classification metrics (normalized accuracy and F1-open score).
\subsection{Impact of $k$-shot test setting}

\begin{figure}[h!]
\begin{center}
\begin{tabular}{cc}
\includegraphics[width=7cm]{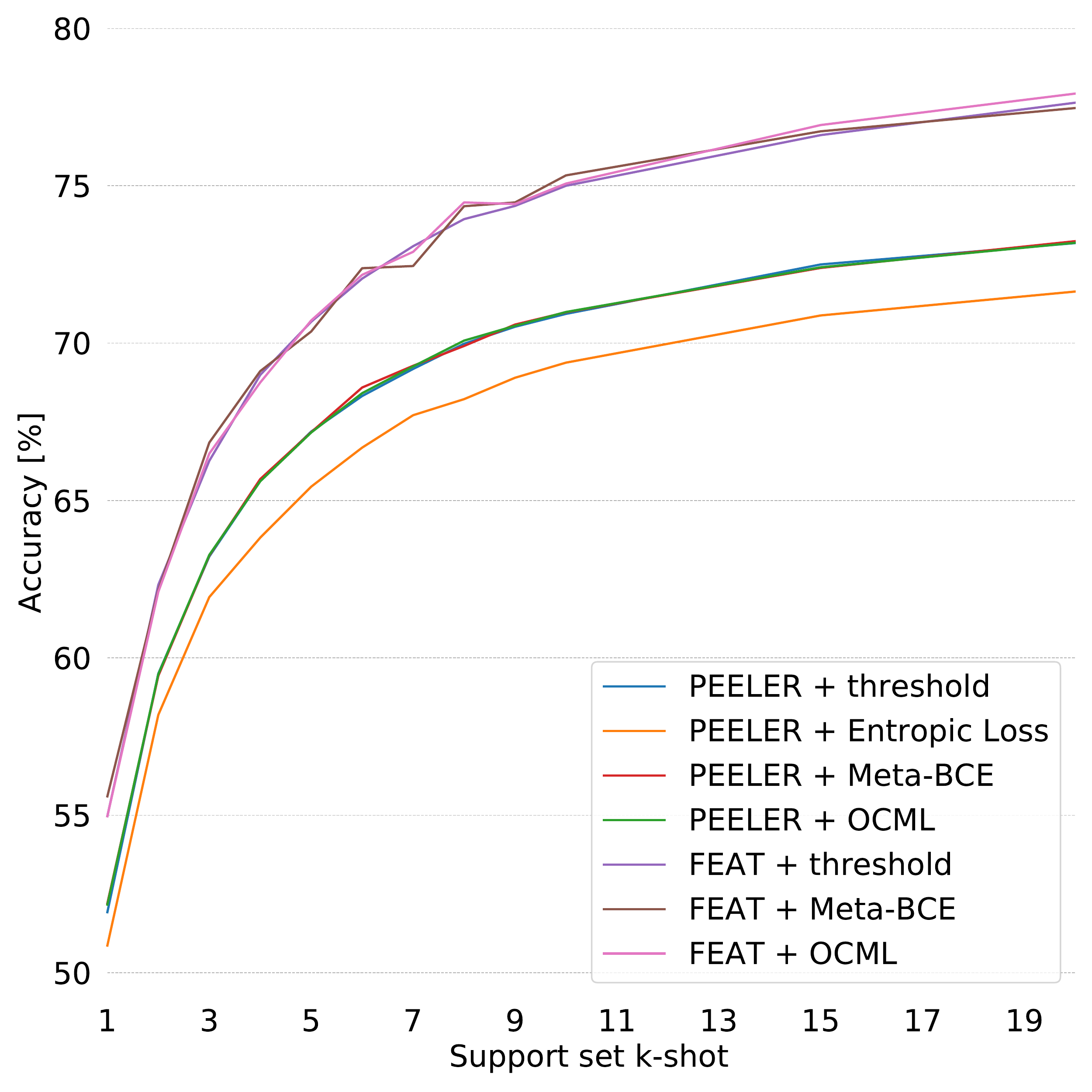}&
\includegraphics[width=7cm]{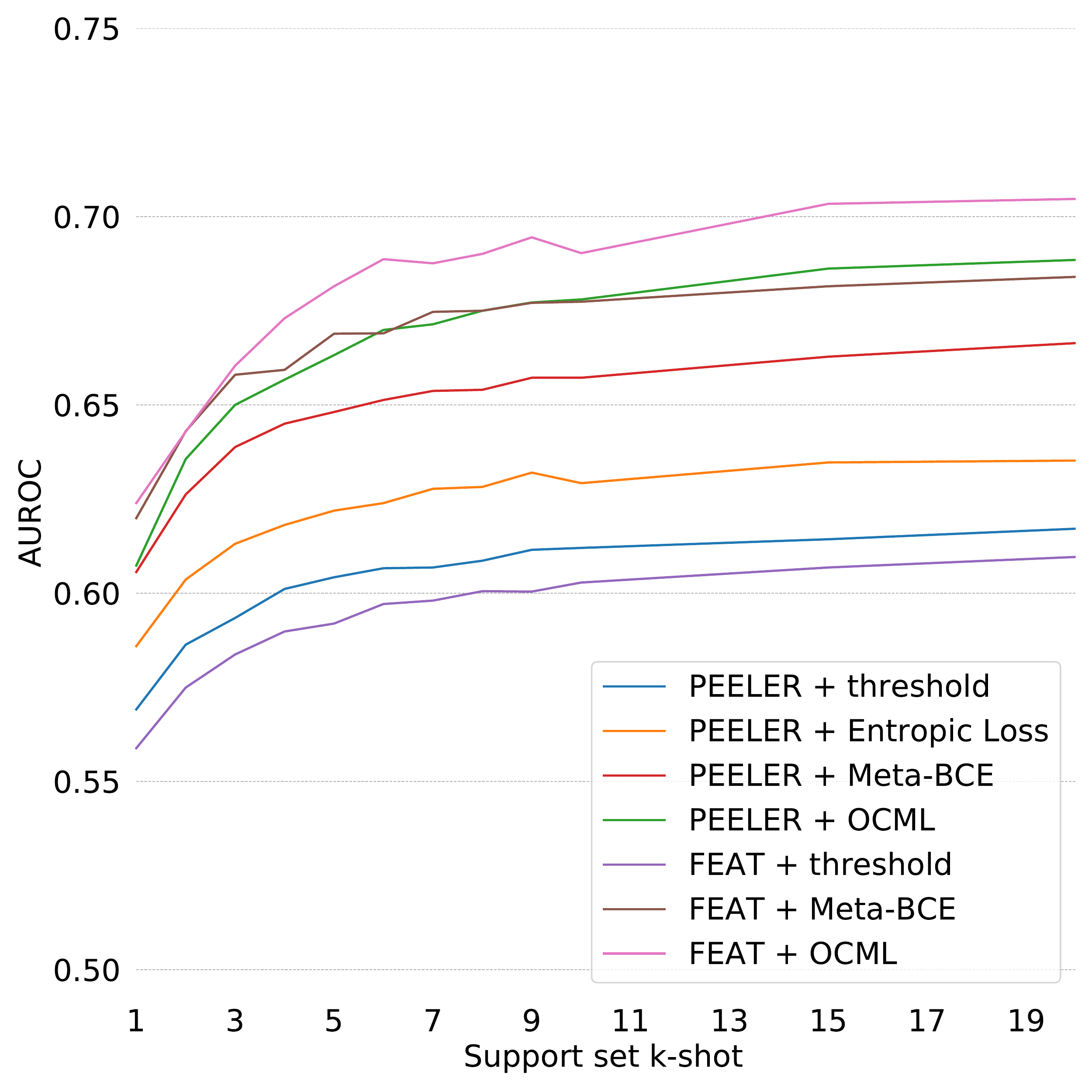}\\
(a) Closed-Set Accuracy & (b) AUROC \\ 
\includegraphics[width=7cm]{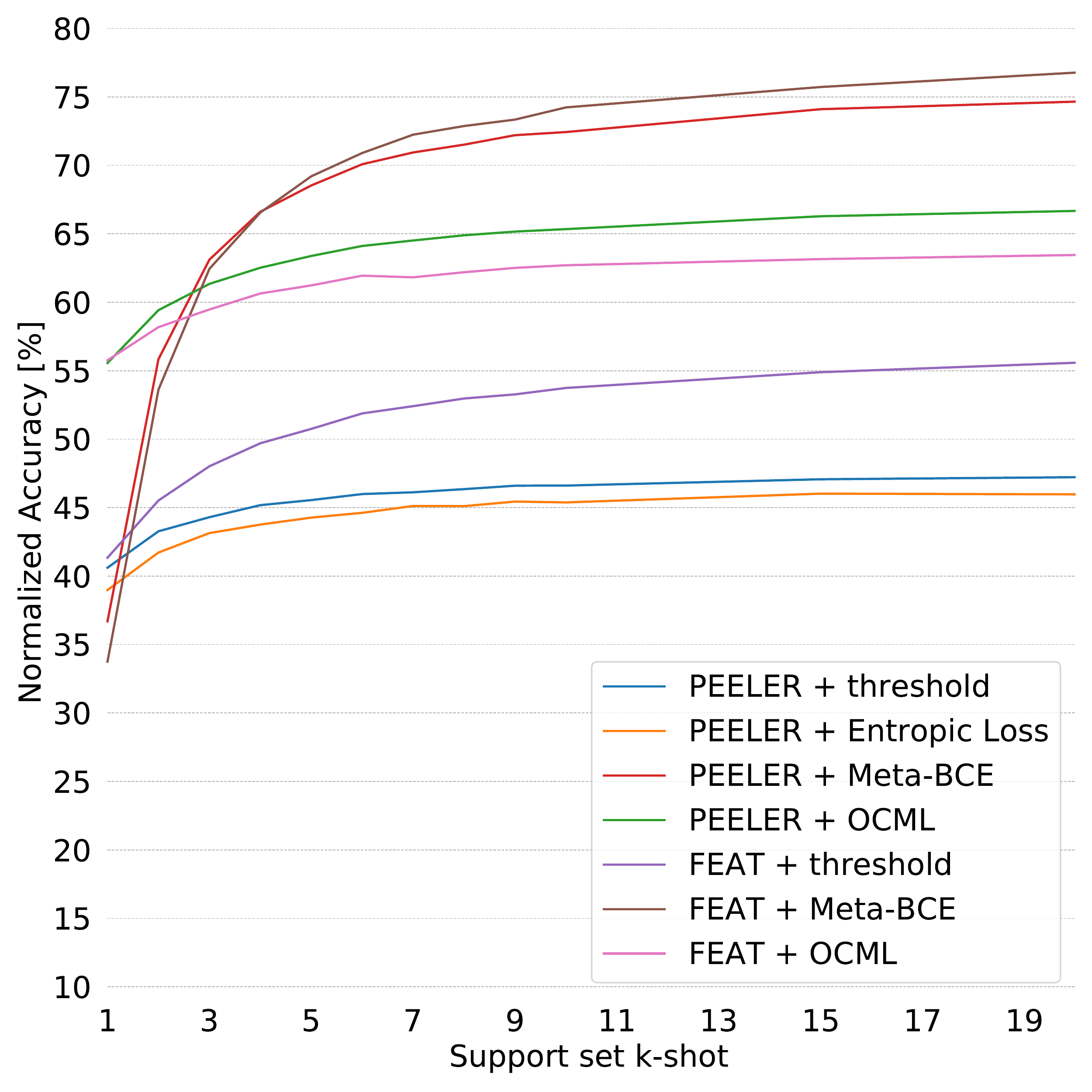}&
\includegraphics[width=7cm]{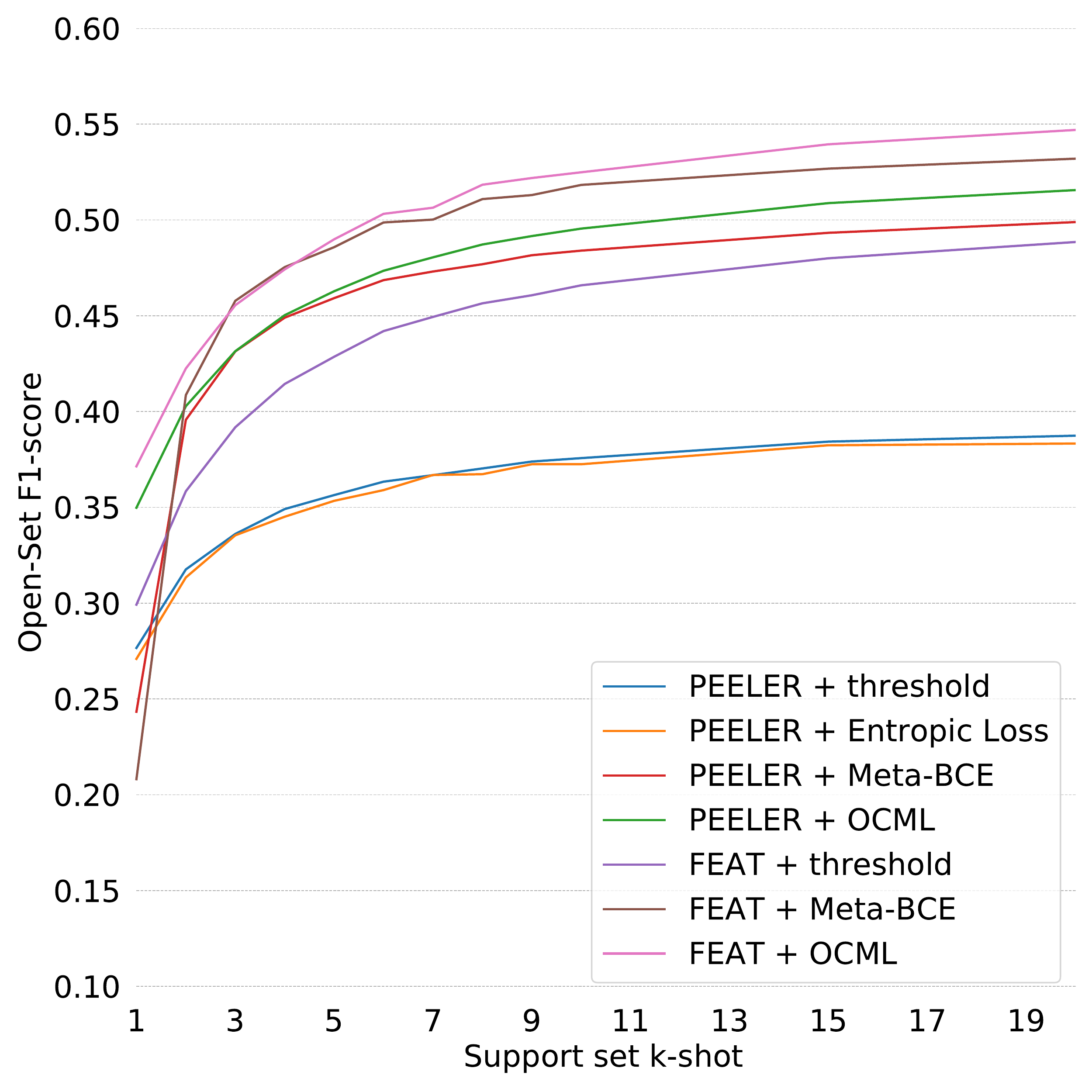}\\
(d) Normalized Accuracy & (e) F1-open score \\ 
\end{tabular}
\caption{Impact of $k$ (number of per-category examples) on the performance for $5$-way $k$-shot classification with $5$ open-set categories.}
\label{fig:fsos_5_way_ablation}
\end{center}
\end{figure}

Number of per-category examples ($k$) has a big effect on the performance of every few-shot approach. In Figure~\ref{fig:fsos_5_way_ablation} we provide a comparison between studied approaches where we showcase impact of number of per-category examples on four performance metrics (closed-set accuracy, AUROC score, normalized accuracy, and F1-open score) for a $5$-way $k$-shot classification with $n_U = 5$ open-set (unknown) categories. The results are from experiments on the miniImageNet dataset~\cite{vinyals2016matching}.

In Figure~\ref{fig:fsos_5_way_ablation}(a) we can see the comparison between methods and their impact on closed-set accuracy. Methods based on PEELER~\cite{liu2020few} have overall lower accuracy for all values of $k$ than those based on FEAT~\cite{ye2020fewshot}. Additionally, it is clear that both our methods (\MetaBCE~and \OCML) do not degrade closed-set accuracy, compared to Entropic Open-Set Loss~\cite{dhamija2018reducing}. In Figure~\ref{fig:fsos_5_way_ablation}(b) we can see the impact on AUROC score. FEAT~\cite{ye2020fewshot} + \OCML~has the best performance for all values of $k$, closely followed by PEELER~\cite{liu2020few} + \OCML~and FEAT~\cite{ye2020fewshot} + \MetaBCE. Threshold-based methods and Entropic Open-Set Loss~\cite{dhamija2018reducing} have the lowest AUROC score across all values of $k$. In Figure~\ref{fig:fsos_5_way_ablation}(c) we can see the normalized accuracy values for the tested methods. Important to see are the high values for \OCML~methods (with PEELER~\cite{liu2020few} and FEAT~\cite{ye2020fewshot}) when $k=1$ and a very high gain of both \MetaBCE~methods (with PEELER~\cite{liu2020few} and FEAT~\cite{ye2020fewshot}) when increasing values of $k$ (surpassing \OCML~performance when $k\geq3$). Very similar behavior can be observed in Figure~\ref{fig:fsos_5_way_ablation}(d) for F1-open score.

\subsection{Impact of $N$-way test setting}

\begin{figure}[h!]
\begin{center}
\begin{tabular}{ccc}
\includegraphics[width=5cm]{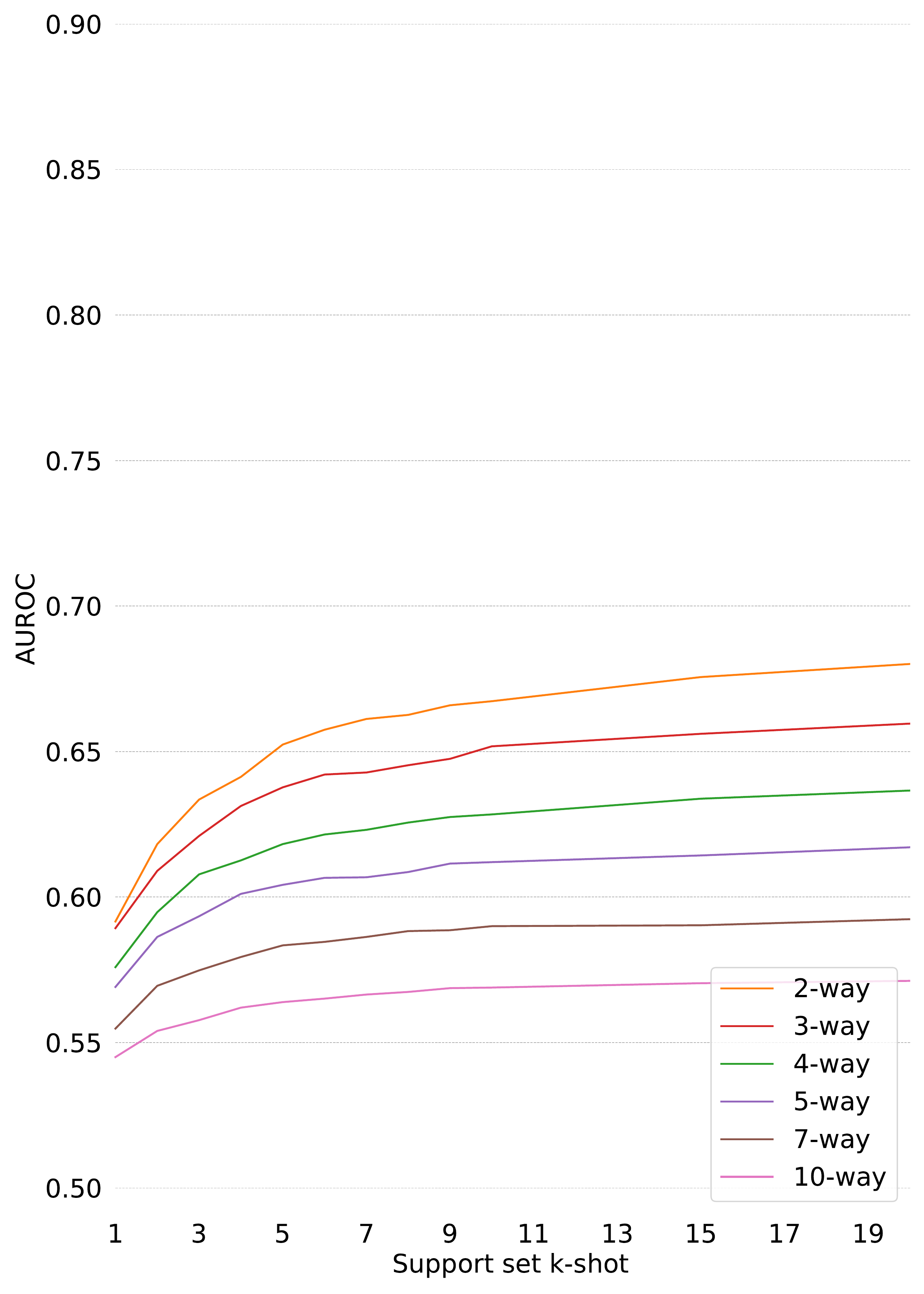}&
\includegraphics[width=5cm]{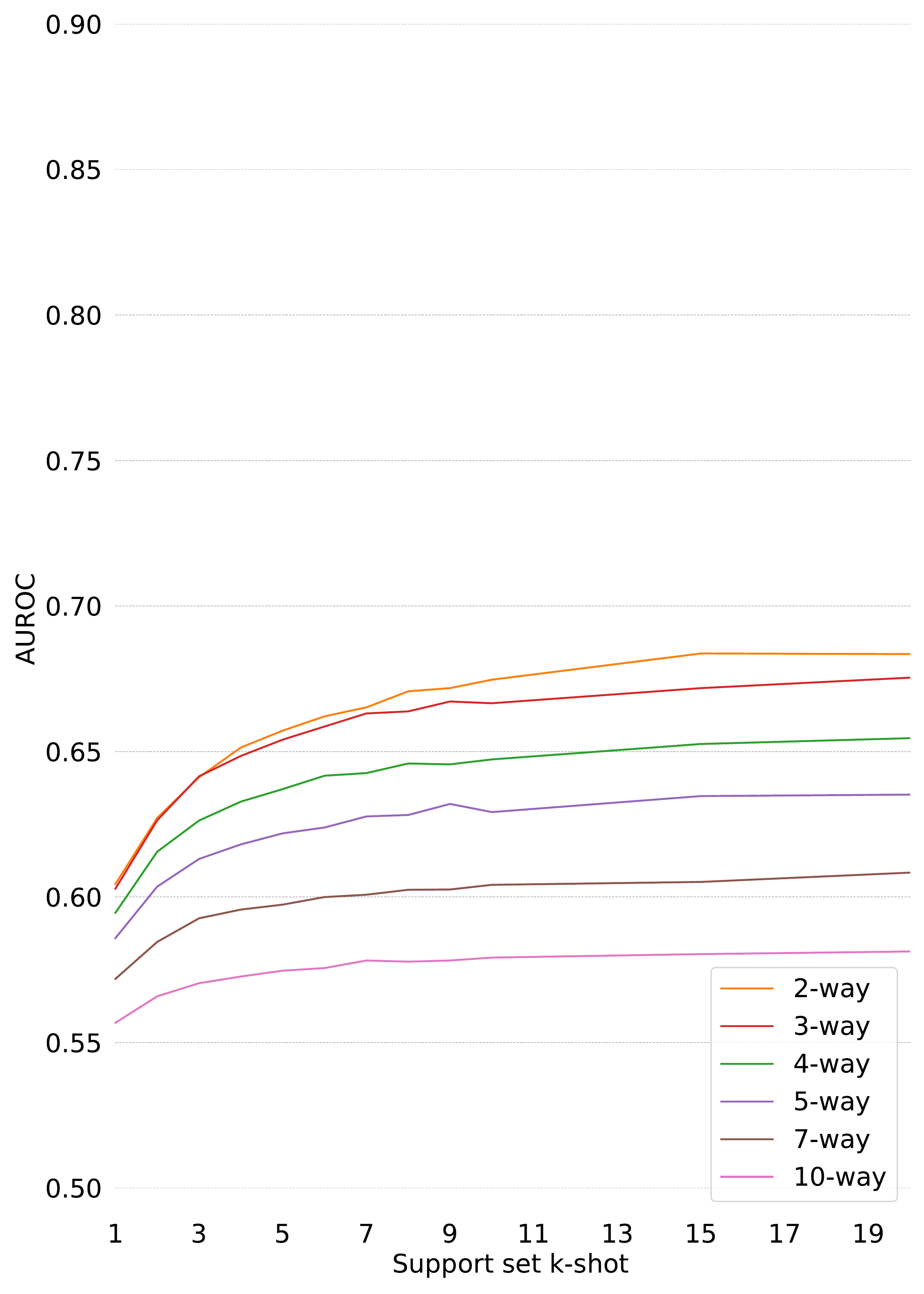}&
\includegraphics[width=5cm]{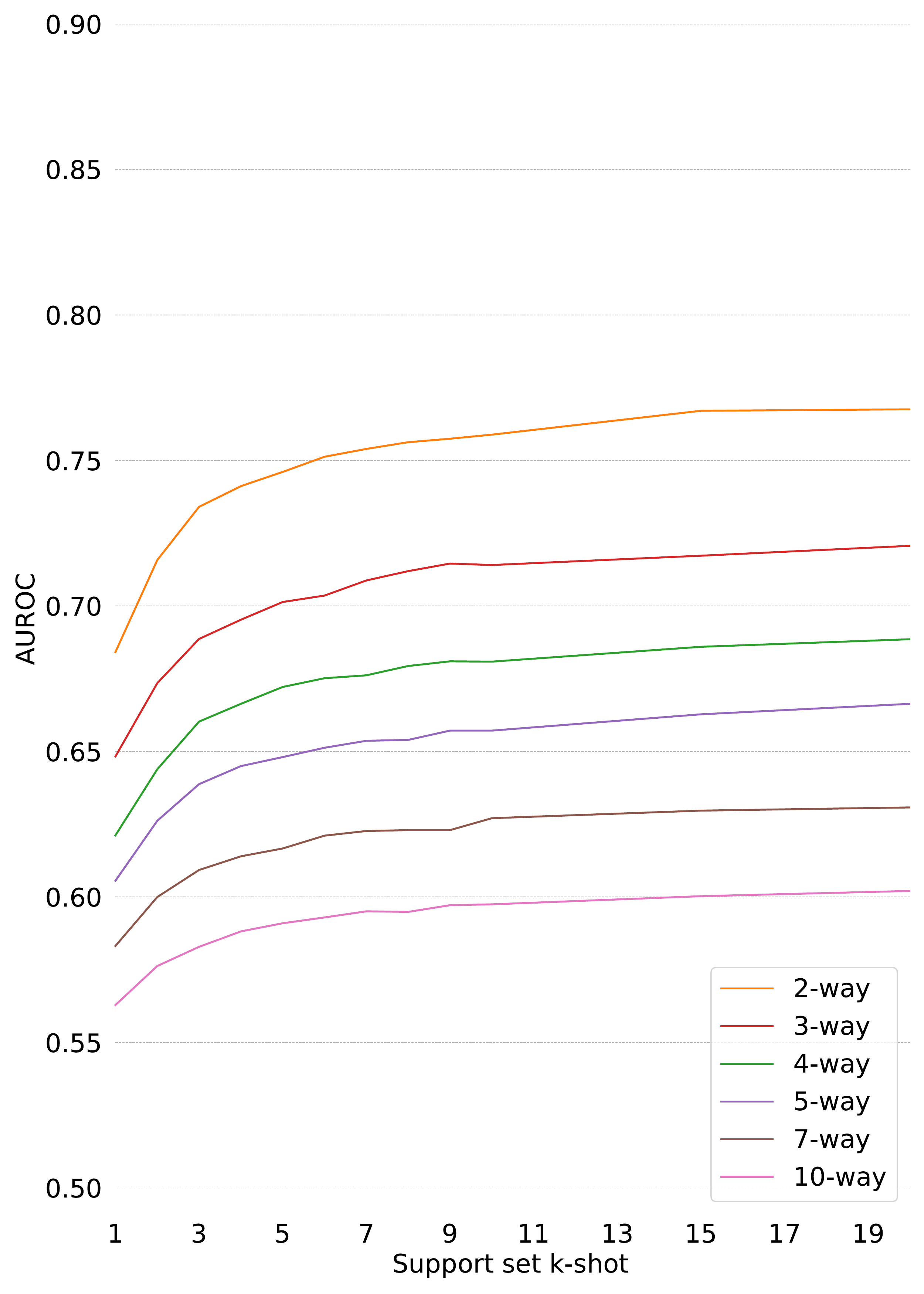}\\
(a) PEELER + threshold & (b) PEELER + Entropic Loss & (c) PEELER + \MetaBCE\\ 
\includegraphics[width=5cm]{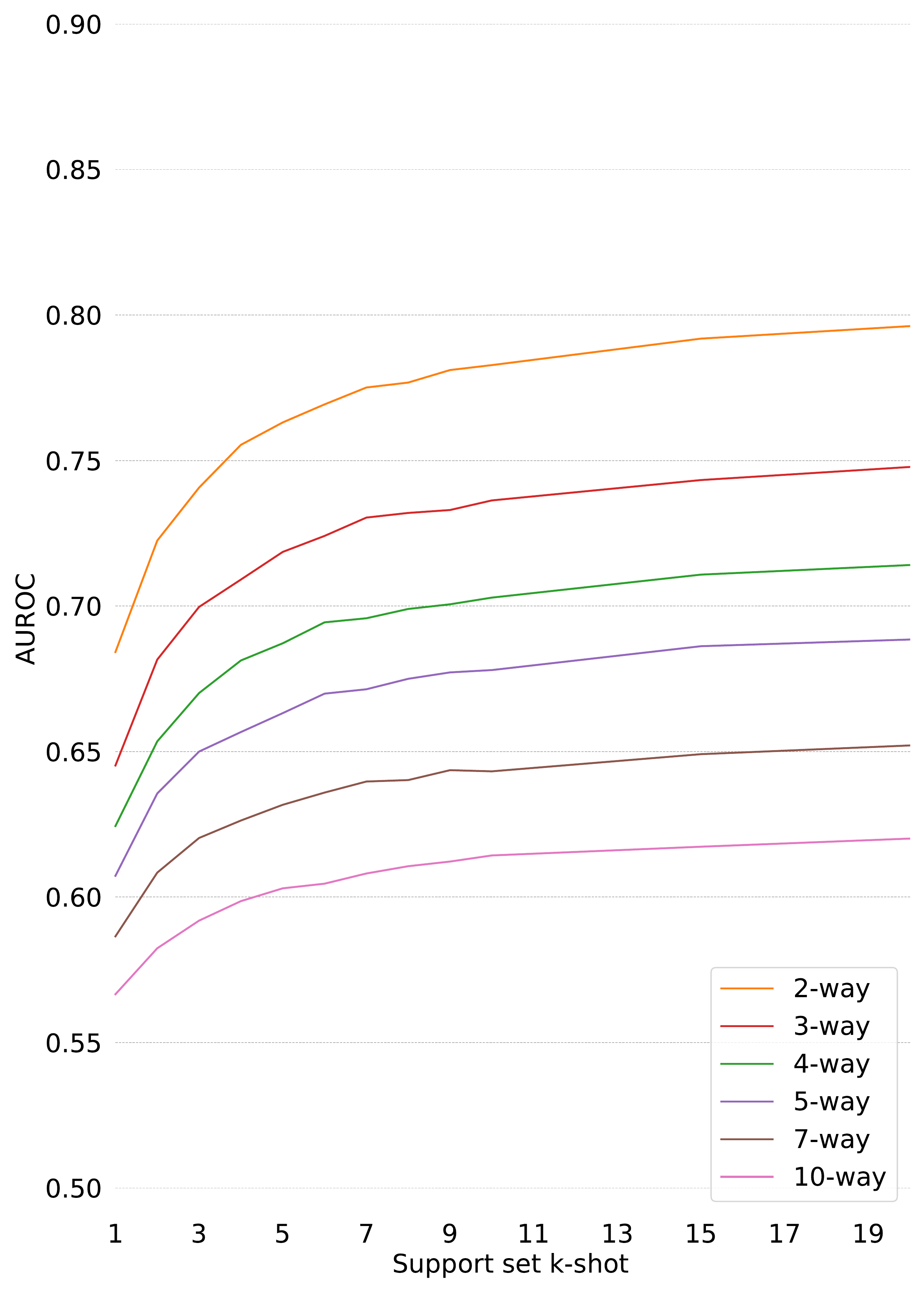}&
\includegraphics[width=5cm]{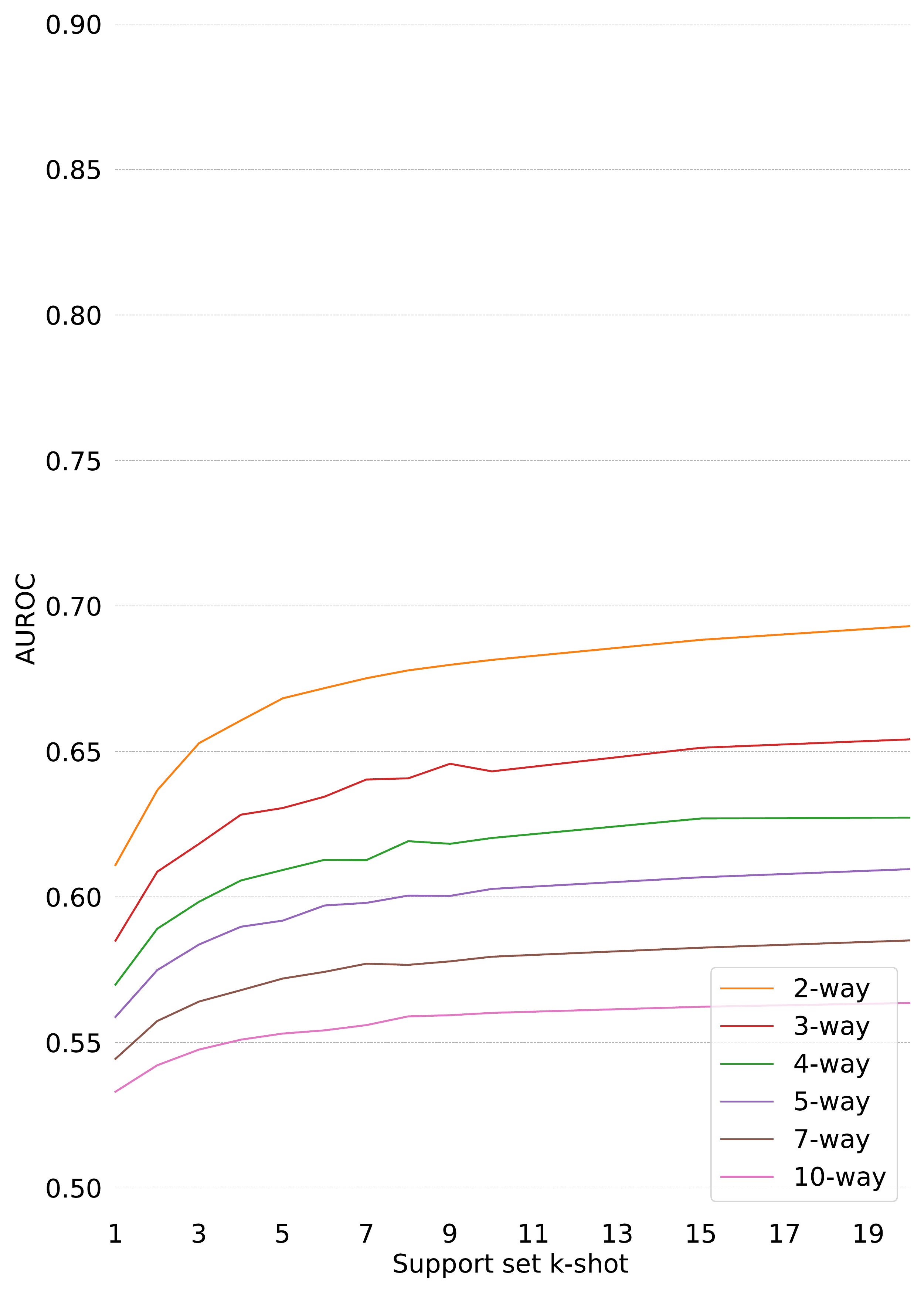}&
\includegraphics[width=5cm]{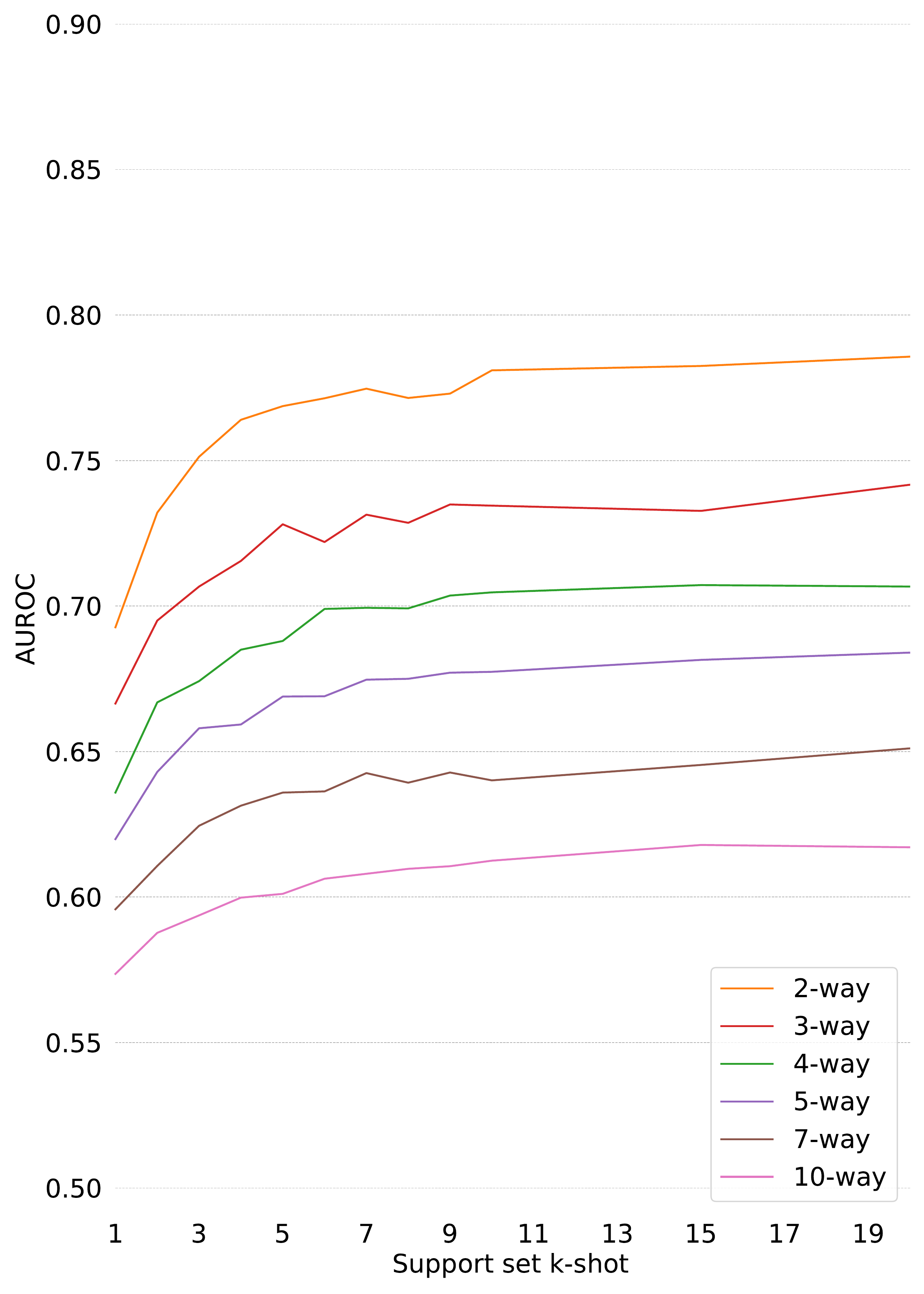}\\
(d) PEELER + \OCML & (e) FEAT + threshold & (f) FEAT + \MetaBCE \\ 
\includegraphics[width=5cm]{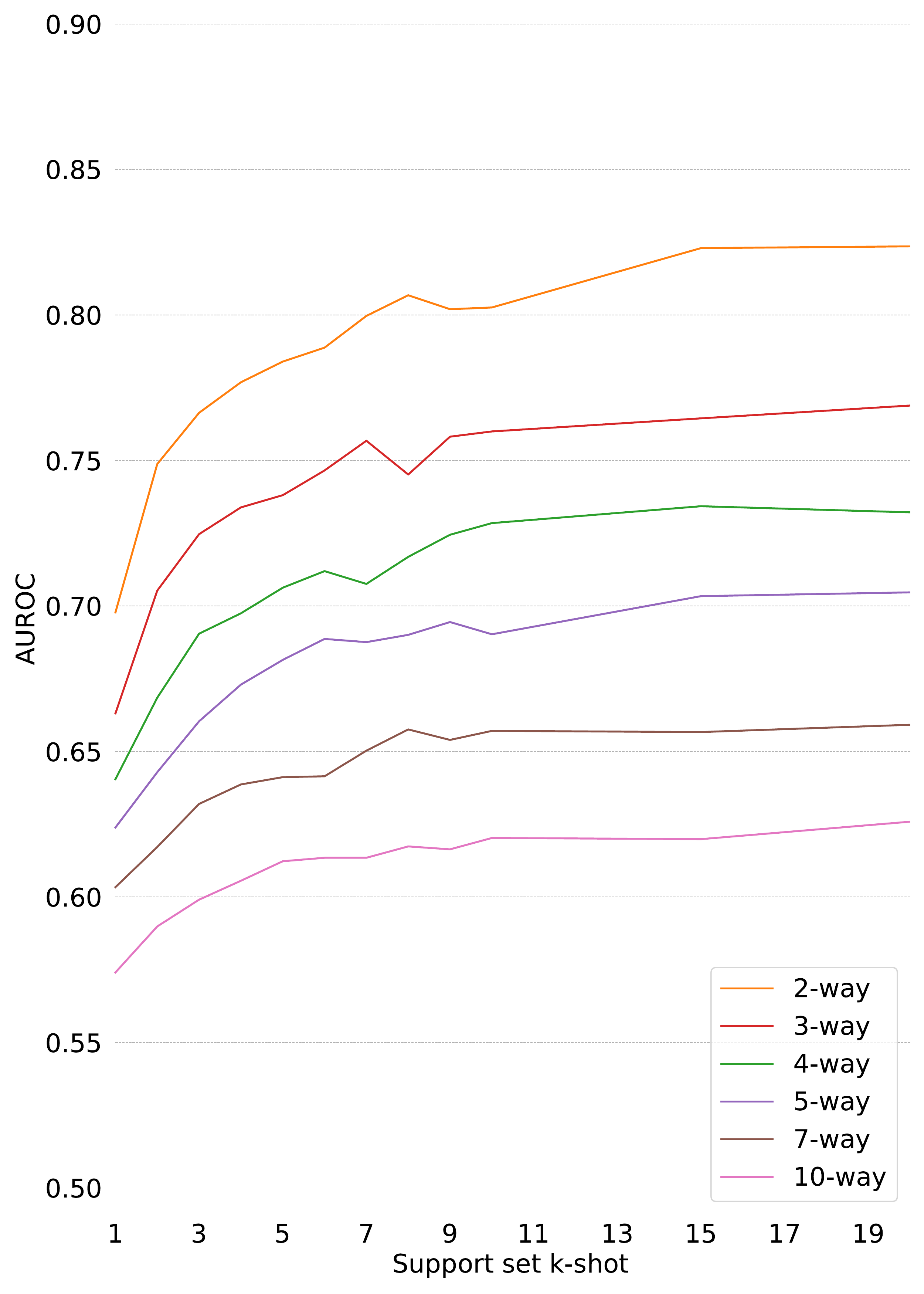} & &\\
(d) FEAT + \OCML &  &\\ 
\end{tabular}
\caption{AUROC score}
\label{fig:fsos_ablation_auroc}
\end{center}
\end{figure}

\begin{figure}[h!]
\begin{center}
\begin{tabular}{ccc}
\includegraphics[width=5cm]{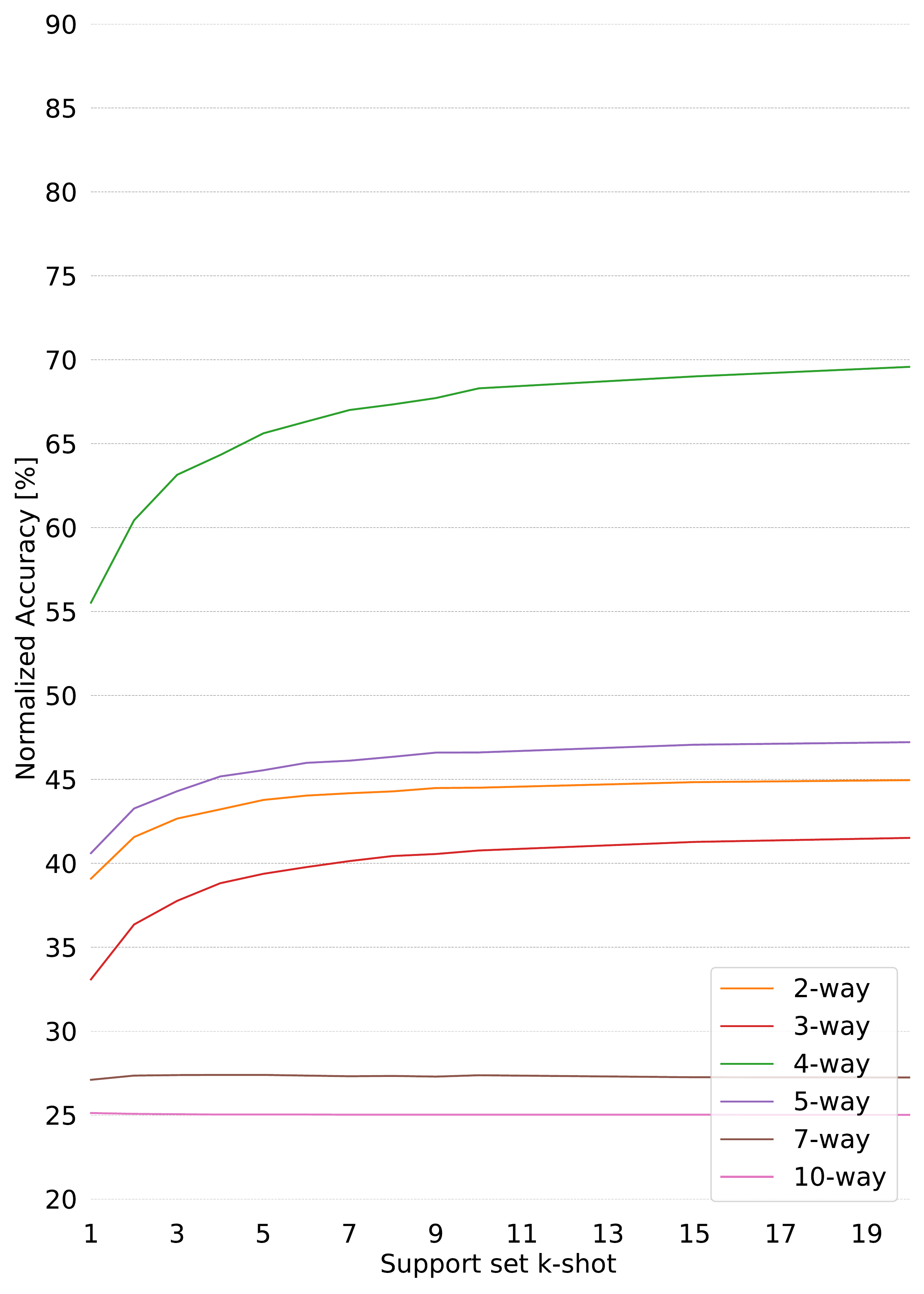}&
\includegraphics[width=5cm]{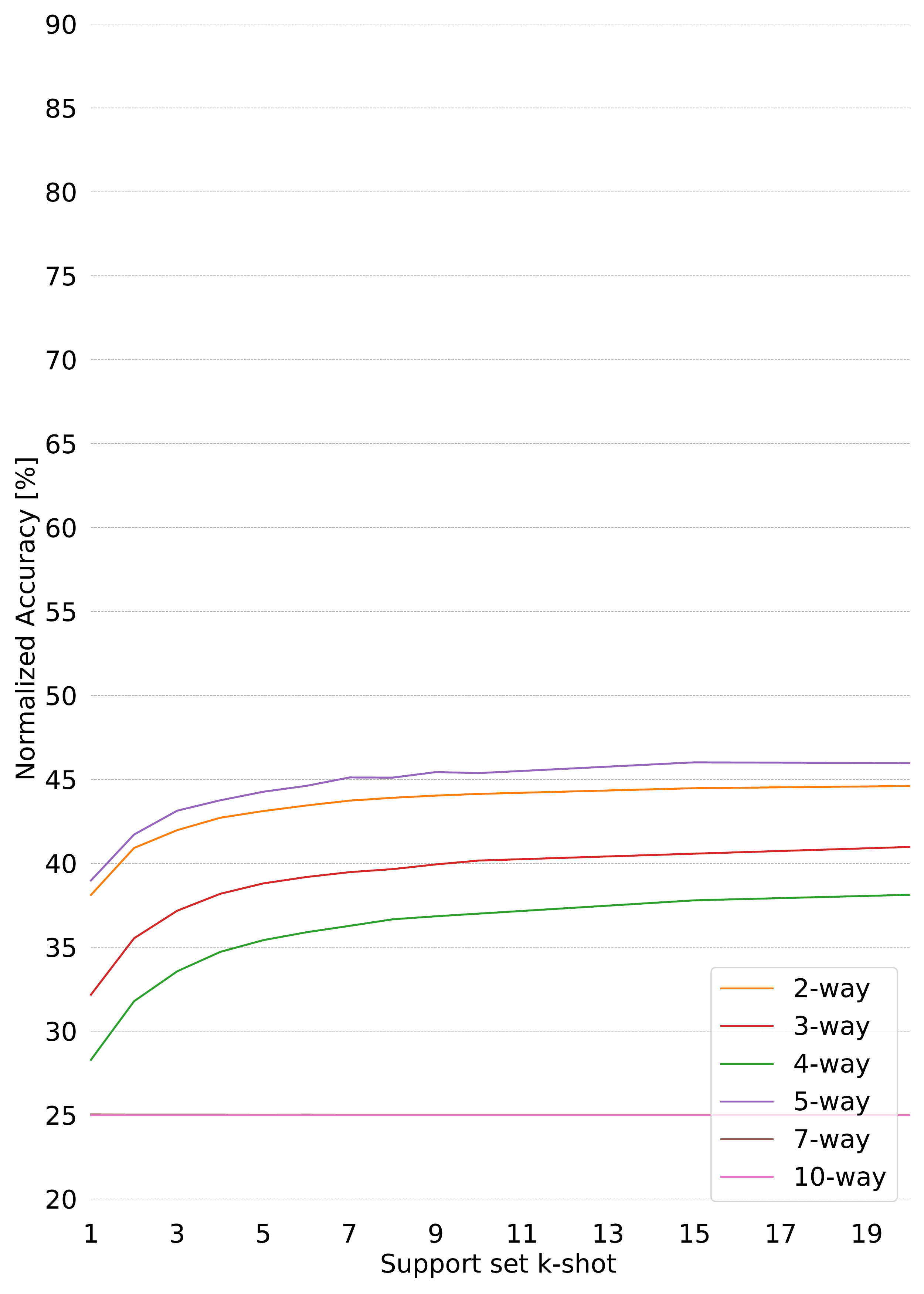}&
\includegraphics[width=5cm]{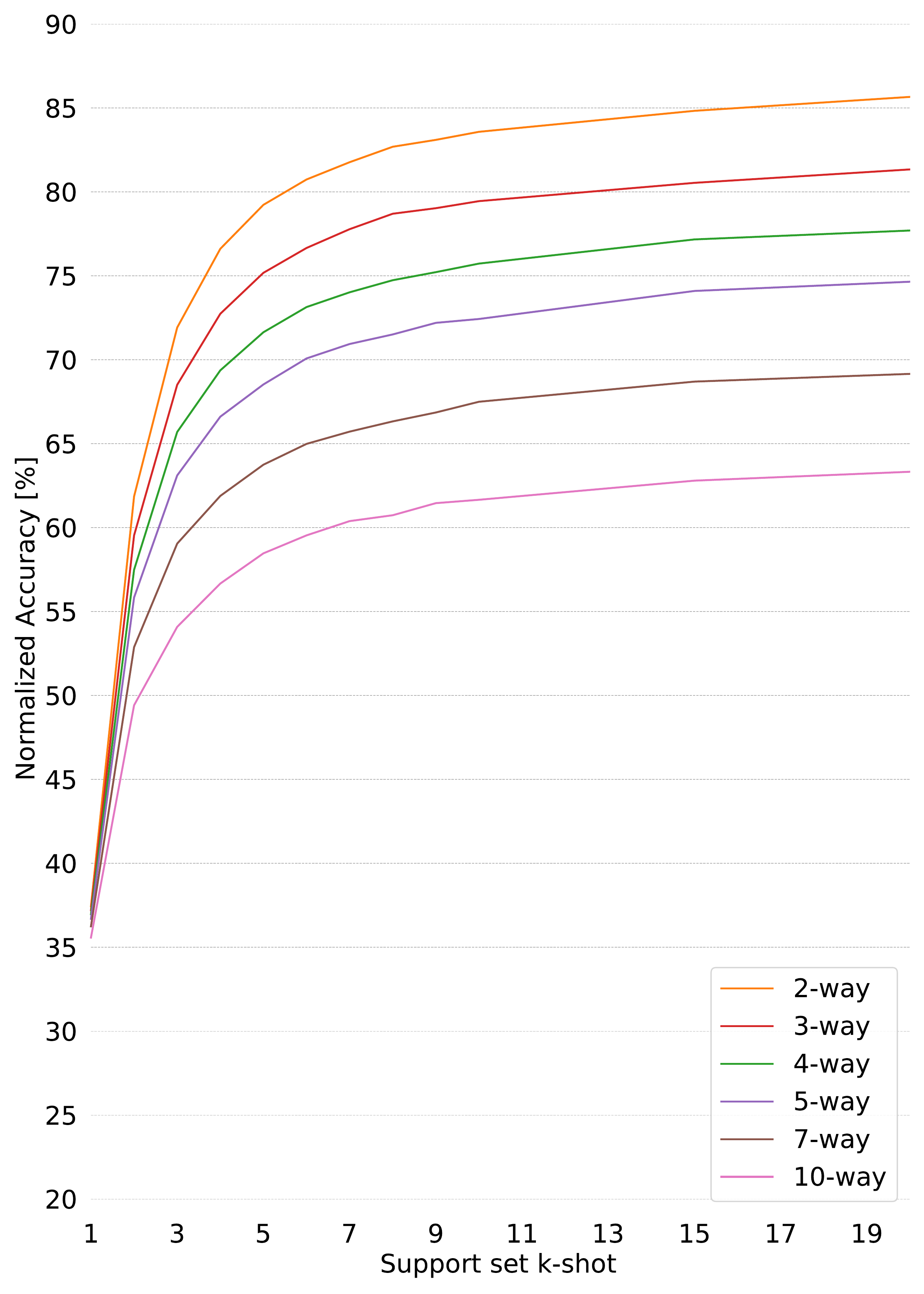}\\
(a) PEELER + threshold & (b) PEELER + Entropic Loss & (c) PEELER + \MetaBCE\\ 
\includegraphics[width=5cm]{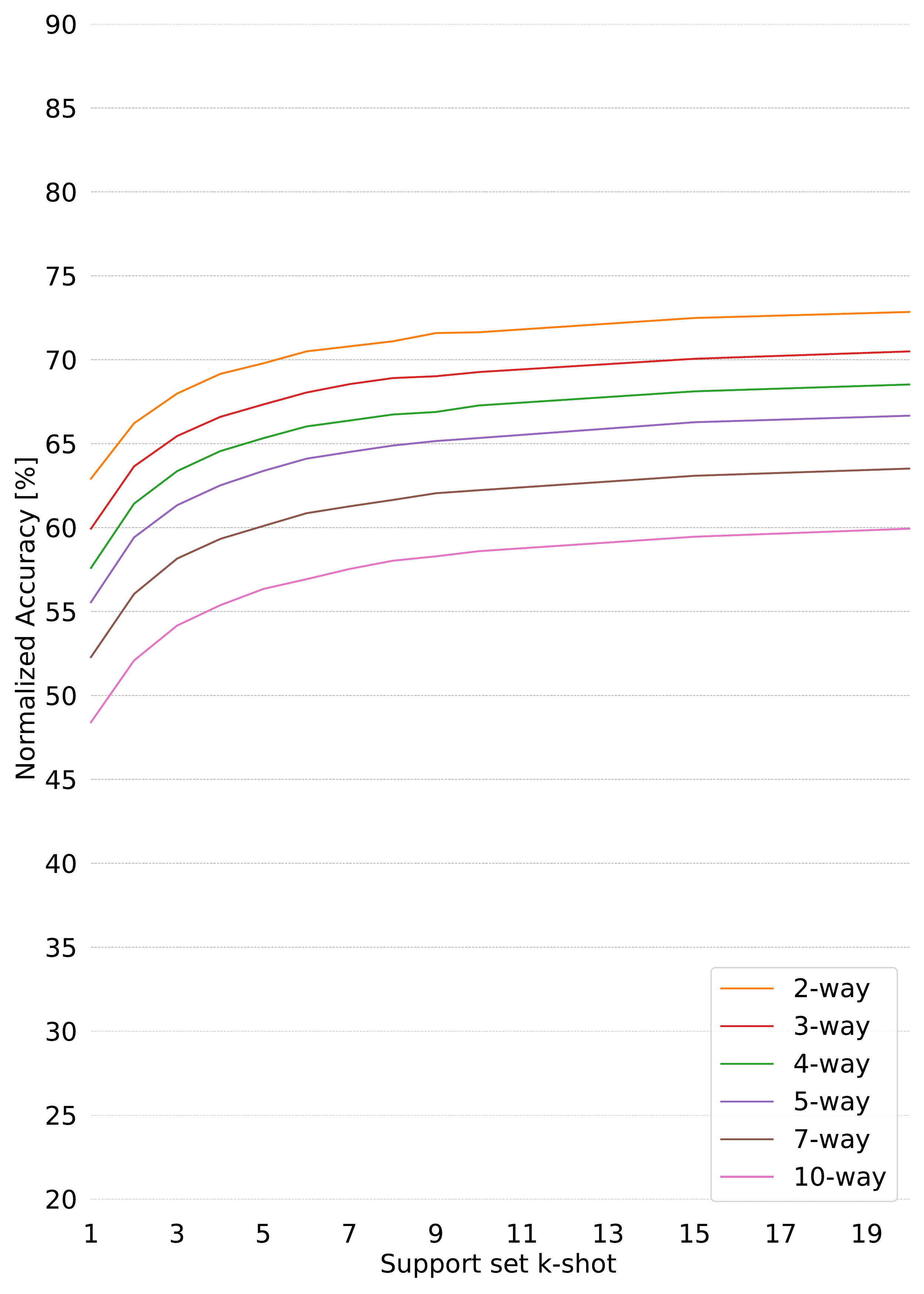}&
\includegraphics[width=5cm]{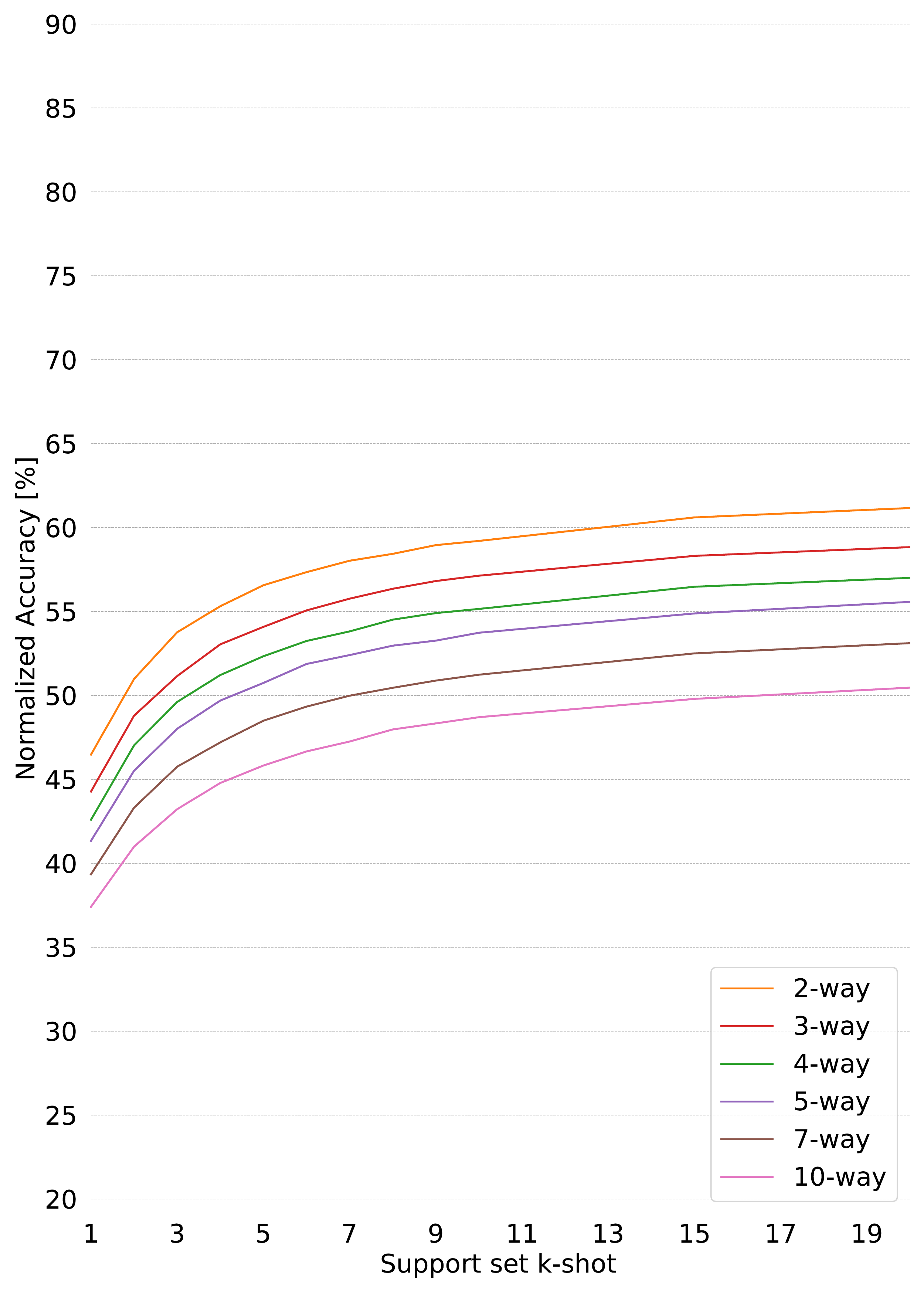}&
\includegraphics[width=5cm]{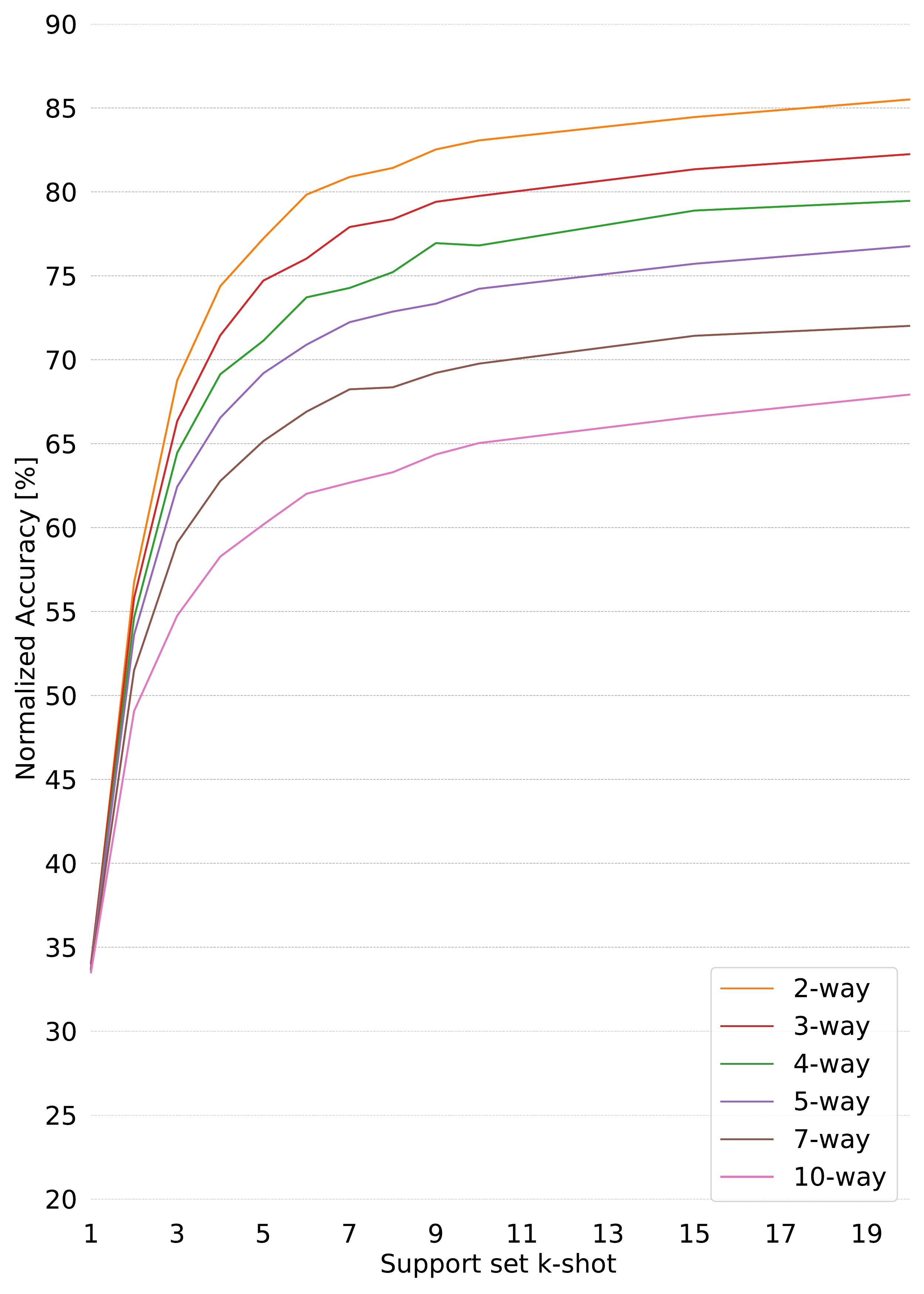}\\
(d) PEELER + \OCML & (e) FEAT + threshold & (f) FEAT + \MetaBCE \\ 
\includegraphics[width=5cm]{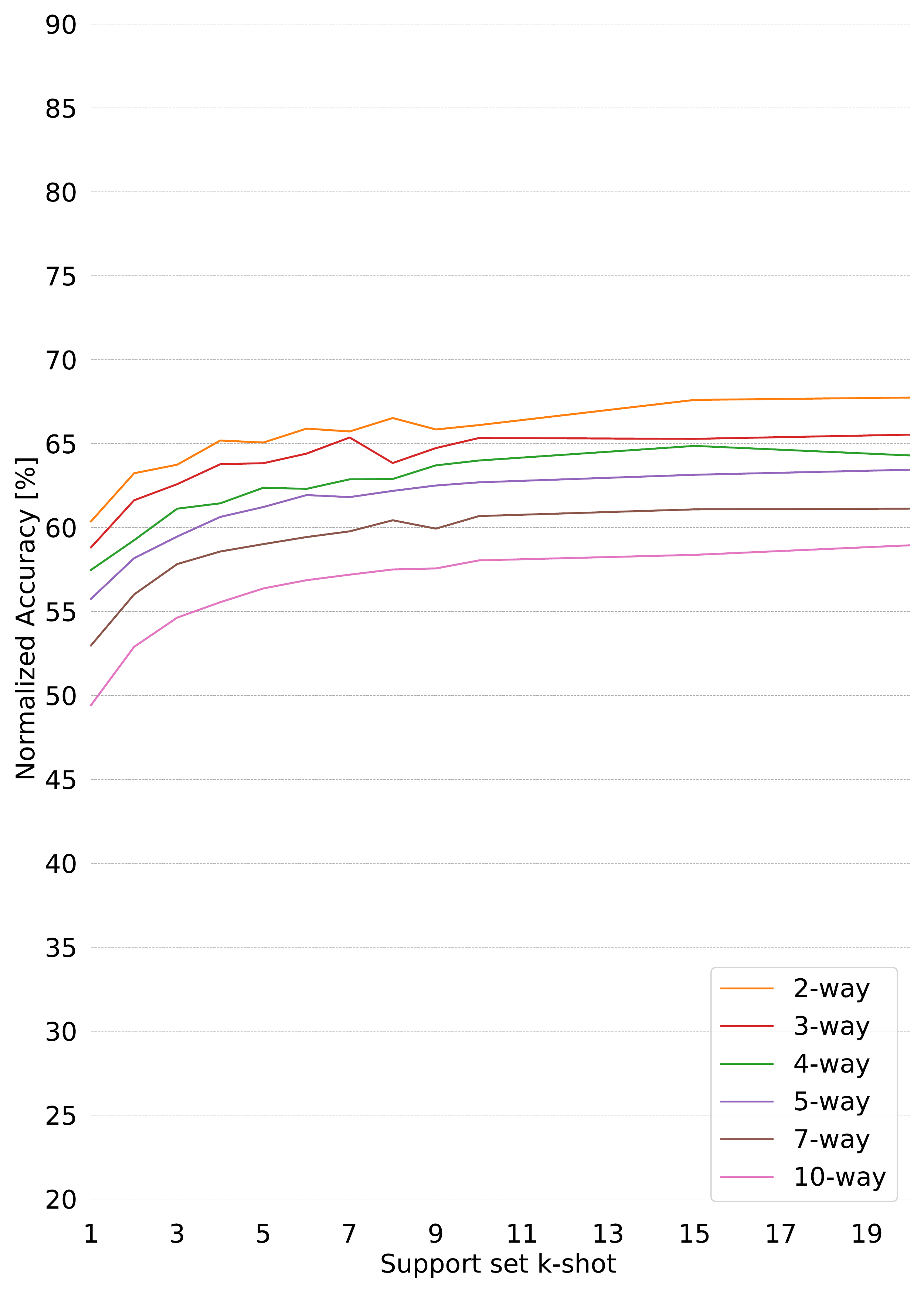} & &\\
(d) FEAT + \OCML &  &\\ 
\end{tabular}
\caption{Normalized accuracy}
\label{fig:fsos_ablation_na}
\end{center}
\end{figure}

\begin{figure}[h!]
\begin{center}
\begin{tabular}{ccc}
\includegraphics[width=5cm]{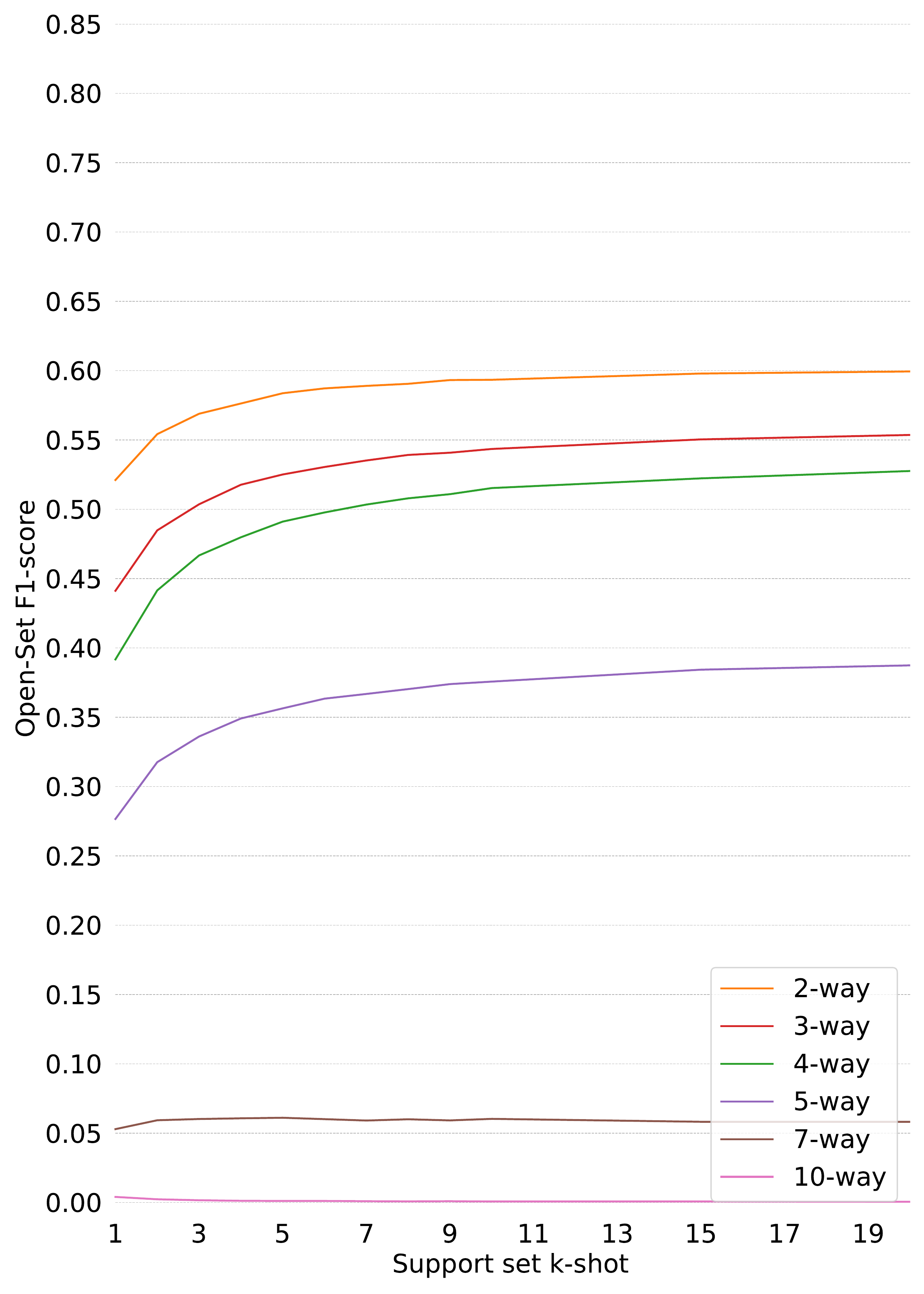}&
\includegraphics[width=5cm]{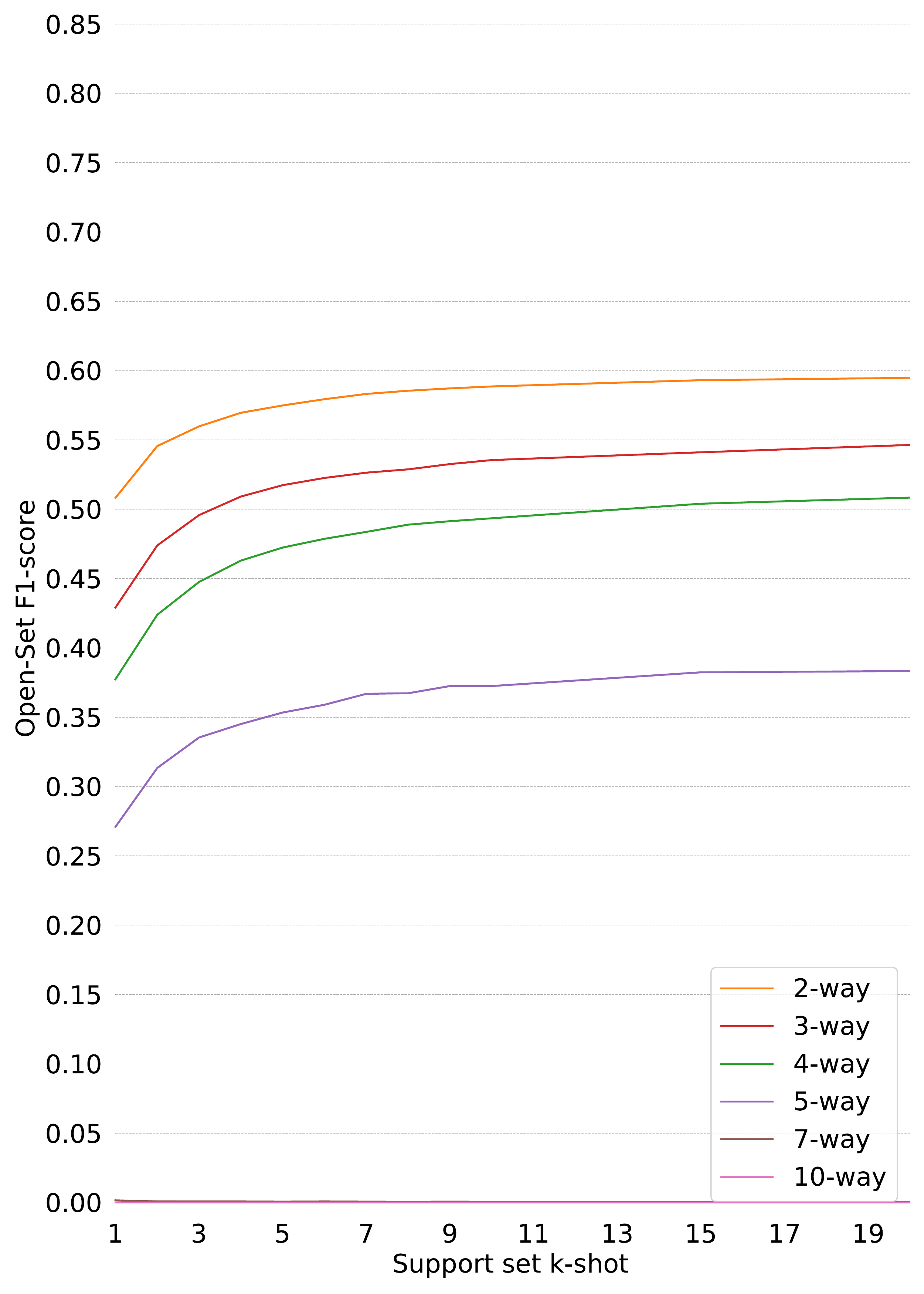}&
\includegraphics[width=5cm]{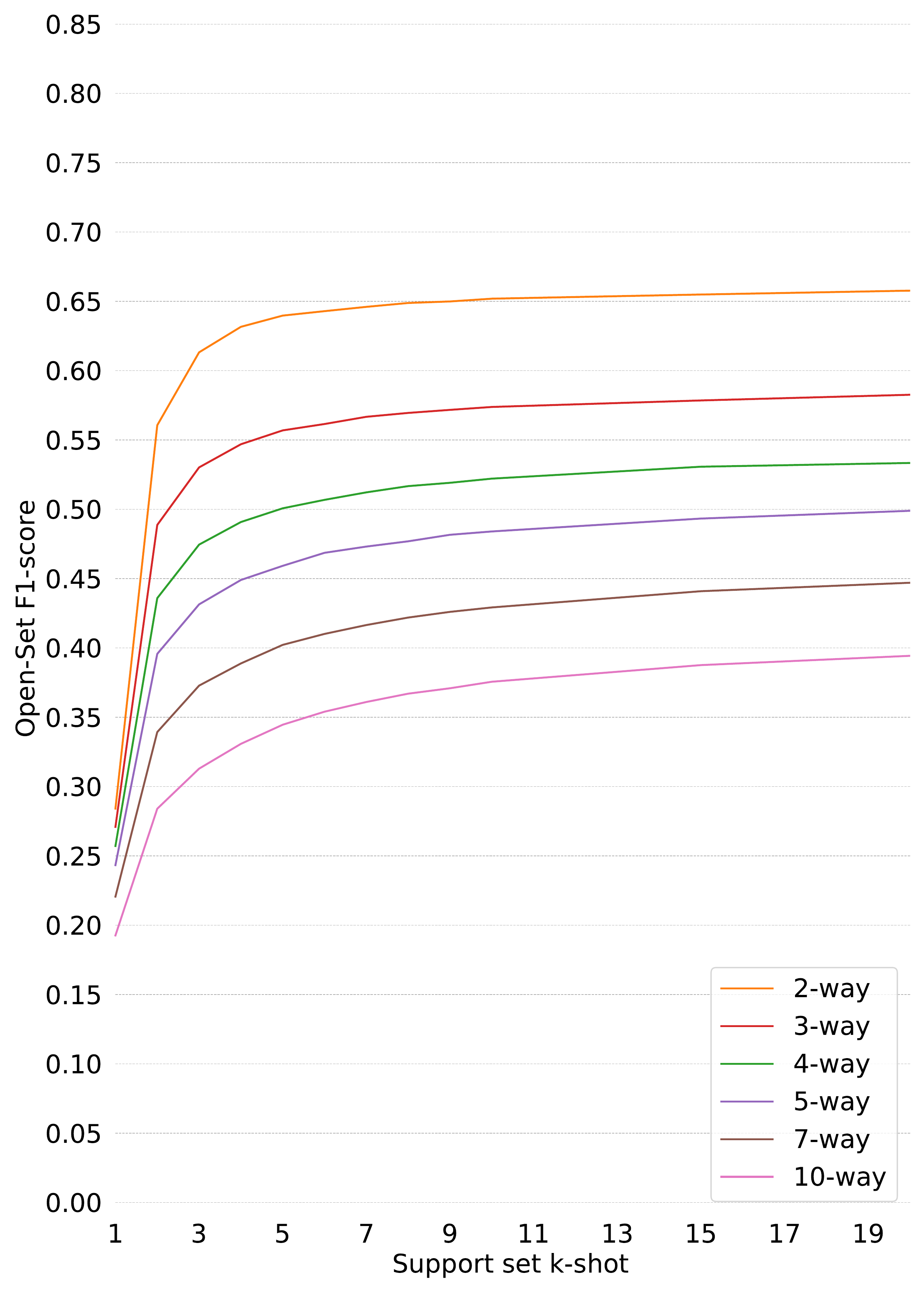}\\
(a) PEELER + threshold & (b) PEELER + Entropic Loss & (c) PEELER + \MetaBCE\\ 
\includegraphics[width=5cm]{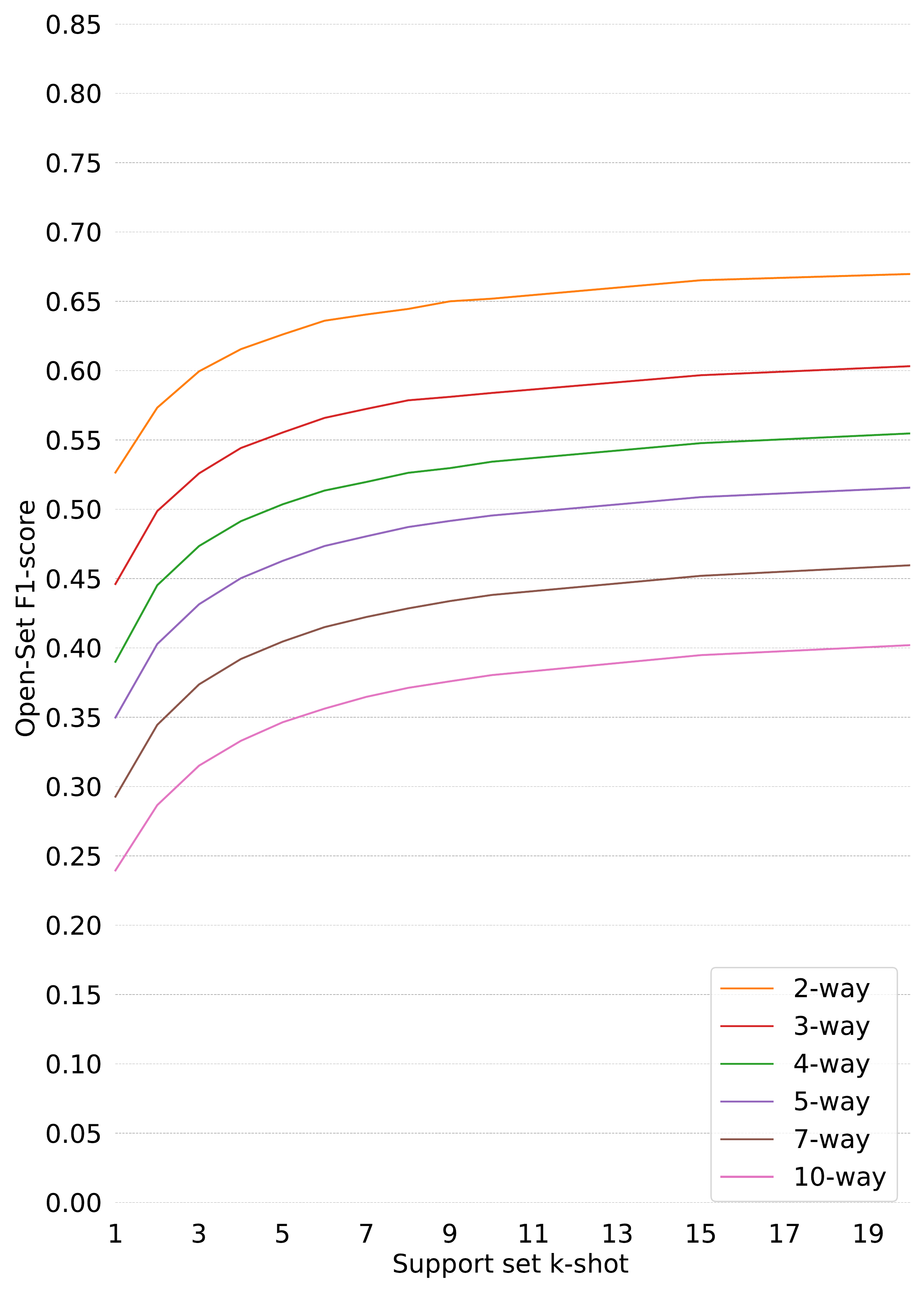}&
\includegraphics[width=5cm]{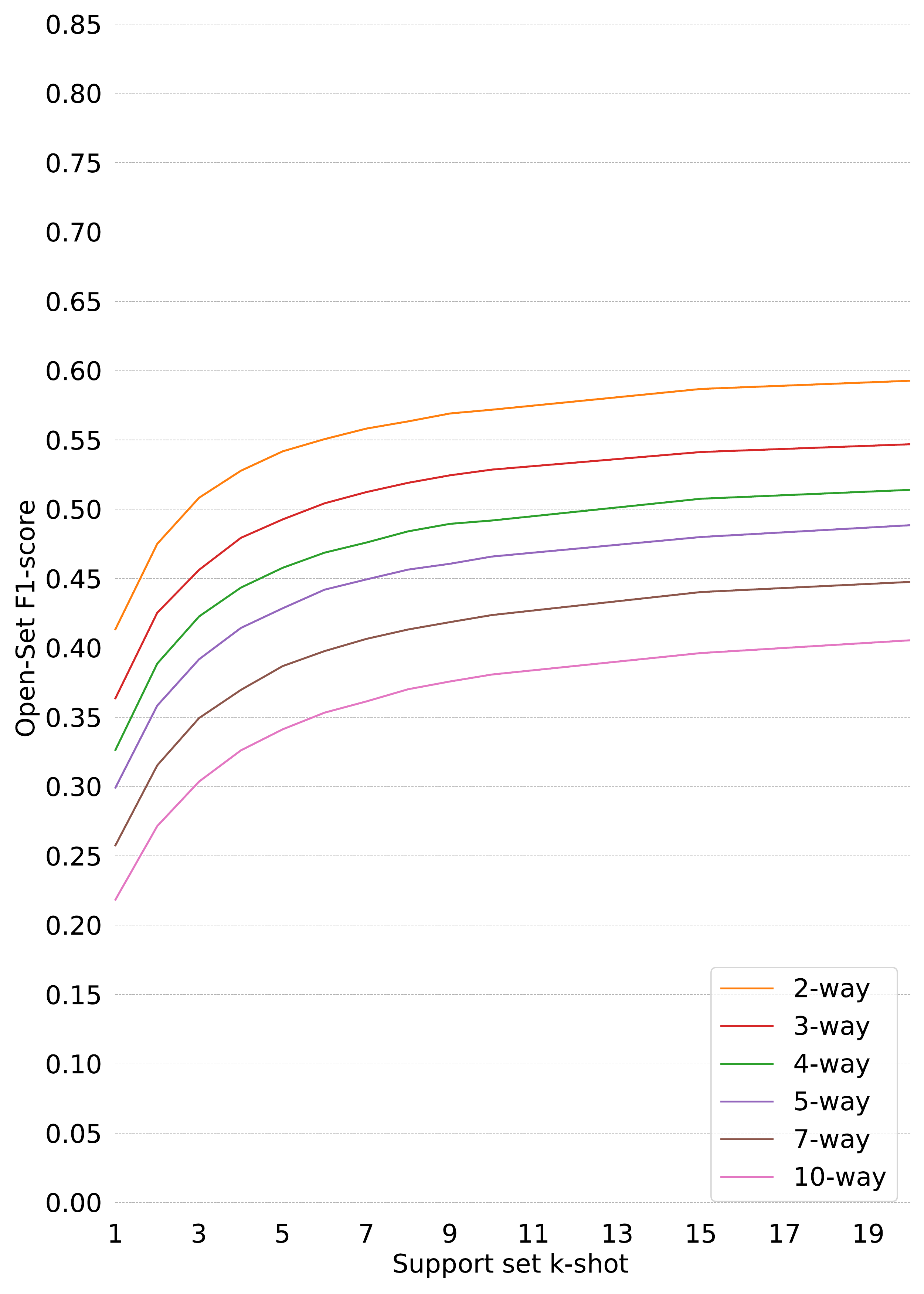}&
\includegraphics[width=5cm]{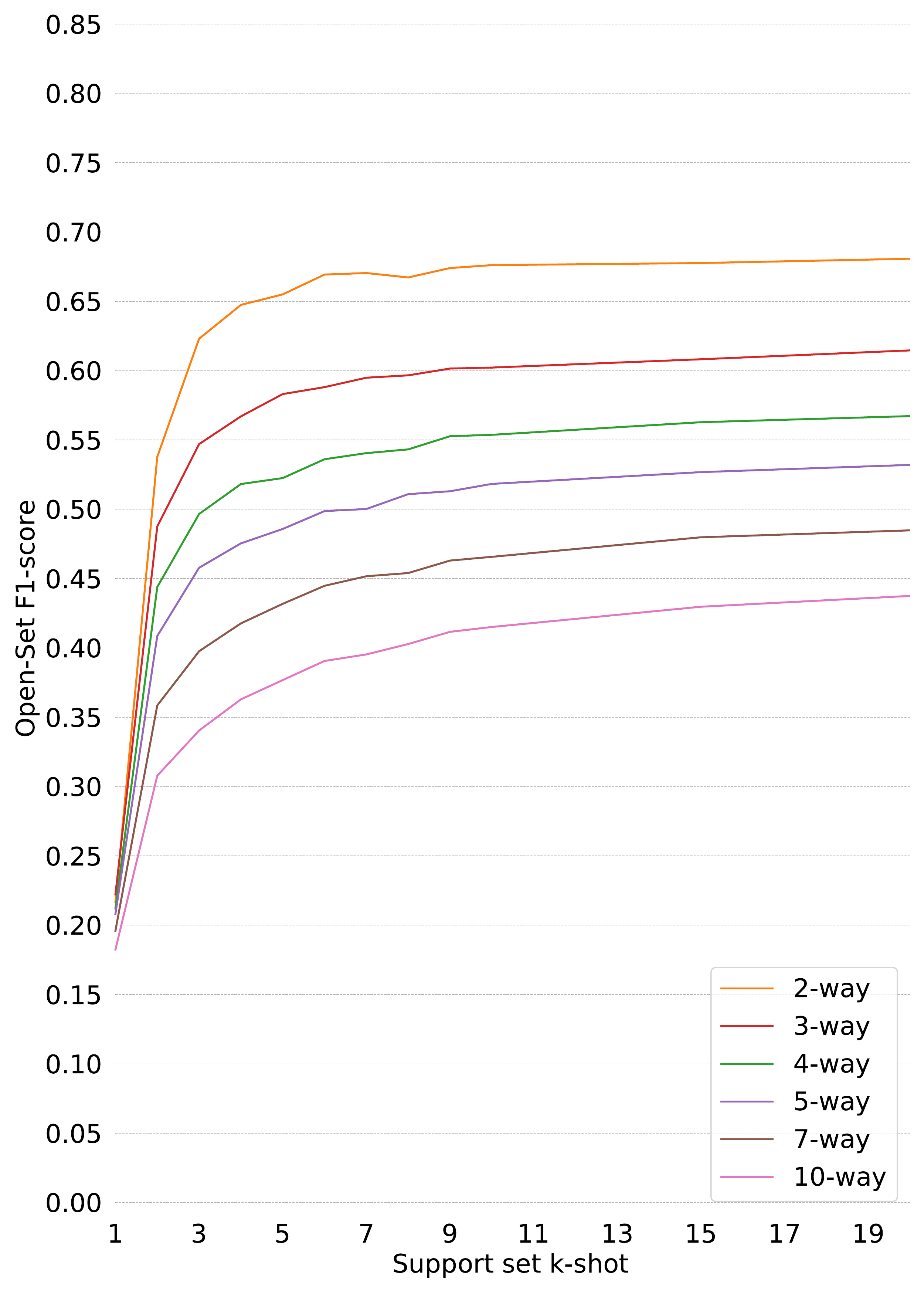}\\
(d) PEELER + \OCML & (e) FEAT + threshold & (f) FEAT + \MetaBCE \\ 
\includegraphics[width=5cm]{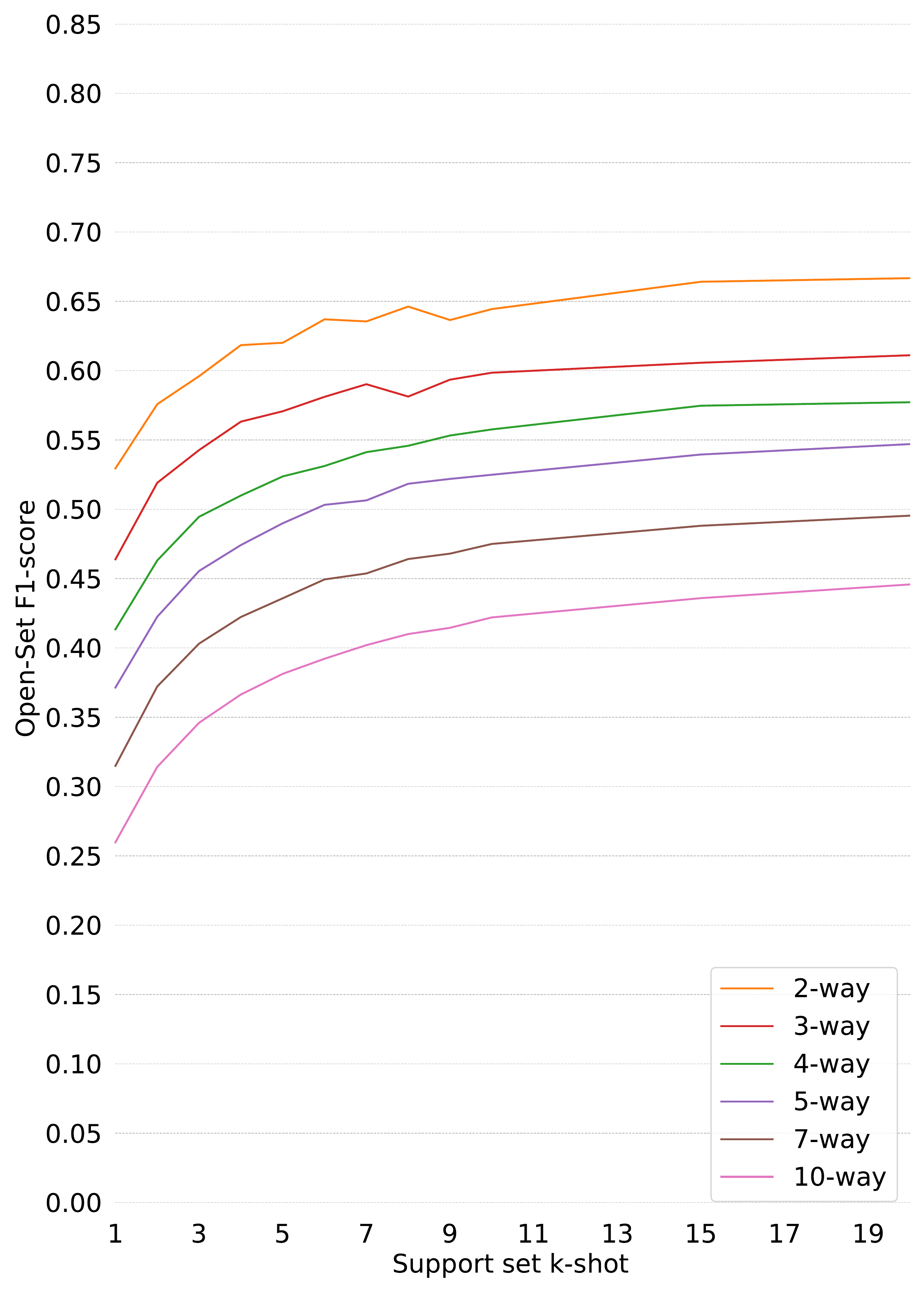} & &\\
(d) FEAT + \OCML &  &\\ 
\end{tabular}
\caption{F1-open score}
\label{fig:fsos_ablation_f1}
\end{center}
\end{figure}

We have performed also a more thorough analysis of studied methods when varying both number of categories and number of per-category examples in the open-set setting. We provide the results for this analysis below in Figure~\ref{fig:fsos_ablation_accuracy} (for closed-set accuracy), Figure~\ref{fig:fsos_ablation_auroc} (for AUROC score), Figure~\ref{fig:fsos_ablation_na} (for normalized accuracy), and Figure~\ref{fig:fsos_ablation_f1} (for F1-open score). In all experiments number of unknown categories ($n_U$) was equal to the number of known categories ($n_u = n$).

When analyzing the closed-set accuracy (Figure~\ref{fig:fsos_ablation_accuracy}) we can see that for all methods (as expected) higher number of per-category examples ($k$) increases the performance and higher number of categories ($n$) decreases the performance. Threshold, \MetaBCE, and \OCML~do not impact the closed-set performance (thus their plots are comparable) and Entropic Open-Set Loss~\cite{dhamija2018reducing} reduces the closed-set accuracy of PEELER~\cite{liu2020few}. AUROC score comparison (Figure~\ref{fig:fsos_ablation_auroc} indicates that \MetaBCE~and \OCML~has much higher scores for all values of $k$ and $n$ than original PEELER~\cite{liu2020few} or PEELER with Entropic Open-Set Loss~\cite{dhamija2018reducing}. And we can further increase the performance by substituting PEELER~\cite{liu2020few} with FEAT~\cite{ye2020fewshot} as the closed-set training method.

Figures showcasing the classification metrics (normalized accuracy in Figure~\ref{fig:fsos_ablation_na} and F1-open score in Figure~\ref{fig:fsos_ablation_f1} indicate few interesting properties. Both PEELER~\cite{liu2020few} and PEELER with Entropic Open-Set Loss~\cite{dhamija2018reducing} have low scores, and for some values of $n$ they classify all query examples as unknown examples. The reason for this behavior is the discrepancy between the training setting of $5$-way $1$-shot and the test setting showcased in Figures and the fact that both methods are basing their classification decision (whether a sample is known or unknown) based on the threshold learned during the training phase. We can see that such problem does not occur in any other method. Another important property is the difference in behavior between \OCML~and \MetaBCE. \OCML~starts with high performance (when $k=1$) and slowly increases it when $k$ increases. \MetaBCE~on the other hand has low initial performance (when $k=1$), but it has a very rapid gain in both normalized accuracy and F1-open score when $k$ increases achieving values for normalized accuracy above $85\%$ when $n=2$ and $k=20$, which is $\sim12\%$ higher than \OCML, $\sim23\%$ higher than possible with a simple thresholding, and $\sim40\%$ higher than with Entropic Open-Set Loss~\cite{dhamija2018reducing}.

\newpage

\begin{figure}[h!]
\begin{center}
\begin{tabular}{ccc}
\includegraphics[width=5cm]{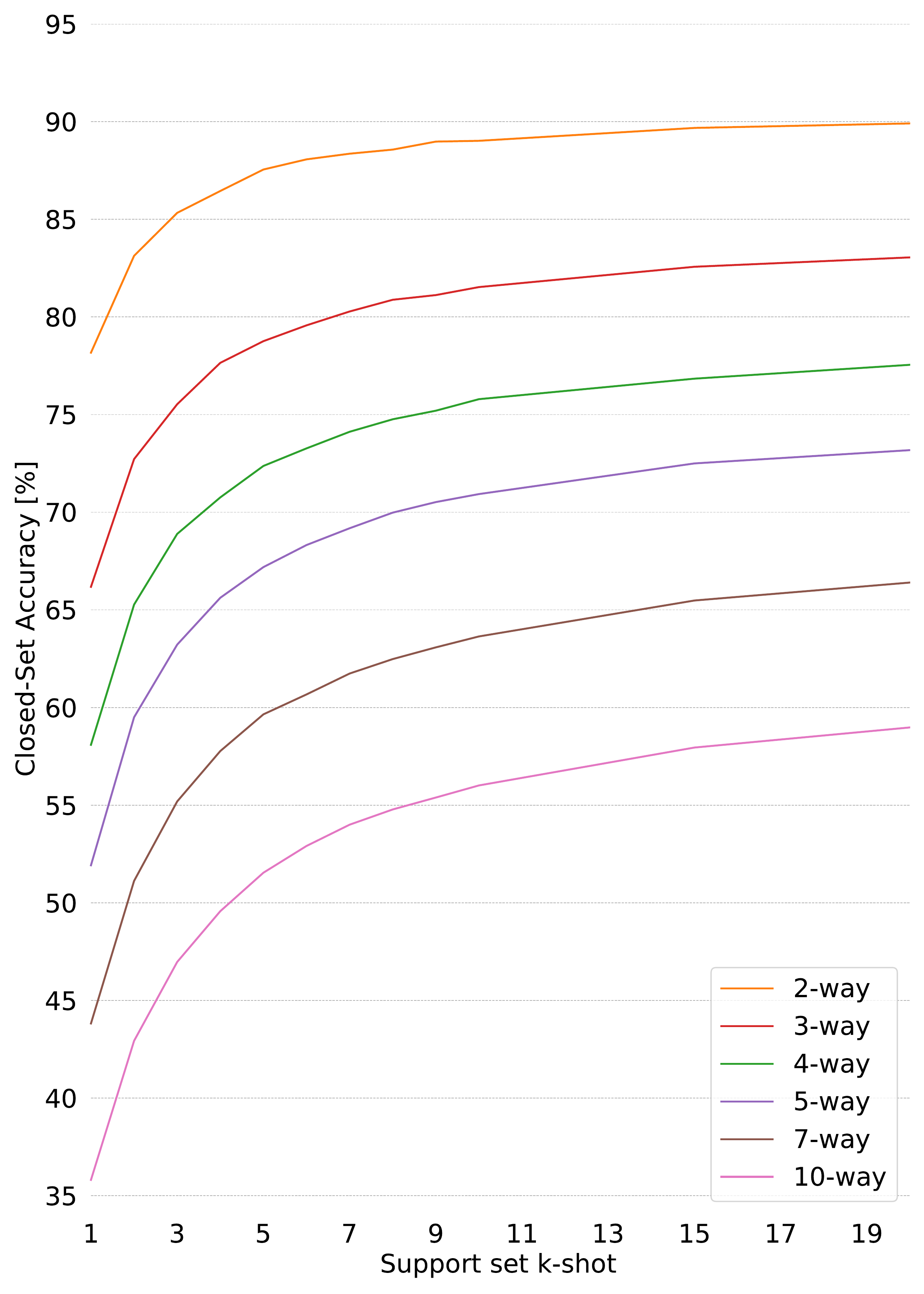}&
\includegraphics[width=5cm]{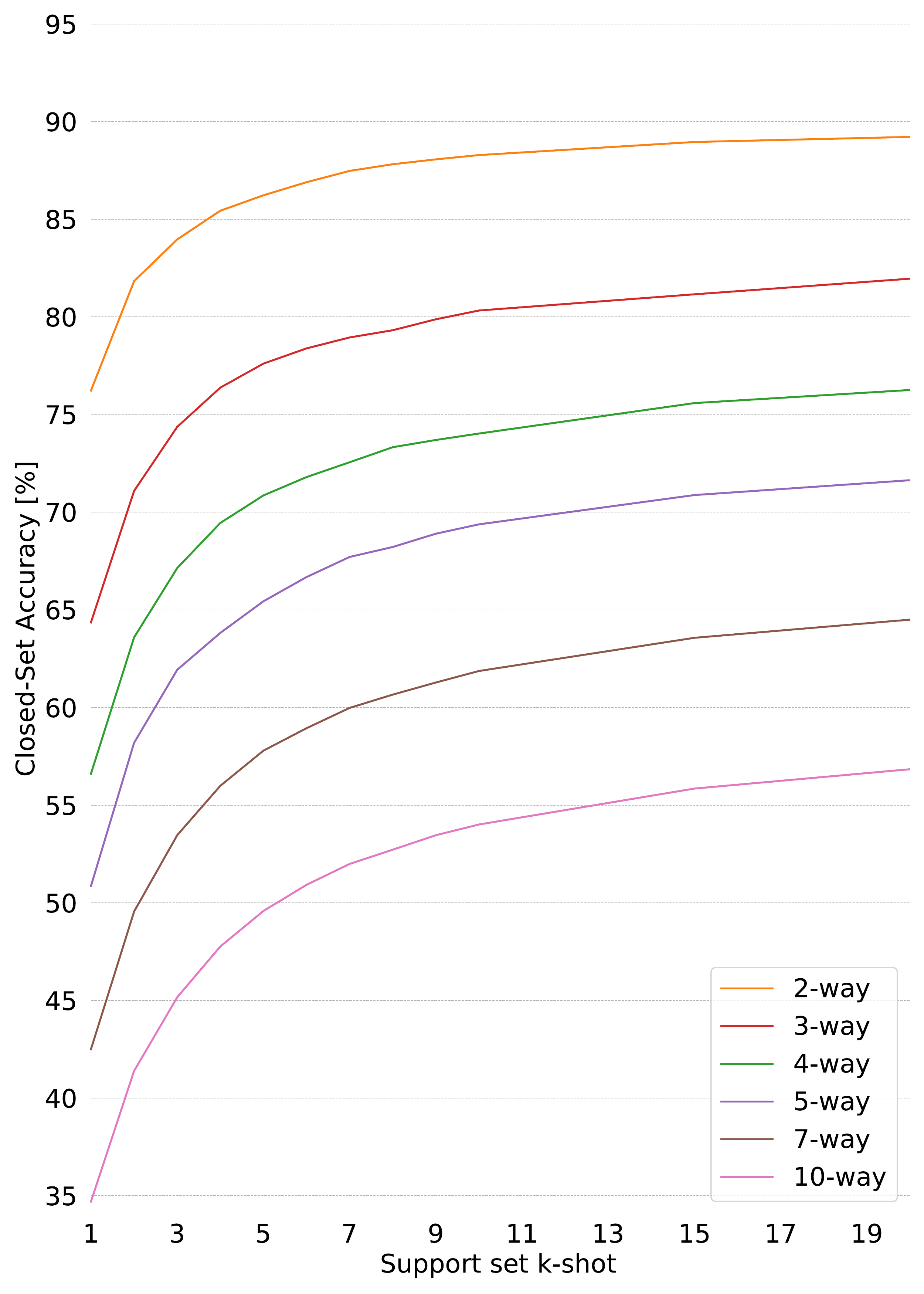}&
\includegraphics[width=5cm]{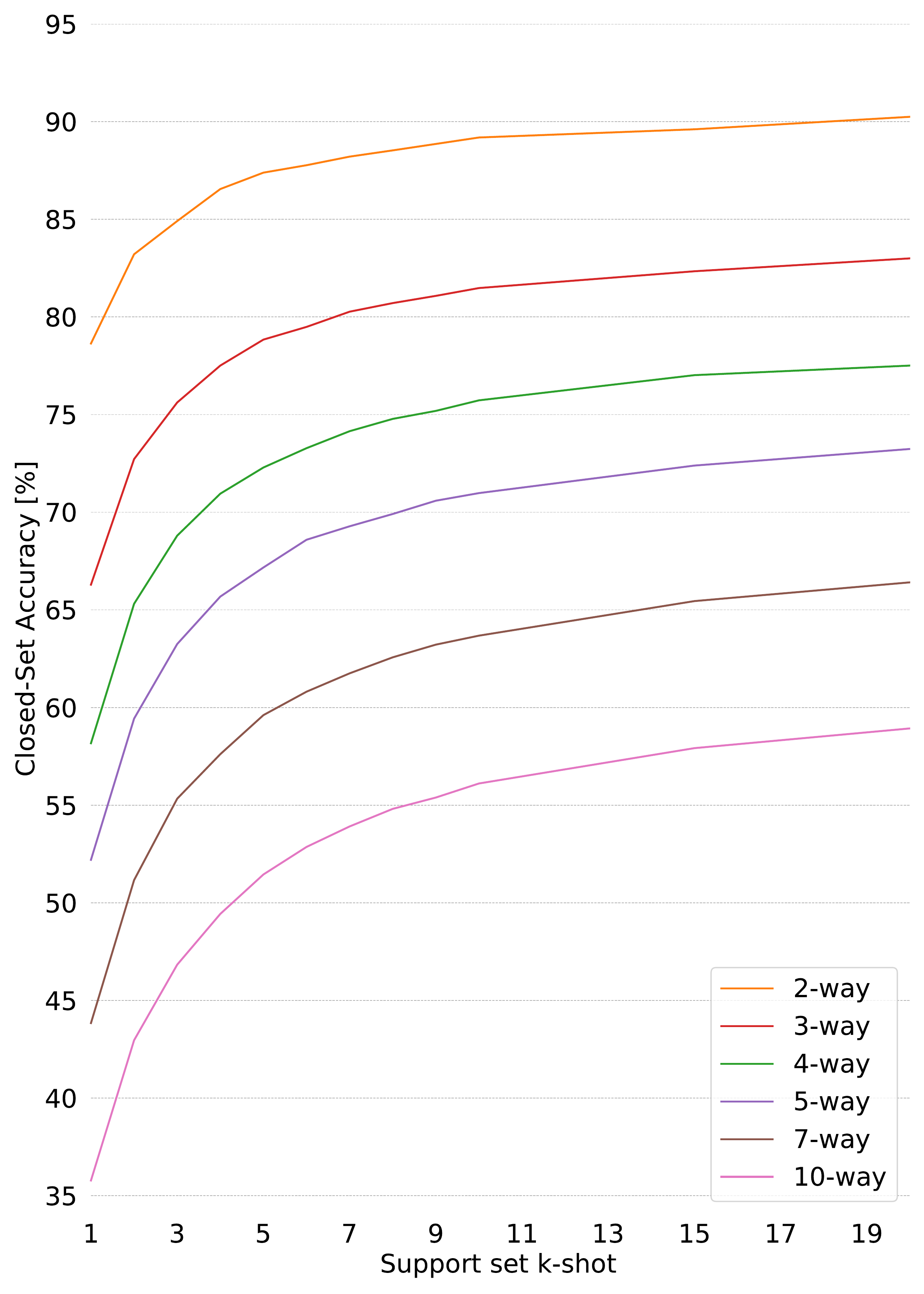}\\
(a) PEELER + threshold & (b) PEELER + Entropic Loss & (c) PEELER + \MetaBCE\\ 
\includegraphics[width=5cm]{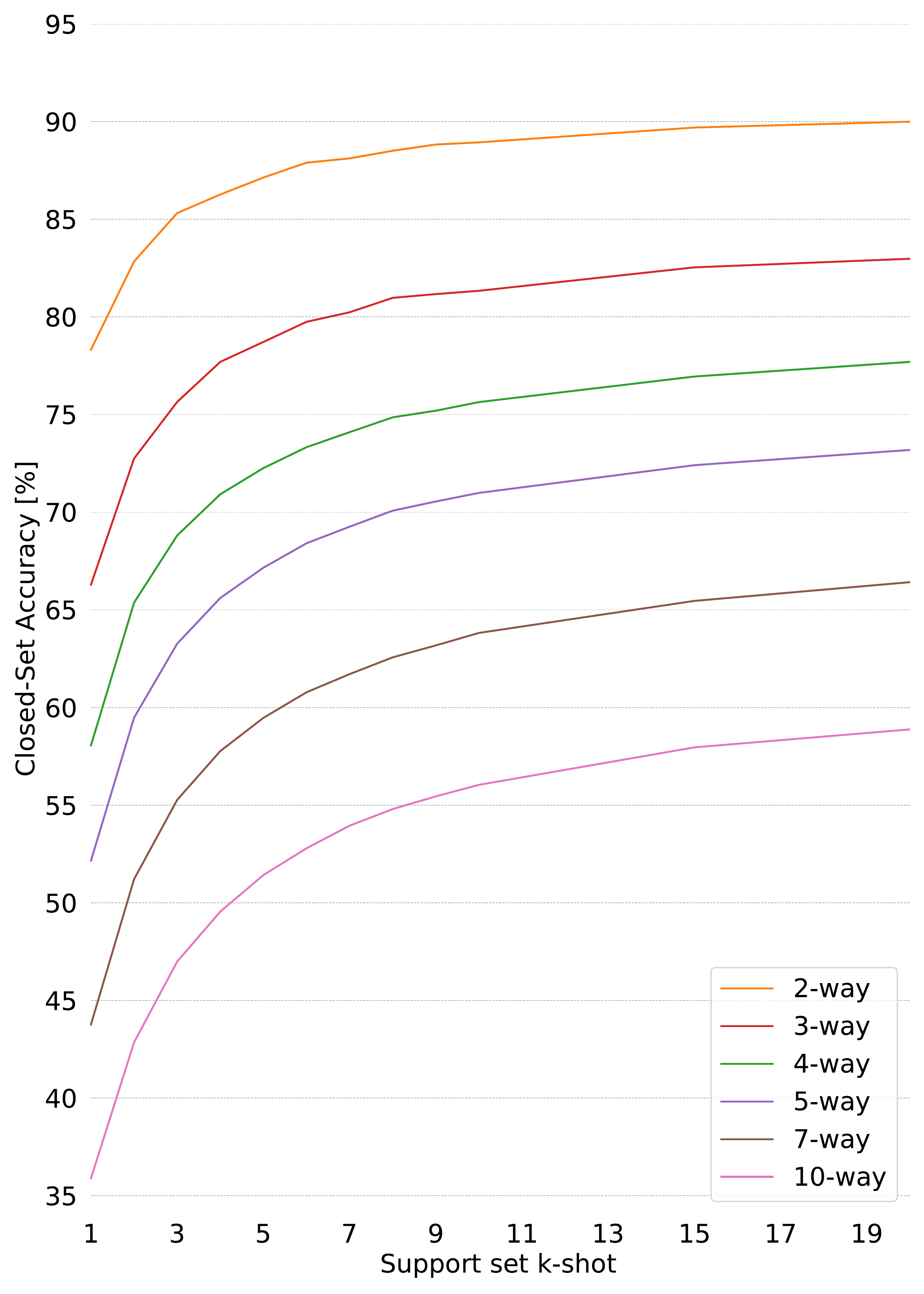}&
\includegraphics[width=5cm]{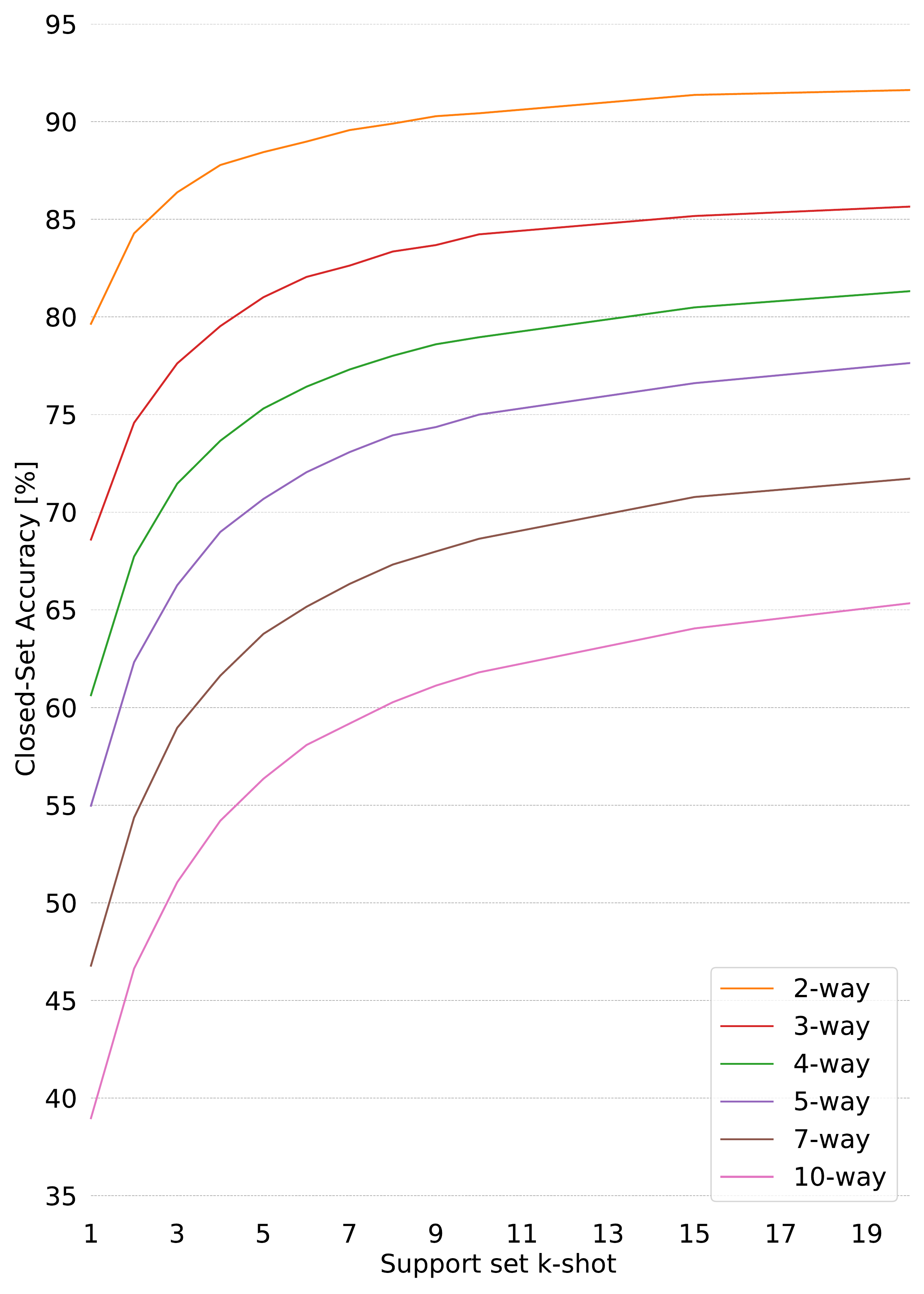}&
\includegraphics[width=5cm]{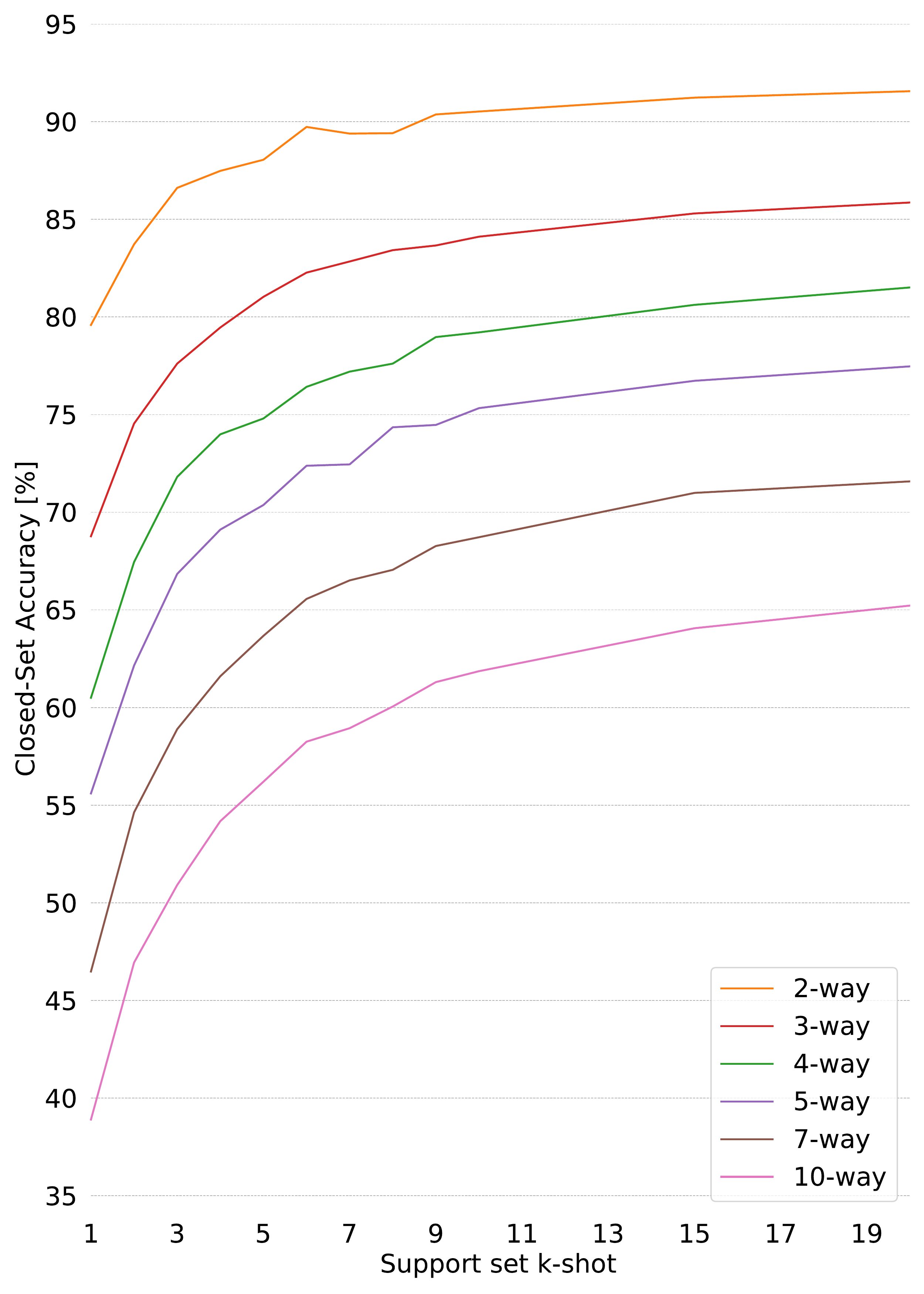}\\
(d) PEELER + \OCML & (e) FEAT + threshold & (f) FEAT + \MetaBCE \\ 
\includegraphics[width=5cm]{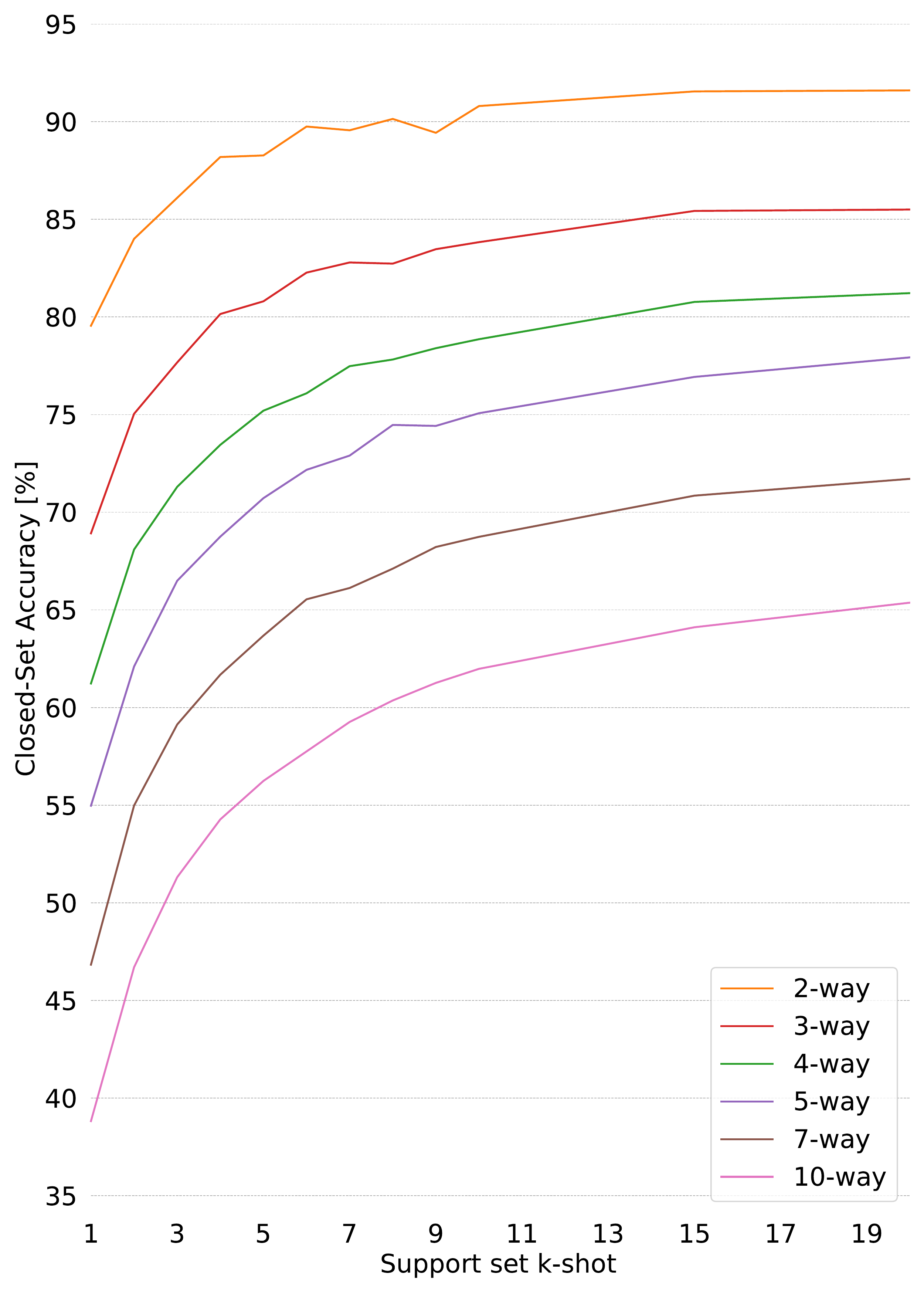} & &\\
(d) FEAT + \OCML &  &\\ 
\end{tabular}
\caption{Closed-set accuracy}
\label{fig:fsos_ablation_accuracy}
\end{center}
\end{figure}

\end{document}